\documentclass[preprint]{elsarticle}

\usepackage{lineno,hyperref}
\usepackage[normalem]{ulem}
\usepackage{seqsplit}
\RequirePackage{fix-cm}

\modulolinenumbers[5]

\journal{Journal of Applied Soft Computing}

\usepackage{mathtools}
\usepackage{commath}
\usepackage{algorithm}
\usepackage{algpseudocode}

\usepackage{multirow}
\usepackage{amsmath}
\usepackage{amssymb}
\usepackage{graphicx}
\usepackage{xcolor}
\graphicspath{./plot}

\begin{document}

\begin{frontmatter}

\title{Semantics in Multi-objective Genetic Programming}

\author{Edgar Galv\'{a}n\corref{mycorrespondingauthor} (ORCID: 0000-0001-8474-5234)}
\address{Naturally Inspired Computation Research Group, Department of Computer Science, Hamilton Institute, National University of Ireland Maynooth, Lero, Ireland}
\ead{edgar.galvan@mu.ie}
\cortext[mycorrespondingauthor]{Corresponding and leading author.}

\author{Leonardo Trujillo (ORCID: 0000-0003-1812-5736)}
\address{Tecnol\'ogico Nacional de M\'exico/IT de Tijuana, Tijuana, BC, M\'exico}
\ead{leonardo.trujillo@tectijuana.edu.mx}

\author{Fergal Stapleton (ORCID: 0000-0002-5347-1573)}
\address{Naturally Inspired Computation Research Group, Department of Computer Science, Hamilton Institute, National University of Ireland Maynooth, Ireland}
\ead{fergal.stapleton.2020@mumail.ie}

\begin{abstract}
Semantics has become a key topic of research in Genetic Programming (GP). Semantics refers to the outputs (behaviour) of a GP individual when this is run on a data set. The majority of works that focus on semantic diversity in single-objective GP indicates that it is highly beneficial in evolutionary search. Surprisingly, there is minuscule research conducted in semantics in Multi-objective GP (MOGP). In this work we make a leap beyond our understanding of semantics in MOGP and propose SDO: Semantic-based Distance as an additional criteriOn. This naturally encourages semantic diversity in MOGP. To do so, we find a pivot in the less dense region of the first Pareto front (most promising front). This is then used to compute a distance between the pivot and every individual in the population. The resulting distance is then used as an additional criterion to be optimised to favour semantic diversity.  We also use two other semantic-based methods as baselines, called Semantic Similarity-based Crossover and Semantic-based Crowding Distance. Furthermore, we also use the Non-dominated Sorting Genetic Algorithm II and the Strength Pareto Evolutionary Algorithm 2 for comparison too. We use highly unbalanced binary classification problems and consistently show how our proposed SDO approach produces more non-dominated solutions and better diversity,
leading to better statistically significant results, using the hypervolume results as evaluation measure, compared to the rest of the other four methods.
\end{abstract}

%\keywords{Multi-objective Genetic Programming \and Semantics \and Diversity}

\end{frontmatter}

%\linenumbers

\section{Introduction}
 Genetic Programming~\cite{Koza:1992:GPP:138936},  one of the four canonical Evolutionary Algorithms paradigms, was popularised by Koza in the early 1990s. Over the years, researchers have been interested in making GP more amenable to evolutionary search. A key element that has been proven to make GP more robust is semantics. The latter has become a key topic of research in GP.  Semantics can be seen as the behaviour of a GP program. This behaviour is the output of a GP program when executed on a set of fitness cases.
 
The number of scientific publications in GP semantics has increased significantly thanks to promising results found by the research community. We discuss in Section~\ref{sec:related} some relevant works of semantics in GP. Interestingly, the vast majority of these work have concentrated on Single-objective GP (SOGP), with minuscule progress in Multi-objective GP (MOGP), with the exception of ~\cite{DBLP:conf/gecco/GalvanS19,Galvan-Lopez2016,Galvan_MICAI_2016,9308386}.  Thus, this scientific work extends significantly this line of research and uses three forms of semantics in a MOGP setting. Each of these are compared independently against two well-established Evolutionary Multi-objective Optimisation (EMO) approaches: the Non-dominated Sorting Genetic Algorithm II (NSGA-II)~\cite{Deb02afast} and the Strength Pareto Evolutionary Algorithm (SPEA2)~\cite{Zitzler01spea2:improving}. The semantic-based MOGP approaches used in this study are:

\begin{description}
    \item \textit{Semantic Similarity-based Crossover.} (SSC). This is motivated by the SOGP approach presented in~\cite{Uy2011}. This approach was one of the early methods in SOGP semantics where the authors were able to promote it in continuous search spaces. We extended this well-known method in MOGP.
    \item \textit{Semantic-based Crowding Distance.} (SCD).  Here, the main idea is to replace the crowding distance, commonly used in EMO algorithms, by a semantic-based distance, originally studied in the first author's MOGP works~\cite{Galvan-Lopez2016,Galvan_MICAI_2016}.
    \item \textit{Semantic-based Distance as an additional criteriOn.} (SDO). This approach draws from SCD and uses the resulting semantic distance as another component to optimise by an EMO algorithm, briefly studied in~\cite{9308386}.%~\cite{DBLP:conf/gecco/GalvanS19}.
    
\end{description}

 Using these three semantic-based methods allow us to show the following: 
\begin{description}
\item {Firstly,} by using SSC, we show how the semantic distance computed in the crossover operator and used to successfully promote semantic diversity in single-objective GP does not have the same positive impact in MOGP. 

\item {Secondly,} by using SCD, inspired by the crowding distance commonly used in EMO, we show that a semantic distance can be naturally computed between every GP tree in the population and a ``pivot". The latter is an individual in the sparsest region of the first Pareto front. SCD then moves away from  SSC, which tries to promote semantic diversity by forcing diversity to emerge by using crossover repeatedly, as proposed in~\cite{Uy2011}.

\item {Finally,} we build from our understanding drawn from SSC and SCD in semantics and propose a robust mechanism for the emergence of semantic diversity in MOGP. Particularly, we use the semantic distance value as an additional indicator to evolve the population. This naturally promotes semantic diversity in MOGP leading to better, statistically significant results based on the average hyper-volume of the evolved Pareto approximations with respect to the other four methods (two semantic-based methods and two EMO methods) in a range of highly imbalanced data sets.
\end{description}

\subsection{Main contributions of this scientific study}
%\subsection{Initial limited study that paved the way for the new contributions of this scientific study}

{In our previous work~\cite{9308386}, we carried out an initial limited study on semantics in Multi-objective Genetic Programming (MOGP). Specifically, we initially proposed and used three semantic-based methods, named Semantic Similarity-based Crossover (SSC), Semantic-based Distance as an additional criteriOn (SDO) and Pivot Similarity Semantic-based Distance as an additional criteriOn (PSDO).}

{The main conclusions from our initial investigation is that the use of a semantic-based distance value as computed in either SDO or PSDO to be used as another objective to be optimised in an EMO setting is robust enough to outperform the results yield by the well-known NSGA-II and SPEA2 approaches. Furthermore, in our initial research, we found out that the distance computed from a pivot, which is the furthest point in the search space, to every individual in the population and used as an additional criterion to be optimised in a EMO setting tends to improve the performance of our semantic-based approaches. Moreover, we were able to fine-tune how this distance can be computed to significantly improve the evolutionary search. This is attained using the SDO approach, which is used again in this work. It is, however, worth saying that these conclusions were drawn from an initial limited study including a restricted statistical analysis impeding drawing general conclusions, limited results as well as a lack of explanation that help us to clearly indicate why SDO yields better results compared to their respective canonical methods as well as the other two semantic-based approaches. }

          {In this work, we have addressed all these issues. More specifically, the main contributions of this scientific study are as follows:}

  \begin{itemize}

   \item We consistently show how Semantic Similarity-based Crossover (SSC) used in single-objective GP and widely reported to be beneficial in GP does not have the same positive impact in a multi-objective GP (MOGP) setting.

\item    {From this, we show how a semantic-based distance approach can enhance the evolutionary search in MOGP. To this end we use two semantic-based approaches:  Semantic-based Crowding Distance (SCD) and a Semantic-base Distance as an additional criteriOn (SDO). }

    \item {We demonstrate how SDO yields better results against all the approaches used in this work, including the semantic-based methods and canonical EMO approaches.}

\item    {Another major contribution of this scientific study is to include detailed results using two well-established EMO approaches NSGA-II and SPEA2. By doing so, as opposed to the limited results reported in~\cite{9308386}, we are now in a position to draw sound conclusions by carrying out a systematic statistical analysis, explained in detail in Section~\ref{sec:results}.}

    \item {Another important contribution in this work is that we are able to explain why the semantic-based technique employed in SDO tends to improve evolutionary search. We do so by extensively analysing the behaviour of the SDO in terms of number of unique solutions, duplicate frequency of solutions over generations, etc. }

            \end{itemize}

This work is organised and presented as follows. Relevant studies to this work are presented in Section~\ref{sec:related}. The fundamental background in semantics and in MOPG is discussed in Section~\ref{sec:background}. Section~\ref{sec:approach} presents the MOGP semantic methods proposed and used in this work. The setup of experiments is presented in Section~\ref{sec:experimental}. Section~\ref{sec:results} presents in detail the results yield by all the MOGP semantic  approaches (SSC, SDO and SCD) and by the EMO methods (NSGA-II and SPEA2). It also offers an explanation as to why SDO finds better results compared to all the other algorithms. In Section~\ref{sec:conclusions}, we draw some conclusions.

\section{Relevant Work}
\label{sec:related}

\subsection{Semantics}

%The number of scientific studies in GP semantics has seen a surge in recent years due to the better results obtained by semantic-based methods compared to those approaches that do not promote semantic diversity explicitly. 

Semantics has become a key topic of research in GP and multiple definitions have been proposed. Semantics can be seen as the \textit{behaviour} (recorded outputs over a data set) of a GP program. We give a formal definition of semantics in Section~\ref{sec:background}. Research in semantics in GP has grown substantially in the last decade as a consequence of the research community reporting better results when semantics has been promoted in evolutionary search as compared to those GP approaches that do not promote it explicitly. The focus of these studies range from dealing with direct semantic methods, such as the use of geometric operators~\cite{DBLP:conf/ppsn/MoraglioKJ12}, to the analysis of indirect semantic methods~\cite{Galvan-Lopez2016,Uy2011}. Next, we discuss some relevant works in this area.

The analysis of McPhee et. al.~\cite{McPhee:2008:SBB:1792694.1792707} laid the foundations for indirect semantic works. Their research focused on analysing the semantics of program subtrees and contexts (a context being the remaining portion of a program after a subtree has been removed) for boolean problems. A key outcome from this research demonstrated that the commonly used ratio 90-10 crossover produced a high proportion of individuals which were semantically equivalent. In other words, the majority of crossover events do not result in an effective search of the semantic space and consequently limits the potential performance benefits of performing this operation. 

To overcome this, Beadle and Johnson~\cite{4630784} proposed an operator that would help promote semantic diversity dubbed Semantically Driven Crossover (SDC). To verify whether parents and corresponding offspring are equivalent or not, the authors used reduced ordered binary decision diagrams to this end. In their studies, the authors applied crossover multiple times when both parents and offspring were semantically equivalent. Later, Beadle and Johnson~\cite{4983099} also explored a similar technique for the mutation operator, called Semantically Driven Mutation (SDM). This verifies the semantic equivalence between a parent and a offspring, where the latter is generated by replacing a subtree with another randomly generated subtree. This equivalence analysis is carried out by reducing the offspring to a canonical form which then can be easily compared to its parent, that has also been reduced. They reported that both techniques increased semantic diversity and led to improved evolutionary search.

A drawback of these methods is that they use discrete fitness-value cases, hence limiting their applicability to continuous search spaces. Uy et al.~\cite{Nguyen:2009:SAC:1533497.1533524} overcame this limitation by using a motivational approach which involves employing ingenious semantic crossover operators in continuous search spaces. The semantics are approximated by evaluating a predefined sample of points from a given problem domain. The semantic equivalence therefore of two expressions (two trees or two subtrees) can be calculated from the absolute difference between the outputs of these expressions. If the difference of these two expressions fall within the bounds of a predefined threshold (a parameter referred to as Semantic Sensitivity) then these expressions can be deemed to be semantically equivalent. Uy et al. proposed four different scenarios for implementing this form of semantics, the first two of which dealt with the semantics of sub-trees. Scenario I sought to promote semantic diversity by checking the semantic equivalence of subtrees used in a crossover operation and if they were deemed to be equivalent, the parents were retained. Otherwise, the crossover was applied again, but now using two different randomly selected crossover nodes. In Scenario II, the subtrees are picked if these are equivalent to each other. The later two scenarios consider semantics by taking into consideration the full program trees. Scenario III checks the semantics of the parents in relation to their offspring. That is, if the child and parent trees are found to be semantically equivalent the offspring are discarded and the parent is retained into the next generation. Scenario IV is the same as Scenario III but in this case the condition is reversed, in other words the offspring are retained into the next generation if the semantic equivalence criteria is met. Scenario I was shown to produce better results on symbolic regression problems when compared with the other three methods.

There is no guarantee that a semantically equivalent offspring will be found immediately after performing the crossover operation and an extension of the above method introduced a trial and error mechanism to perform the crossover operation multiple times until a suitable candidate is found or until some predefined iteration is reached \cite{Uy2011}. However, a significant downside is that the Uy et al. method can be computationally high-intensive. To address this drawback, Galv\'an et al. put forward a cost-effective approach by trying to promote semantic diversity through the tournament selection operator. The first parent is selected as usual. The second parent is selected by considering: semantic dissimilarity and fitness, in that order. If there is no individual that is dissimilar to the first parent, then the second parent is selected as commonly done in tournament selection. The semantic dissimilarity is obtained in the same manner as done by Uy et al.~\cite{Nguyen:2009:SAC:1533497.1533524} and described above in Scenario I. This cost-effective approach performs similar to Uy el al. method with the benefit of removing the expensive trial and error mechanism. 

%\TODO I need to include Pawlak's work and then make a reference to this from Dou and Rockett's work

 Forstenlechner et al.~\cite{Forstenlechner:2018:TES:3205455.3205592} investigated the use of semantics for program synthesis, a different domain from those discussed before (Symbolic Regression and Boolean problems). A variety of different data types are used in this approach as opposed to just a single data type. The semantics of two GP subtrees are stored in a pair of vectors. In their work, the authors used two conditions for checking semantic similarity 1) `Partial change’ which denotes a vector whose semantics have stayed the same for at least one entry but have changed for another entry; and 2) `Any change’ which denotes a vector where any entry is different. Two subtrees undergo crossover if a partial change occurs. If this is not present, then the `any change' is checked. The authors reported better results when adopting these two conditions in four out of the eight problems used in their studies. An analysis of semantic methods have also been applied to local search methods in respect to indirect semantics. For example, Dou and Rockett~\cite{Dou2018} tested a mixture of various GP and local search variants, including three subtree selection methods and four replacement strategies. The authors found that a semantic-based local search following a steady-state or generational GP performed significantly better compared to baseline GP methods with statistically smaller tree sizes.

Direct semantic approaches have also received attention in the GP community thanks to the improvement on search performance compared to a standard GP system. The driving motivation by incorporating direct semantics is that previous indirect methods were deemed to be wasteful~\cite{Nguyen:2009:SAC:1533497.1533524,Uy2011}. To tackle this issue, Moraglio et al.~\cite{DBLP:conf/ppsn/MoraglioKJ12} used his previous theoretical results~\cite{Moraglio2004} to allow modifications on the genotype of GP trees to correlate to geometric operators. This had the consequence of inheriting their properties. This results in having a cone landscape by construction, providing to the evolutionary process an ``easier'' search direction. One potential limitation of this approach, however, is that it allows for the presence of neutrality. This can be beneficial or detrimental depending on the features of the problem at hand -- see, for example,~\cite{DBLP:conf/gecco/LopezP06,DBLP:journals/evs/LopezPKOB11,DBLP:journals/tec/PoliL12,10.1007/978-3-540-73482-6_9}   where Galv\'an and Poli give an in-depth explanation on the effects of neutrality in different type of problems. One more limitation in the approach proposed in~\cite{DBLP:conf/ppsn/MoraglioKJ12} is that this modification to GP trees tends to produce larger individuals. To deal with the latter, Vanneschi et al.~\cite{Vanneschi2013}  proposed a cache implementation of Moraglio's approach. To this end, the authors store the semantics of GP trees in a table making the process efficient indeed. However, one limitation in Vanneschi et al. approach is that the reconstruction of GP individuals is cumbersome and difficult to obtain in some cases. This is a drawback specially for applications where the expression is required. Uy et al.~\cite{Nguyen2016} took a different approach compared to Vanneschi et al.~\cite{Vanneschi2013} to deal with the size of individuals. Uy et al.  proposed subtree semantic geometric crossover (SSGX) operator that allows them to control the size of individuals. However, their approach has some limitations too such as determining the right values for multiple parameters that are needed to apply SSGX in a GP system. Moreover, it is also based on an expensive trial-and-error mechanism (a maximum of 20 trials is set in their work to search for suitable subtrees).

          {In~\cite{9308386}, we proposed semantic-based methods to integrate them into a MOGP system. In this preliminary study, we were able to promote semantics naturally in a MOGP framework by using a pivot taken from the best Pareto front to compute the semantic distance between this and every individual in the population. We found that this distance and a variant of it tend to produce better results compared to its canonical evolutionary multi-objective variant, in this case, against the well-known NSGA-II~\cite{Deb02afast}  and SPEA2 \cite{Zitzler01spea2:improving} algorithms. However, the findings reported in~\cite{9308386} were limited: a lack of explanation as to why SDO tends to yield better results compared to the other methods used in the initial study, a lack of an in-depth statistical analysis and of a discussion on the limitation of the SDO approach. In contrast, the current work addresses all these issues, as we should see in the next sections. In particular, we incorporated in detail the results yielded by the SPEA2 algorithm as well as the semantic-based variants using this EMO approach, we explain in detail why SDO works by using different elements such as the frequency of duplicated individuals in the population over generations. We also carried out a detailed statistical analysis of the results. }

\subsection{Multi-Objective Genetic Programming}

The objective of a multi-objective optimisation problem is to discover candidate solutions based on the simultaneous consideration of multiple, potentially conflicting, objectives. In the context of GP there are a number of ways that this can be achieved. Broadly speaking there are two primary methods. One is to incorporate multiple objectives into a single fitness function. Another method is to consider the objectives separately making use of the  \textit{Pareto dominance} relationship between candidate solutions~\cite{1597059,CoelloCoello1999,Deb:2001:MOU:559152}. Since the aim of Evolutionary Multi-objective Optimisation (EMO) is to discover the best balance of solutions amongst the objectives of an evolutionary run, Pareto dominance allows for an intuitive approach to handle multiple (conflicting) objectives.  The details of EMO will be explained further in Section~\ref{sec:backgroundMOO}. EMO is one of the most popular and active research area in EAs, with many applications and often achieves impressive results~\cite{1597059,CoelloCoello1999,Deb:2001:MOU:559152}. Next, we discuss just a few works that have been adopted EMO within GP systems. 

%Bloat, dramatic increase of GP individuals' size as evolution progresses, is a phenomenon commonly seen in GP runs.
Bleuler et al.~\cite{934438} proposed an EMO approach to naturally control bloat on even-n-parity problems. The authors defined two objectives to be optimised simultaneously within a MOGP framework; the first objective being the fitness of a program and the second objective being the size of the program tree. This method was compared against other well-known techniques for controlling bloat, such as using an aggregated single-objective function that includes both fitness and parsimony pressure (both Constant and Adaptive Parsimony Pressure were tested) and also via a two stage optimisation process. This approach was not only novel at the time, but also demonstrated how to successfully control for GP bloat, leading to solutions which could be evaluated faster when compared with the other methods.

Related to this study is the work carried out by Bhowan et al.~\cite{6198882}, where the authors used a MOGP to find high accuracy on binary unbalanced classes achieving good results compared to well-established machine learning classification methods. Galv\'an et al. also used MOGP for the same type of problems~\cite{DBLP:conf/gecco/GalvanS19,Galvan-Lopez2016,Galvan_MICAI_2016}. In the same vein, Zhao showed how MOGP can be successfully employed to define partial preferences on the objectives by carefully inserting this bias into the fitness function~\cite{Zhao2007809}. The motivating reason to incorporate this embedding is that classification errors are often cost-sensitive in real life scenarios, where the benefits of a correct prediction on one class may significantly outweigh the correct prediction on another. For example, it is more costly to approve a bad loan than denying a good loan. 

Shao et al.~\cite{6683022} demonstrated how images can be classified using domain-adaptive global features that are automatically generated. Similar to how Bleuler et al.~\cite{934438} implemented their optimisation process, Shao et al. use the length of the individuals as an objective to be optimised in their MOGP framework, where the second objective is the classification error rate.  Shao et al. reported that their MOGP approach consistently yielded better performance when compared against fourteen other methods, including two neural network-based approaches.
%Other examples include the use of MOGP to detect interest points in images, a low-level feature extraction procedure that is often used as an initial step in feature-based scene analysis and object recognition techniques \cite{ip1,ip2}. In this domain, MOGP has produced novel image operators with unique properties that are not present in standard techniques \cite{ip1}.

\section{Background}
\label{sec:background}
This section defines some of the basic concepts relevant to this work, namely semantics, MO and EMO algorithms.

\subsection{Semantics}

We use a well-established definition of semantics originally defined in~\cite{6808504}. For a supervised learning task in general, and for GP in specific, the problem is normally specified as a set of input--output pairs,
also known as fitness cases, of the form $T = \{(in_i,o_i)\}$, where $in_i \in I$ are the inputs and $o_i \in O$ are the desired
outputs, and $i=\{1, \cdots, l\}$, where $l$ is the number of the fitness cases. The semantics $s(p)$ of a program $p$ is defined as the vector of output values computed by this program based on the inputs
given in the set of fitness cases of a problem. This is formally defined as,

\[
s(p) = [p(in_1), p(in_2), \cdots, p(in_l)] \ .
\label{expression:semantics}
\]

\subsection{Multi-Objective Optimisation}
\label{sec:backgroundMOO}
 
The goal in Multi-objective optimisation (MO) is to simultaneously optimise two or more objective functions. When multiple objective functions are considered, often these will be in conflict, so the focus then is to search for a set of trade-off solutions as a global optimum becomes unattainable. A natural form to solve this problem is to use the Pareto dominance relation: a solution $x_1$ in the search space is said to {\em Pareto-dominate} solution $x_2$ if $x_1$ is at least as good as $x_2$ for all objectives and strictly better for at least one of the objectives.

In this work, the objectives are to be maximised. The class of problems used in this work, defined in Section~\ref{sec:experimental}, is binary classification problems where the goal is to  maximise the classification accuracy of two conflicting objectives. The \textit{Pareto dominance} concept is defined as in Eq. (\ref{eq:paretoDominance}):

\begin{equation}
  S_i \succeq S_j \longleftrightarrow \forall_m [(S_i)_m \ge (S_j)_m] \wedge \exists k [(S_i)_k > (S_j)_k] \
\label{eq:paretoDominance}
\end{equation}

\noindent where $(S_i)_m$ denotes the value of solution $S_i$ in the  $m$th objective.
Conversely, solutions are considered to be \textit{non-dominated} if no other solution in the population dominates them. The set of optimal trade-off solutions of a MO problem at hand is referred to as the {\em Pareto optimal set}. Thus, the objective of an EMO algorithm is to find (a good approximation) of this set. The {\em Pareto optimal front} is the objective space representation of the Pareto optimal set.

Pareto-dominance has been used in different ways to handle such criterion to bias the search in an EMO algorithm. Some of the most widely-known are dominance rank~\cite{Deb02afast} and dominance count~\cite{934438}. 
Dominance rank is based on the number of solutions that dominate another within a population, this means that a lower value is desirable.
On the other hand, dominance count is the number of solutions that a particular solution dominates, in this case a higher value is preferable.
There are two popular EMO algorithms that include the aforementioned ways to use Pareto-dominance and they are adopted in this study: the Non-dominated Sorting Genetic Algorithm II (NSGA-II)~\cite{Deb02afast} that uses dominance rank, and the Strength Pareto Evolutionary Algorithm (SPEA2)~\cite{934438} that uses dominance rank and dominance count.

In NSGA-II, dominance rank is used as a fitness value for solution S$_i$. This is expressed as

\begin{equation}
  \label{eq:fitnessNSGAII}
   \textnormal{NSGA-II}(S_i) =\{\nexists j, j \in Pop | S_j \succeq S_i\} \ .
  \end{equation}

SEA2, on the other hand, uses and dominance count and dominance rank when computing the fitness value of an individual. First, a \textit{strength} value $D$ is given to every candidate solution in the population. This is formally defined by, 

\begin{equation}
D(S_i) = | \{j|j \in Pop \wedge S_i \succeq S_j\}|.
\label{eq:fitnessSPEA2-1}
\end{equation}

$D$ then determines the number of solutions that a particular solution (S$_i$) dominates. To determine the fitness value of a candidate solution $i$ in SPEA2, we then use the strengths of all the solutions that dominate solution $i$. This is expressed as,

\begin{equation}
\textnormal{SPEA2}(S_i) = \sum_{j \in Pop, S_j \succeq S_i} D(S_j) \ .
  \label{eq:fitnessSPEA2}  
\end{equation}

%\noindent As in NSGA-II, fitness is minimized in SPEA2, such that non-dominated solutions have the best possible fitness of zero. 

\subsubsection{Diversity Preservation in Multi-objective Optimisation Through the Use of the Crowding Distance}
\label{sec:crowding}

%In general, in MO problems it is necessary to promote diversity in an evolutionary algorithm since the goal is to attain solutions that provide different trade-offs in objective space. However, 

A total order is not provided by Pareto dominance, such that another criterion is required to effectively
compare different points in the search space, which is necessary to perform selection and survival during evolution.
One approach is to use a \textit{crowding distance} measure,
such that regions that are sparsely populated are preferable compared to denser regions.
The crowding distance is normally used in objective space to differentiate between individuals having the same Pareto rank, preferring those that are in the less populated regions. Algorithm~\ref{alg:crowding} shows how this crowding distance is computed, adapted from~\cite{Deb02afast}.

\begin{algorithm}
  \caption{Crowding distance. Adapted from~\cite{Deb02afast}.}
\begin{algorithmic}[1]
\Procedure{Crowding-distance-assignment($I$)}{}
\State  $l$ $\gets$ $|I|$ \Comment{number of solutions in $I$}
\For {each $i$}
 \State $I$[$i$]$_{distance}$ $\gets$ 0 \Comment{initialise distance}
\EndFor
\For {each objective $m$}
 \State $I$ $\gets$ sort($I$,$m$) \Comment{sort using each objective value}
 \State $I$[1]$_{distance}$ $\gets$  $I$[$l$]$_{distance}$ $\gets \infty$ \Comment{so that boundary points are} \newline \raggedleft \text{always selected} 
 \For  {$i$ $\gets$ 2 to ($l-1$)}  \raggedright \Comment{for all other points} 
 \State $I$[$i$]$_{distance}$ $\gets$ $I$[$i$]$_{distance}$ + ($I$[$i$+1].$m$-$I$[$i$-1].$m$)/($f^{max}_m$-$f^{min}_m$)
 \EndFor
\EndFor
\EndProcedure
\end{algorithmic}
\label{alg:crowding}
\end{algorithm}

%Because Pareto dominance does not provide a total order, an additional criterion must be used so as to allow the comparison of any pair of points of the search space, i.e., \textit{crowding distance} measure. This is the Manhattan distance between solutions in the objective space, where sparsely populated regions are preferred over densely populated regions. This distance is only used to resolve selection when the primary fitness is equal between two or more individuals. Thus, the crowding distance promotes diversity among individuals having the same Pareto rank: in objective space, for each objective, the individuals in the population are ordered, and the partial crowding distance for each of them is the difference in fitness between  its two immediate neighbours. The crowding distance is the sum over all objectives of these partial crowding distances~\cite{Deb02afast}. Intuitively, it can be seen as the Manhattan distance between the extreme vertices of the largest hypercube containing the point at hand and no other point of the population. Selecting points with the largest crowding distance amounts favours sparse regions of the objective space, i.e., it favours  diversity.

\subsubsection{MOGP Algorithm} 

\label{sec:mogp}
The MOGP framework employed in this work is based on NSGA-II, which is described next. This is the framework that we use for all our experiments and we only change the way fitness of individuals is computed by using either NSGA-II  or SPEA2, formally defined in Equations~\ref{eq:fitnessNSGAII} or~\ref{eq:fitnessSPEA2-1}, respectively.

Both populations (parents and offspring) are joint at every generation. The best individuals from this overall population are copied into the archive population, which contains the same number of individuals as the original population. The archive population then serves as the parent population for the next generation. The archive population provides elitism preserving the set of non-dominance solutions throughout the evolutionary process.

\section{Semantic-based MOGP Methods}
\label{sec:approach}

Next, we present the semantic-based approaches employed in this work that are incorporated into the 
baseline MOGP algorithms, namely NSGA-II and SPEA-2.

\subsection{Semantic Similarity-based Crossover MOGP}
\label{sec:scc}
The first approach that we consider to incorporate semantics within MOGP is the Semantic Similarity-based Crossover (SSC), originally proposed by Uy et al.~\cite{Uy2011} for single-objective genetic programming.

SSC requires a semantic distance. The distance is computed as the average of the absolute difference of values for every $in \in I$ (or a partial set of inputs)  between parents and offspring.
When the distance value falls within a specific range, defined by one or two bounds, an offspring is generated by way
of crossover. Given that this  may be difficult to meet, the original approach encourages diversity by repeating crossover with a maximum of 20 attempts.
If the condition is not satisfied, then crossover is execute as usual.

%To use SSC in single-objective GP a semantic distance must be computed first. This distance is obtained by computing the average of the absolute difference of values for every $in \in I$ between parent and offspring. If the distance value lies within a range, defined by one or two bounds, then crossover is used to generate offspring. Because this condition may be hard to satisfy, the authors tried to encourage semantic diversity by repeatedly applying crossover up to 20 times. If after this, the condition is not satisfied, then crossover is executed as usual.

%To promote semantic diversity, Uy et al.~\cite{Uy2011} calculated a semantic distance between parents and offspring. Using Def.~\ref{def:semantics}, the authors computed this distance by calculating the average of the addition of difference values  for every $in \in I$ between parent and offspring. If the distance value lies within  a range, then crossover is promoted to generate offspring. Because this condition may be hard to satisfy, the authors tried to encourage semantic diversity by repeatedly applying crossover up to 20 times. If after this, the condition is not satisfied, then crossover is executed as normal.

SSC was a notable contribution to GP since it showed that it was feasible to promote semantic diversity in continuous search spaces, leading to several subsequent studies~\cite{DBLP:conf/gecco/GalvanS19,DBLP:conf/eurogp/AnhNNO13,6557931,Krawiec:GPEM2013,6063448}.
In this work we implement SSC as proposed in~\cite{Uy2011}, but for the first time the method is evaluated as part of a MOGP framework, using both NSGA-II and SPEA2 as the baseline algorithms. Our results, discussed in Section~\ref{sec:results}, show that unlike for single-objective GP, MOGP SSC did not lead to notable performance improvements.

\begin{figure}[tb!]
   \includegraphics[width=0.85\columnwidth]{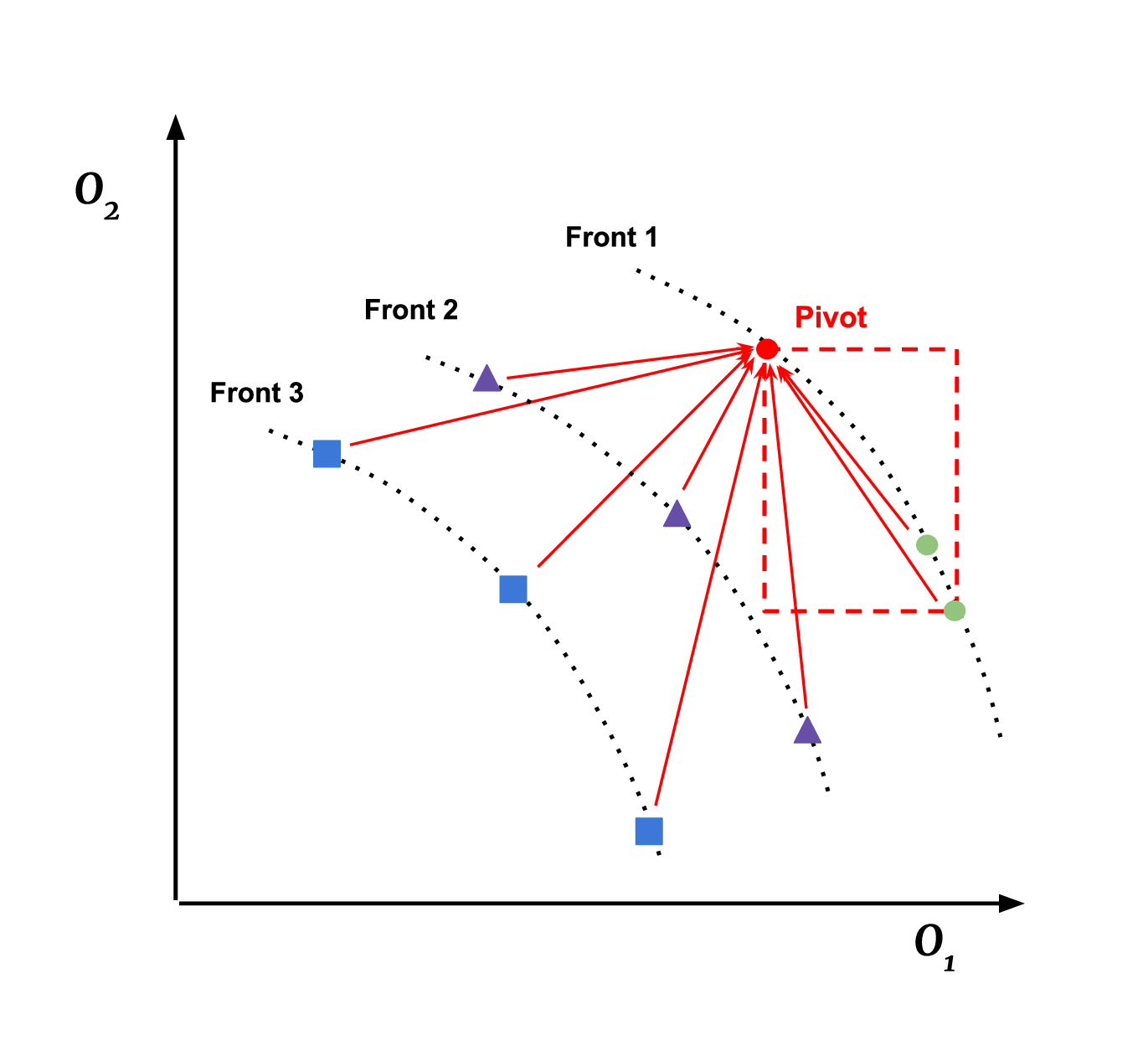}
\caption{Semantic-based Distance as an Additional CriteriOn. First, we get non-dominated solutions using either dominance rank (see Eq.~\ref{eq:fitnessNSGAII}) or the dominance rank and dominance count to determine the strength of a given solution (see Eq.~\ref{eq:fitnessSPEA2}) and store these in R$_t$. Second, once the non-dominated fronts have been found, we proceed to find a pivot from the first non-dominated front. To do so, we use the crowding distance defined in Section~\ref{sec:crowding}. This pivot is in the sparsest region of the front (the dotted red rectangle depicts this idea). Third, we compute the semantic distance (either using Eq.~\ref{semantic:distance:two:values} or Eq.~\ref{semantic:distance:one:value}), from the pivot to each individual in R$_t$. Fourth, the distance values are used as an additional criterion for the EMO to optimise, along with conflicting objectives, O$_1$ and O$_2$, namely the TPR and the TNR for unbalanced binary classification problems used in this work.}
\label{fig:SDO_aproach}
\end{figure}

\subsection{Semantic-based Crowding Distance}
\label{sec:semanticBasedDistance}

In Semantic-based Crowding Distance (SCD), the key element is to define an individual (pivot) that can be used to compute a semantic distance between this and some elements from the population. Algorithm~\ref{alg:sbcd} shows how the SCD MOGP works. It creates a new population P$_{t+1}$ using the non-dominated sorted solutions from the parent population P$_t$ and the offspring population Q$_t$, which are merged into population R$_t$ (Lines 2 -- 3).  This continues until the size of P$_{t+1}$ is equal to the size of P$_t$ (Lines 4 -- 12). It may be the case that a particular front does not fit entirely into the new population and a second criterion is necessary to complete P$_{t+1}$. If this is the case, we proceed to store (F$_r$), the remaining of those individuals that have not been used to complete the population (Lines 13 -- 24). We then use the semantic distance as a criterion to select those points from F$_r$ to complete P$_{t+1}$.  To do so, we proceed to find a pivot $v$ which is the furthest point from the first front using the crowding distance, explained in Section 3.2.1 (Lines 15 – 16). The dotted red rectangle in Figure~\ref{fig:SDO_aproach} exemplifies how the pivot (red dot) is chosen from the first front.

The semantic distance between every point in F$_r$ and the pivot is computed (Line 17). Thus, only one pivot is necessary to compute this distance.  Formally, this distance is computed as in Eq. (\ref{semantic:distance:two:values}):

\begin{equation}
  \label{semantic:distance:two:values}
    d(p_j,v) = \sum_{i=1}^l 1 \text{ if } \text{LBSS} \leq |p(in_i) - v(in_i)| \leq \text{UBSS} 
  \end{equation}

\noindent where $p_j$ is an individual in $R_t$, $l$ is the number of fitness cases, and LBSS and UBSS are the lower bound and upper bound for the semantic similarity values, respectively. These last two values are used to promote semantic diversity within a range, as reported beneficial in~\cite{Uy2011}.

There is also a number of studies that have concluded that only one bound is necessary to promote semantic diversity~\cite{DBLP:conf/gecco/GalvanS19,Galvan-Lopez2016,Nguyen2016}. We can compute the semantic distance between the pivot $v$ and every individual in R$_t$ using the following distance formally described in Eq. (\ref{semantic:distance:one:value}),

\begin{equation}
\label{semantic:distance:one:value}
  d(p_j,v) = \sum_{i=1}^l 1 \text{ if }   |p(in_i) - v(in_i)| > \text{UBSS}  
\end{equation}

%The semantic distance between every point in R$_t$ and the pivot is computed using either Eq.~\ref{semantic:distance:two:values}  (two threshold values are necessary) or Eq.~\ref{semantic:distance:one:value} (only one threshold value is neeeded).

We then use the semantic distance values on the stored  F$_r$  (Line 18) and select the individuals that are in sparse regions
of the search space, until we complete P$_{t+1}$ (Lines 20 -- 23).

%In SCD, we show how it is possible to compute the semantic distance within a EMO framework and encourage semantic diversity. Because this is promoted only when a particular front does not fit entirely into the new population, we should expect a minor positive impact in evolutionary search, if any.

\begin{algorithm}
  \caption{Semantic-based Crowding Distance}

\begin{algorithmic}[1]

\Procedure{Forming New Population}{}
\State R$_t \gets P_t \cup Q_t$ %\Comment{Parent and Children pop. are merged}
\State F $\gets$ \text{non\_dominated\_sort} (R$_t$)% \Comment{all nondominated fronts of R$_t$}
\State P$_{t+1} \gets 0$; i $\gets$ 1;
\While {$|$P$_{t+1}|$ $\leq$ $|$P$|$} %\Comment{Fill Parent pop. until $|$F$_i|$ fits into P$_{t+1}$ }
 \If {$|$P$_{t+1}|$ + $|$F$_i|$ $<$ $|$P$|$} 
   \State P$_{t+1}$ $\gets$ P$_{t+1}$ $\cup$ F$_i$
%\State F$_{t+1}$ $\gets$ F$_{t+1}$ + F$_i$
   \State i $\gets$ i+1
 \Else
   \State break
 \EndIf
\EndWhile 
\If {$ |$P$_{t+1}|$ $<$  $|$P$|$}
\State F$_r$ $\gets$ F $\cap$ $\{$F$_1\cdots$F$_i\}$

\State CD$_1$ $\gets$ \text{crowding\_distance}(F$_1$)
\State pivot $\gets$ \text{furthest\_point}(CD$_1$)
\State SV $\gets$ \text{compute\_semantics(F$_r$, pivot)}
\State F$_r$ $\gets$ \text{crowding\_distance}(SV(F$_r$))

\State  j $\gets$ 0
\For {j + $|$P$_{t+1}|$ $<$ $|$P$|$}
 \State P$_{t+1} \gets$ P$_{t+1}$ $\cup$ F$_{r[j]}$
 \State  j $\gets$ j + 1
\EndFor
\EndIf

\EndProcedure
\end{algorithmic}

\label{alg:sbcd}
%}

\end{algorithm}

\begin{algorithm}
\caption{Semantic-based Distance as an Additional Criterion}
\begin{algorithmic}[1]
\Procedure{Forming New Population}{}
\State R$_t \gets P_t \cup Q_t$ %\Comment{Parent and Children pop. are merged}
\State F $\gets$ \text{non\_dominated\_sort} (R$_t$)% \Comment{all nondominated fronts of R$_t$}
\State CD$_1$ $\gets$ \text{crowding\_distance}(F$_1$)
\State pivot $\gets$ \text{furthest\_point}(CD$_1$)
\State R$_t$ $\gets$ \text{compute\_semantics(R$_t$, pivot)}
%\State F$_i$ $\gets$ \text{crowding\_distance}(SV$_i$)
\State F $\gets$ \text{sort} (R$_t$)% \Comment{all nondominated fronts of R$_t$}
\State  P$_{t+1}$ $\gets$ 0;  j $\gets$ 0
\For {j $<$ $|$P$|$}
 \State P$_{t+1} \gets$ P$_{t+1}$ $\cup$ F[j]
 \State j $\gets$ j + 1
\EndFor
\EndProcedure
\end{algorithmic}
\label{alg:sbdo}
\end{algorithm}

\subsection{Semantic-based Distance as an Additional CriteriOn}
\label{sec:sdo}

 We further expand SCD by now using the resulting semantic distance values as another indicator to select solutions by the MO process.
We refer to this approach as Semantic-based Distance as an additional criteriOn (SDO). As with SSC, we continue using the MOGP framework described above, for SDO. This approach is described in detail in Algorithm~\ref{alg:sbdo}.  

First,  we merge  the parent population P$_t$ and the offspring population Q$_t$ into R$_t$. We then get the non-dominated sorted  solutions (Lines 2 -- 3).
To compute the semantic distance for each point contained in R$_t$ we proceed as follows. We first compute the crowding distance from the first front and select the point that is the furthest away. We use this as pivot $v$ to then compute the semantic distance between $v$ and each individual contained in R$_t$ (Lines 4 -- 5). We then use the resulting distance value, using either Eq.~(\ref{semantic:distance:two:values}) (two bound values are required) or Eq.~(\ref{semantic:distance:one:value}) (one bound value is necessary), as another criterion to be computed (Line 6). Next, we sort R$_t$   (Line 7) and form the new population P$_{t+1}$ (Lines 8 -- 12). In case the size of P$_{t+1}\cup$ F[j] is greater than $\abs{P_t}$, we proceed to take only the individuals needed to complete P$_{t+1}$. Figure~\ref{fig:SDO_aproach} depicts this idea.

\section{Experimental Setup}
\label{sec:experimental}
  \begin{table*}
\caption{Summary of the binary unbalanced classification datasets.}
\centering
\footnotesize
%\resizebox{1.00\textwidth}{!}{ 
\begin{tabular}{llccc}
\hline
Dataset & Positive/Negative Class    & Number of examples                  & Ratio     & Features \\
         & Brief description          &  Total/Positive(\%)/Negative(\%)     &           & No.(Type) \\ \hline
Ion      & Good/Bad:                            & 351/126(35.8\%)/225(64.2\%)       &1:3     &34(Real)  \\
         & Ionosphere radar signals                                     & & &  \\
Spect  & Abnormal/Normal                        & 267/55(20.6\%)/212(79.4\%)        &1:4     & 22(Binary)  \\
       & Tomography scan                & & &  \\
Yeast$_1$& MIT/Other                            &1482/244(16.5\%)/1238(83.5\%)     &1:6     & 8(Real)  \\
         & Protein sequences                                     & & &  \\
Yeast$_2$& ME3/Other                            & 1482/163(10.9\%)/1319(89.1\%)     &1:9     & 8(Real)  \\
         & Protein sequences                                     & & &  \\
Abal$_1$ & 9/18                                 & 731/42(5.7\%)/689(94.2\%)     & 1:17       &  8(Real) \\
         & Biology of abalone                                     & & &  \\
Abal$_2$ & 9/Other                             & 4177/32(0.77\%)/4145(99.2\%)   &  1:130      &  8(Real) \\
         & Biology of abalone                                    & & &  \\
\hline
\end{tabular}
%}
\label{tab:datasets}
\end{table*}

            {The use of benchmark problems has allowed the research community to test, validate and explain a plethora of evolutionary algorithms. In this work, we also adopt well-known, robust and tested benchmark problems used in other studies~\cite{6198882,10.1007/978-3-642-10439-8_38,9308386} that will allow us to (i) test the algorithms used in this work, (ii) to use well-defined metrics that allow us to compare one method against another one, (iii) to allow us to explain why one particular method behaves better than others, (iv) to draw sound conclusions on the results reported in the following sections. Thus, for} this study, the impact of semantics in MOGP are analysed using several
 unbalanced binary classification problems, taken from the UCI Machine Learning repository~\cite{Asuncion+Newman:2007}. Table~\ref{tab:datasets}, adapted from~\cite{6198882}, gives the details of all datasets used in this work. These binary classification problems have various degrees of class imbalance, from 1:3 to 1:130, for the Ion and the Abal$_2$ data set, respectively. The 50-50\% training/test sets also range from being well-represented (Abal$_2$ has around 2,100 instances) to sparsely represented (Spect has around 133 instances, where around 27 are from the minority class). Moreover, these data sets range between low dimensionality (Yeast$_1$ has 8 features) to high dimensionality (Ion has 34 features). Finally, our data sets include binary and real-valued features. Thus, these data sets represent class imbalance problems of various degrees of difficulty, size, dimensionality and types of features reasonably well.  Moreover, we carefully choose these benchmark problems so that our evaluations on the three semantic-based methods (SSC, SCD and SDO) and the two EMO approaches (NSGA-II and SPEA2) are not-problem dependant.

The terminal and function sets used in this work are as follows. The terminals are the problem features. The function set contains the most common arithmetic operators, namely ${\cal F} = \{+, -, *, /\}$, where the division operator is protected by returning the numerator when the denominator has a value of zero.
The models evolved by GP map each input pattern in a dataset to a single output value.
When the output of a GP model is greater than, or equal to, zero the pattern is labeled
as part of the minority class, and it is labeled as a majority class pattern, otherwise.

The common way to measure fitness in a classification task is to use the overall classification accuracy: for binary classification, the four possible cases are shown in Table~\ref{tab:outcome:classification}. Assuming the minority class is the positive class, the accuracy is given by  Acc = $\frac{TP+TN}{TP+TN+FP+FN}$, where $TP$ are the true positives,
$TN$ are the true negatives, $FP$ are the false positives and $FN$ are the false negatives. The drawback of using Acc alone is that it rapidly biases the evolutionary search towards the majority class~\cite{6198882}. A better approach is to treat each class 'separately' using a MO approach. Two objectives considered are thus the true positive rate, given by  TPR = $\frac{TP}{TP+FN}$, and the true negative rate given by  TNR = $\frac{TN}{TN+FP}$. They measure the distinct accuracy for the minority (TPR) and majority class (TNR), respectively. 

\begin{table}
\caption{Confusion Matrix}
\centering
\resizebox{0.75\columnwidth}{!}{ 
\begin{tabular}{l|c|c} \hline 
                    & Predicted positive & Predicted negative \\ \hline \hline
  Actual positive     &  True Positive (TP) & False Negative (FN) \\ \hline
  Actual negative & False Positive (FP) & True Negative (TN) \\ \hline
\end{tabular}
}
\label{tab:outcome:classification}
\end{table}

\begin{table}
\centering
\caption{Summary of parameters}
\resizebox{0.65\columnwidth}{!}{ 
\small\begin{tabular}{l|r} \hline 
\emph{Parameter} &
\emph{Value} \\ \hline 
Population Size & 500 \\ 
Generations & 50 \\ 
Type of Crossover & 90\% internal nodes, 10\% leaves  \\ 
Crossover Rate  & 0.60  \\ 
Type of Mutation & Subtree \\ 
Mutation Rate & 0.40 \\ 
Selection & Tournament (size = 7) \\
Initialisation Method & Ramped half-and-half \\ 
Initialisation Depths: & \\ 
\hspace{.3cm}Initial Depth & 1 (Root = 0)\\ 
\hspace{.3cm}Final Depth & 5 \\ 
Maximum Size & 800 nodes \\
Maximum Final Depth & 8\\ 
Independent Runs & 50 \\ 
Semantic Bounds &  UBSS = \{0.25, 0.5, 0.75, 1.0\} \\
 &  LBSS = \{0.001, 0.01, 0.1\} \\ \hline
%\multirow{1}{*}{Changes}  & Every 50 generations \\ \hline 

\end{tabular}
}
\label{tab:parameters}
\end{table}

The experiments were conducted using a generational approach. Tree size was controlled by using a maximum length or a maximum final depth, whatever happens first.  
When a child tree exceeds any of these, the offspring is generated again until these conditions are satisfied. The parameters used  are shown in Table~\ref{tab:parameters}. These include the use of different bounds, defined as Upper Bound Semantic Similarity (UBSS) = \{0.25, 0.50, 0.75, 1.0\} values and Lower Bound Semantic Similarity (LBSS) = \{0.001, 0.01, 0.1\} values,  to compute the semantic distances defined in Eqs.~(\ref{semantic:distance:two:values}) and~(\ref{semantic:distance:one:value}). This results in conducting a thorough analysis: for each of the semantic-based approaches and for each of the datasets used, we have 16\footnote{4 values for UBSS and 4 cases for LBBS: 3 values and 1 case where LBSS is not defined.} independent results, each being the result of 50 independent runs.  To obtain meaningful results, we carried out an  extensive empirical experimentation (29,400  independent runs in total)\footnote{50 independent runs, 6 datasets, 3 semantic-based MOGP approaches (SSC, SCD, SDO), 16 different combination of values for UBSS and LBSS, 2 canonical EMO methods (NSGA-II, SPEA2).}.

The results reported in the following section are based on the testing data set.

\section{Results and Analysis}
\label{sec:results}

%% \begin{table}
%%   \caption{Number of unique solutions found by either NSGA-II \textit{vs.} SDO, SSC or SCD, setting UBSS at 0.5}
%% \centering
%% \resizebox{1.0\columnwidth}{!}{
%% \begin{tabular}{|c|ccc|cc|cc|cc|}\hline

%%   \multirow{1}{*}{Data }     & \multicolumn{7}{c|}{NSGA-II}\\ \cline{2-8}%  & \multicolumn{2}{c|}{NSGA-II}  &   \multicolumn{2}{c|}{NSGA-II}\\
%%                Set                &  Canonical & SDO  & Ratio & Canonical  & SSC  & Canonical  & SCD  \\
%%  \hline

%% Ion & 162.5  $\pm$  20.1   &751.1 $\pm$ 187.9  & $\approx$4.6& 269.7  $\pm$  11.2 & 252.8 $\pm$ 20.5 & 270.0  $\pm$  25.8 &  245.4 $\pm$ 27.5   \\

%%  Spect  &   81.1  $\pm$  13.9 &  331.7 $\pm$ 56.7 & $\approx$4.1&151.8  $\pm$  9.5   & 155.8 $\pm$ 14.3 &156.7  $\pm$  9.1  & 159.5 $\pm$ 24.0\\

%% Yeast$_1$ &  798.5  $\pm$  33.3 &  1991.5 $\pm$ 149.0 & $\approx$2.4  &  1166.5  $\pm$  26.1  & 1050.4 $\pm$ 30.6 & 1168.7  $\pm$  19.4 &  1058.0 $\pm$ 44.5 \\

%% Yeast$_2$ &  235.9  $\pm$  15.53 &  855.1 $\pm$ 138.5 & $\approx$3.6 &   418.8  $\pm$  13.3 & 414.9 $\pm$ 16.9 & 418.0  $\pm$  9.9 &  398.1 $\pm$ 16.7   \\

%% Abal$_1$ & 128.6  $\pm$  13.2 & 485.8 $\pm$ 55.3 & $\approx$3.7 &  192.6  $\pm$  5.9 &    205.1 $\pm$ 9.1 & 193.4  $\pm$  6.0 &204.5 $\pm$ 6.2   \\

%% Abal$_2$ & 183.5  $\pm$  21.3 &  1034.7 $\pm$ 284.8&  $\approx$5.6 &   248.9  $\pm$  2.4 &  187.9 $\pm$ 11.8& 244.4  $\pm$  4.2 &  215.7 $\pm$ 13.2 \\

%% \hline
%% \end{tabular}
%% }
%% \label{tab:unique:nsgaii}
%% \end{table}

\begin{table}
  \caption{Number of unique solutions (average and standard deviation) found by NSGA-II \textit{vs.} SDO, SSC or SCD.}
\centering
\resizebox{1.0\columnwidth}{!}{
\begin{tabular}{c|cc|cc|cc}\hline

  \multirow{1}{*}{Data}     & \multicolumn{6}{c}{NSGA-II}\\ \cline{2-7} %\multicolumn{2}{c|}{NSGA-II}  &   \multicolumn{2}{c|}{NSGA-II}\\
               Set                &  Canonical & SDO   & Canonical  & SSC  & Canonical  & SCD  \\
 \hline

Ion & 162.5  $\pm$  20.1   &751.1 $\pm$ 187.9 & 269.7  $\pm$  11.2 & 252.8 $\pm$ 20.5 & 270.0  $\pm$  25.8 &  245.4 $\pm$ 27.5   \\

 Spect  &   81.1  $\pm$  13.9 &  331.7 $\pm$ 56.7 &151.8  $\pm$  9.5   & 155.8 $\pm$ 14.3 &156.7  $\pm$  9.1  & 159.5 $\pm$ 24.0\\

Yeast$_1$ &  798.5  $\pm$  33.3 &  1991.5 $\pm$ 149.0  &  1166.5  $\pm$  26.1  & 1050.4 $\pm$ 30.6 & 1168.7  $\pm$  19.4 &  1058.0 $\pm$ 44.5 \\

Yeast$_2$ &  235.9  $\pm$  15.53 &  855.1 $\pm$ 138.5 &   418.8  $\pm$  13.3 & 414.9 $\pm$ 16.9 & 418.0  $\pm$  9.9 &  398.1 $\pm$ 16.7   \\

Abal$_1$ & 128.6  $\pm$  13.2 & 485.8 $\pm$ 55.3 &  192.6  $\pm$  5.9 &    205.1 $\pm$ 9.1 & 193.4  $\pm$  6.0 &204.5 $\pm$ 6.2   \\

Abal$_2$ & 183.5  $\pm$  21.3 &  1034.7 $\pm$ 284.8 &   248.9  $\pm$  2.4 &  187.9 $\pm$ 11.8& 244.4  $\pm$  4.2 &  215.7 $\pm$ 13.2 \\

\hline
\end{tabular}
}
\label{tab:unique:nsgaii}
\end{table}

\begin{table}
  \caption{Number of unique solutions (average and standard deviation) found by SPEA2 \textit{vs.} SDO, SSC or SCD.}
\centering
\resizebox{1.0\columnwidth}{!}{
\begin{tabular}{c|cc|cc|cc}\hline

  \multirow{1}{*}{Data}     & \multicolumn{6}{c}{SPEA2}\\ \cline{2-7}%  & \multicolumn{2}{c|}{NSGA-II}  &   \multicolumn{2}{c|}{NSGA-II}\\
             Set                  &  Canonical & SDO & Canonical  & SSC & Canonical  & SCD \\
 \hline

Ion &  147.8  $\pm$  15.7 & 775.5 $\pm$ 161.9  &247.2  $\pm$  9.1 &  82.6 $\pm$ 14.0&  242.1  $\pm$  6.6  &  273.9 $\pm$ 18.3 \\

 Spect  &  95.5  $\pm$  18.1 & 334.1 $\pm$ 45.2  & 169.8  $\pm$  16.0  &  168.8 $\pm$ 16.5 &  183.5  $\pm$  12.5 & 147.8 $\pm$ 13.4 \\

Yeast$_1$ & 749.1  $\pm$  40.7 & 2076.6 $\pm$ 189.5   &1144.0  $\pm$  22.8  & 1067.5 $\pm$ 36.9 &  1131.9  $\pm$  35.6 & 1087.6 $\pm$ 44.4  \\

Yeast$_2$ & 253.1  $\pm$  21.4 & 886.3 $\pm$ 153.1  &  439.1875  $\pm$  14.3 &  399.8125 $\pm$ 10.1 &  430.25  $\pm$  8.6&  411.5 $\pm$ 26. \\

Abal$_1$ &  149.3  $\pm$  12.7 &  498.1 $\pm$ 70.9  & 213.7  $\pm$  6.6 &  198.8 $\pm$ 7.7 & 213.7  $\pm$  8.1  & 205.0 $\pm$ 11.6  \\

Abal$_2$ & 166.1  $\pm$  13.9 & 1120.4 $\pm$ 233.3  &218.8  $\pm$  2.5  & 198.1 $\pm$ 14.7 & 216.6  $\pm$  3.8 & 230.5 $\pm$ 14.3  \\

\hline
\end{tabular}
}
\label{tab:unique:spea2}
\end{table}

\subsection{Diversity}
Diversity in GP can be quantified in many different ways, in this work we focus on
phenotypic diversity, based on the number of unique fitness values \cite{div1,div2}.
This measure of diversity is particularly relevant in a MO problem, since the spread
of solutions in the Pareto front is a desired feature of the optimisation process.
Specifically, we quantify diversity  by computing the number of unique solutions 
produced in objective space, considering the TPR and the TNR (see Section~\ref{sec:experimental}) of each solution
found by either a canonical EMO approach (NSGA-II or SPEA2) \textit{vs.} a semantic-based approach (SDO, SSC or SCD) for each of the six datasets used in this work (see Table~\ref{tab:datasets}). A unique solution is defined  as a solution that was obtained by a particular approach (for instance NSGA-II) but not obtained by another approach (for instance SDO). Consequently, a non-unique solution is one that was obtained by two approaches.

We know from Section~\ref{sec:experimental} that we used 16 different configurations of values to compute the semantic distance for each of the semantic-based approaches in each of the datasets used. We summarised these by computing the average and standard deviation of these 16 values and compared it against a canonical MO approach. Table~\ref{tab:unique:nsgaii} reports the number of unique solutions found by either NSGA-II or SDO, NSGA-II or SSC and NSGA-II or SCD.  As can be seen from this table, our proposed SDO approach, which treats semantic distance values as an additional criterion, third column from left to right of Table~\ref{tab:unique:nsgaii}, produces more unique solutions compared to canonical NSGA-II for any of the datasets used. For example, the lowest ratio of SDO \textit{vs.} NSGA-II is $\approx$2.4 on the Yeast$_1$ dataset. That is, there are 2.4 more solutions produced by SDO compared to NSGA-II (1991.5 $\pm$ 149.0 \textit{vs.} 798.5 $\pm$ 33.3). At the opposite end, we have that the highest ratio of SDO \textit{vs.} NSGA-II is $\approx$5.6 on the Abal$_2$ dataset (183.5  $\pm$  21.3 \textit{vs.}  1034.7 $\pm$ 284.8). When we then turn our attention to the other two semantic-based approaches, we can see that there is no real advantage for SSC or SCD over NSGA-II, in terms of producing more solutions on the two conflicting classes. 

Table~\ref{tab:unique:spea2} reports the number of unique solutions found by either SPEA2 \textit{vs.} any of the semantic-based approaches used in this work. Similarly to what has been reported in Table~\ref{tab:unique:nsgaii}, as well as discussed before, we can observe that our proposed SDO produces significantly more solutions compared to canonical SPEA2. For example, the lowest ratio of SDO \textit{vs.} SPEA2 is $\approx$2.7 on the Yeast$_2$ dataset (253.1  $\pm$  21.4 \textit{vs.} 886.3 $\pm$ 153.1) and the highest ratio of SDO \textit{vs.} SPEA2 is $\approx$6.7 on the Abal$_2$ dataset (166.1  $\pm$  13.9 \textit{vs.} 1120.4 $\pm$ 233.3). Similarly to what can be seen in Table~\ref{tab:unique:nsgaii}, the other two semantic-based approaches (SSC and SCD) do not show a significant improvement when compared to SPEA2.

\begin{figure*}
  \centering
%  \resizebox{0.75\textwidth}{!}{
  \begin{tabular}{ccc}
    \scriptsize{Ion} & \scriptsize{Spect} &  \scriptsize{Yeast$_1$} \\
    \hspace{-0.42cm}  \includegraphics[width=0.360\textwidth]{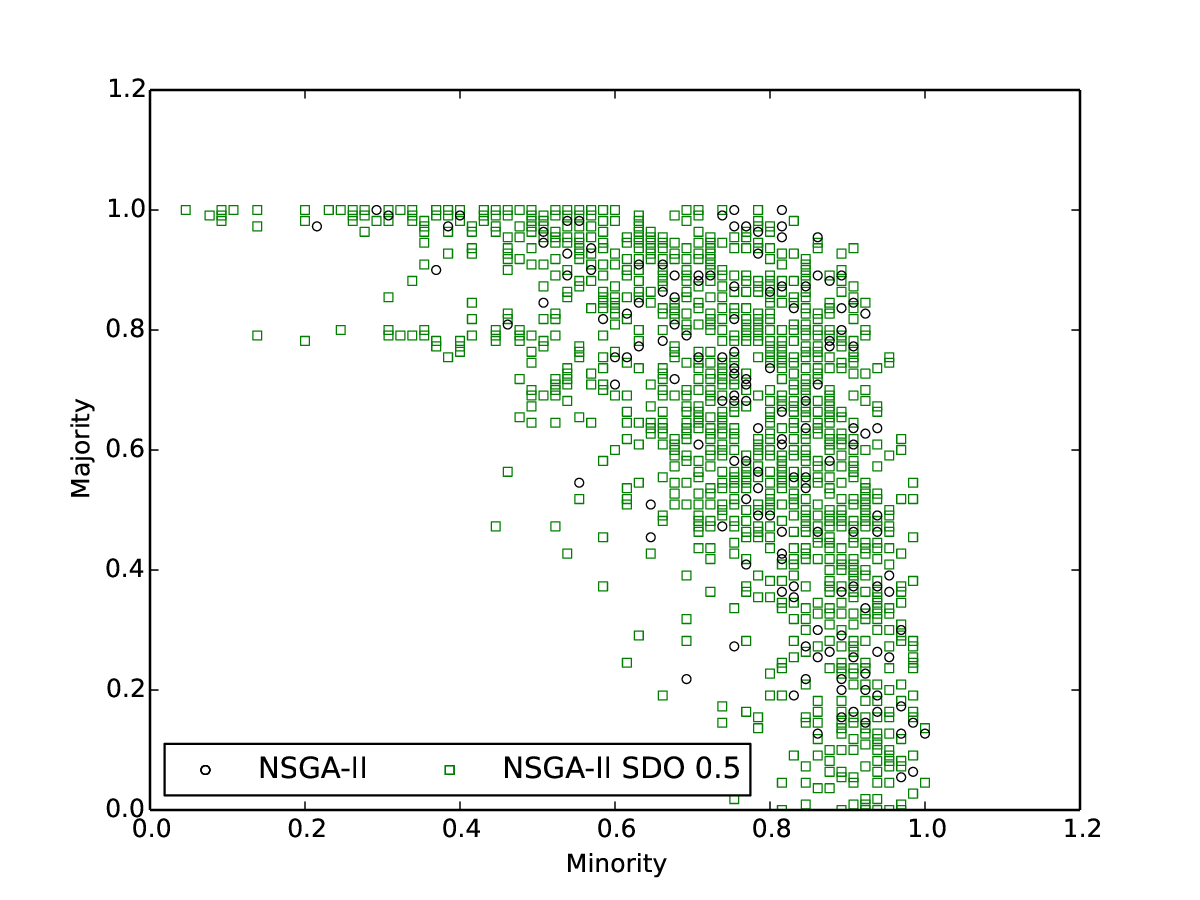}  & \hspace{-0.95cm}   \includegraphics[width=0.360\textwidth]{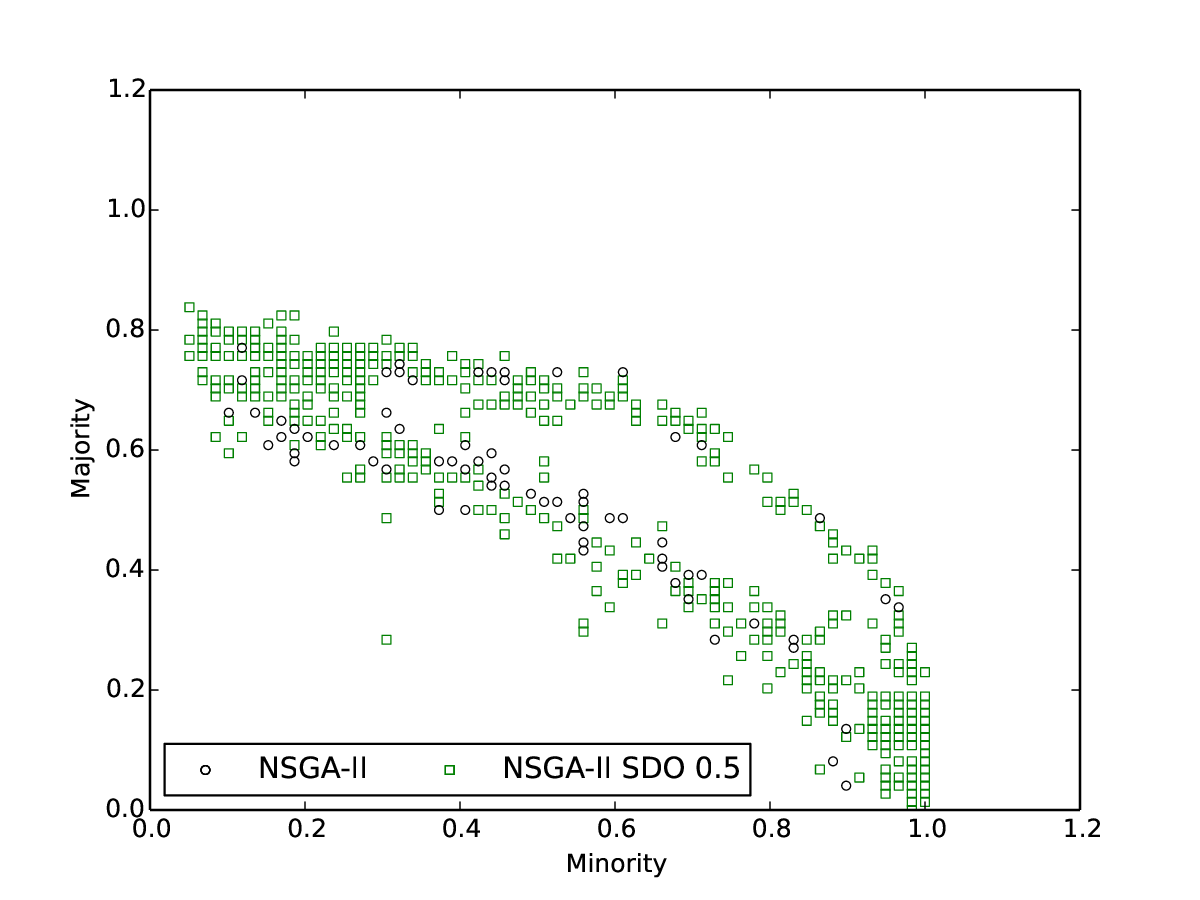} & \hspace{-0.95cm}  \includegraphics[width=0.36\textwidth]{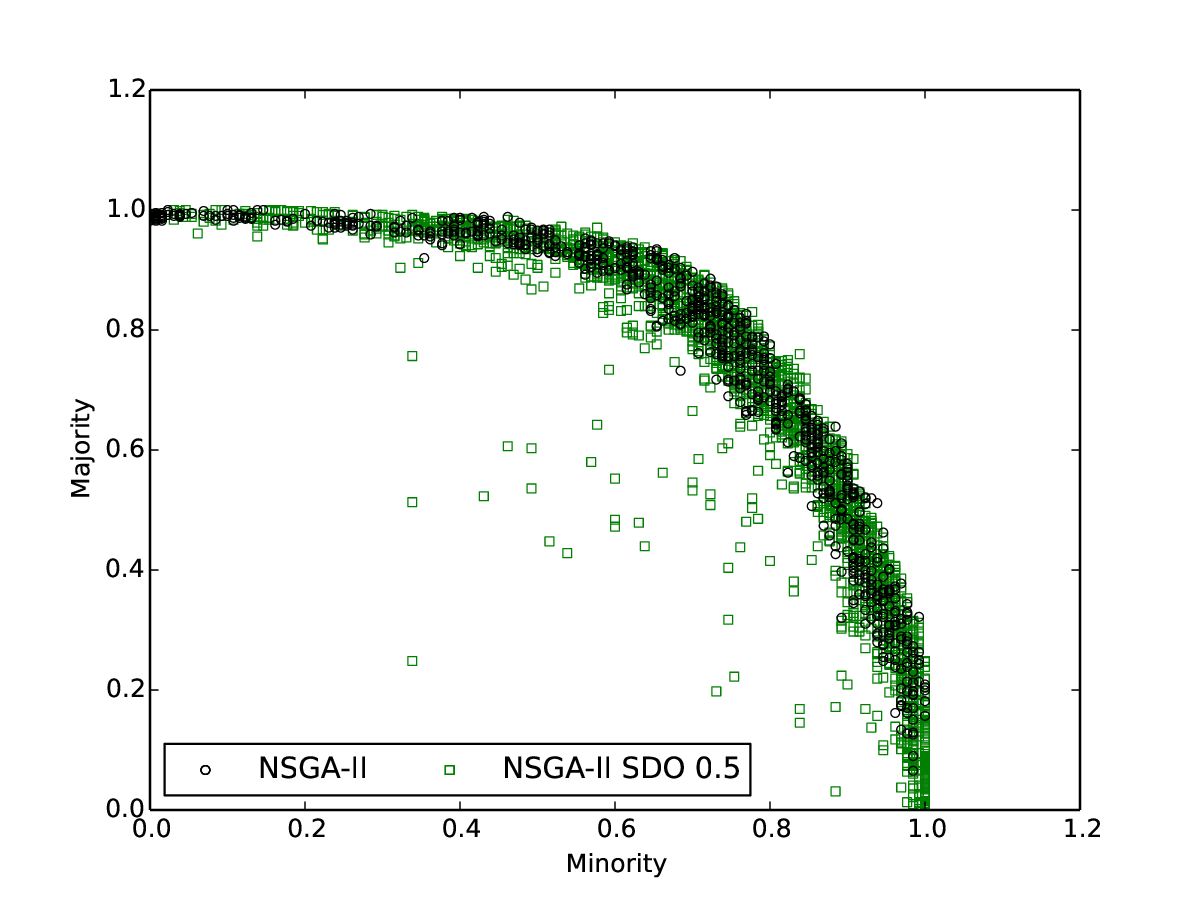} \\
    
    \scriptsize{Yeast$_2$} & \scriptsize{Abal$_1$} &  \scriptsize{Abal$_2$}\\
    
    \hspace{-0.42cm}   \includegraphics[width=0.360\textwidth]{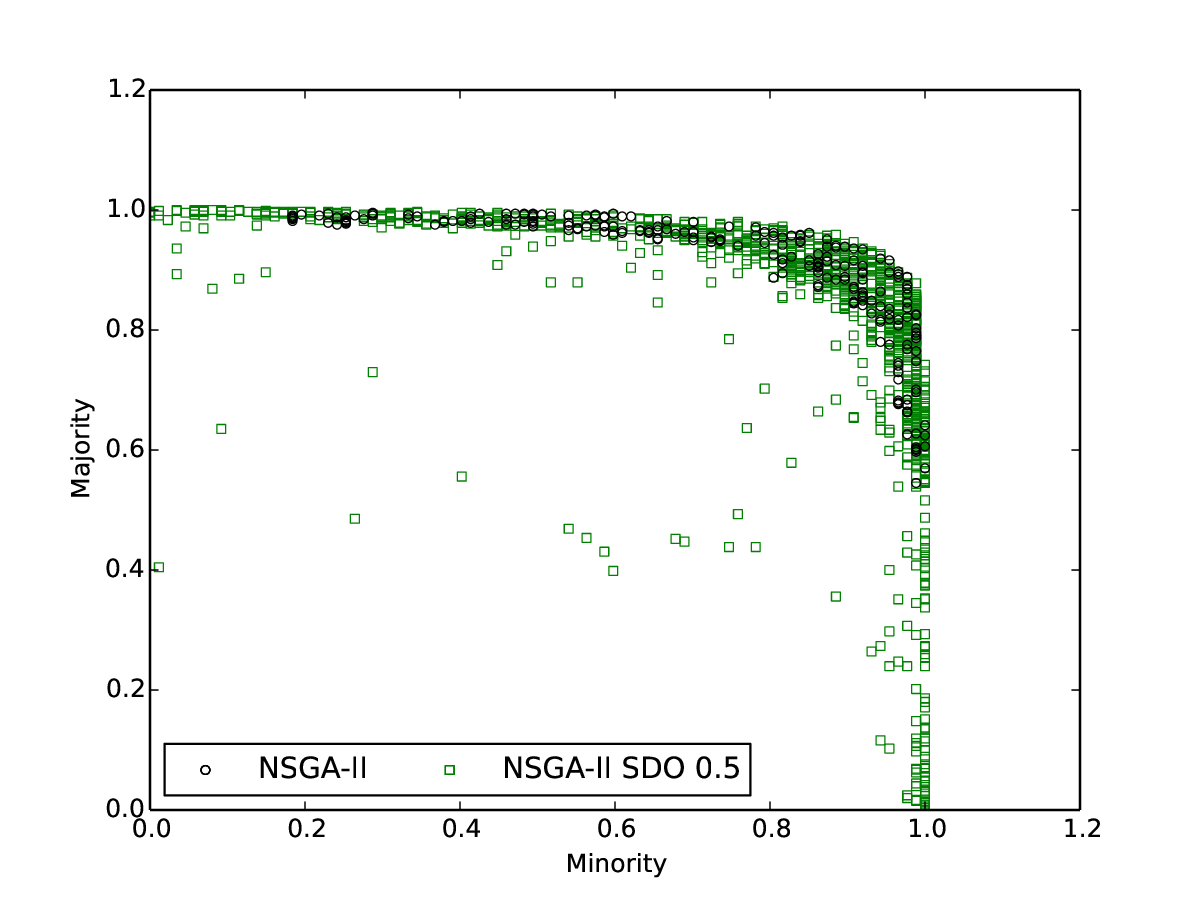} & \hspace{-0.95cm}  \includegraphics[width=0.360\textwidth]{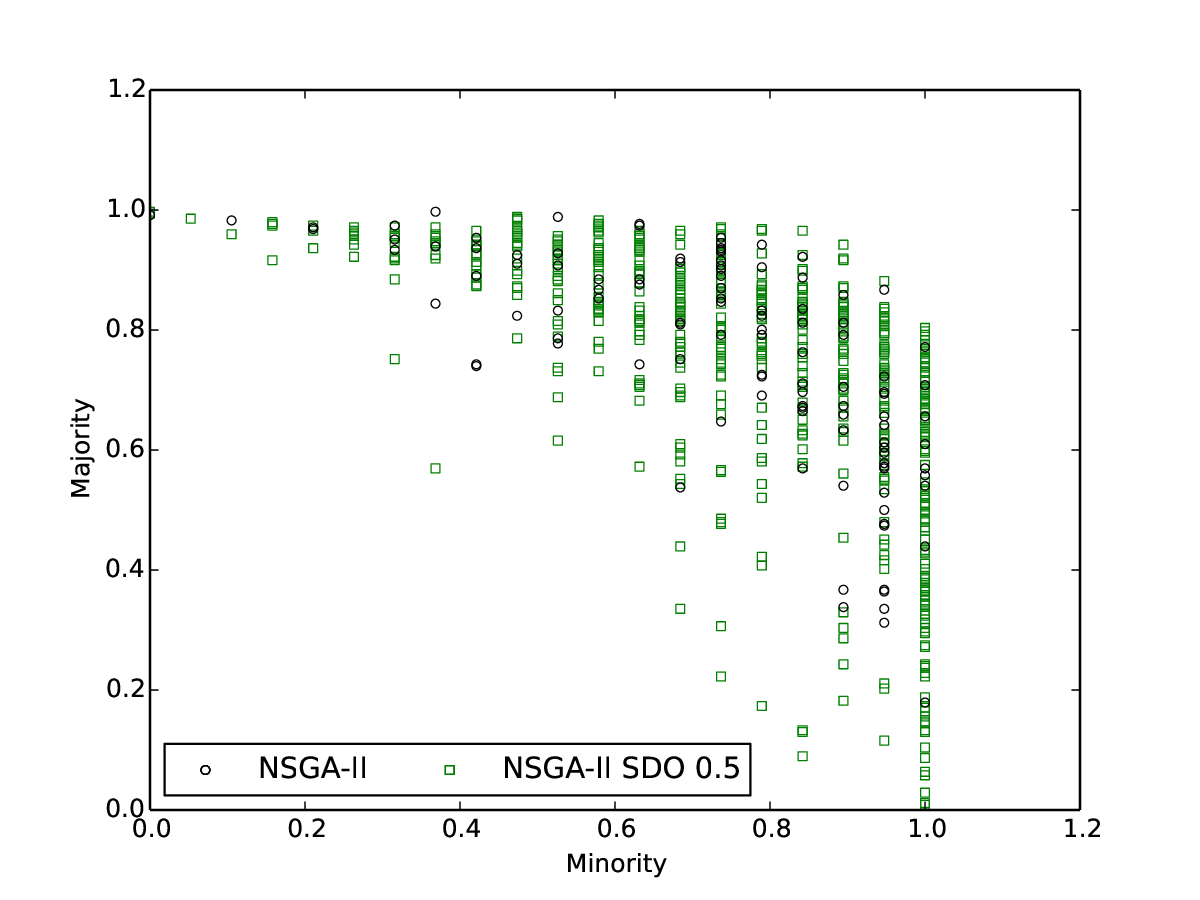}   & \hspace{-0.95cm}   \includegraphics[width=0.360\textwidth]{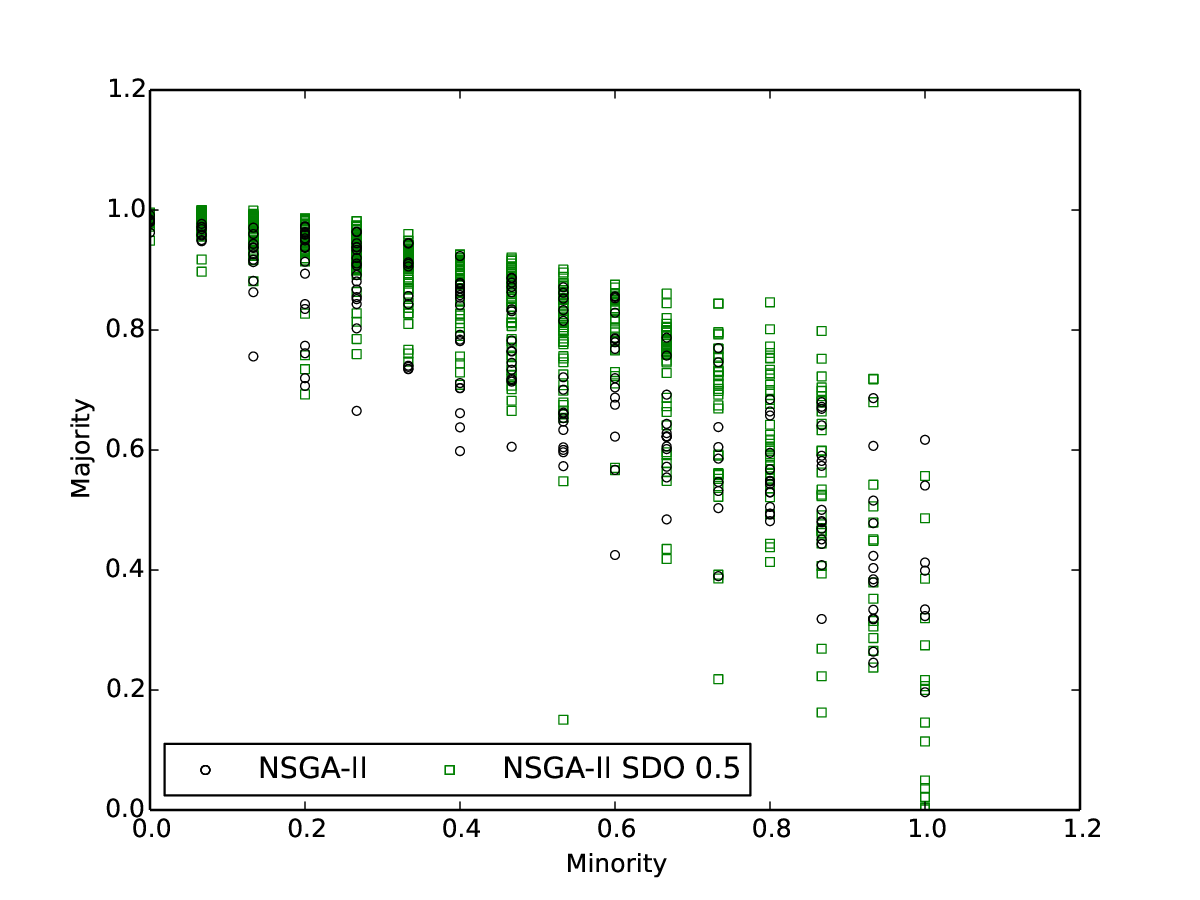} \\

  \end{tabular}
 % }
\caption{Evolved solutions that were exclusively found  by either \underline{NSGA-II}, represented by black hollow circle symbols, and its SDO variant, represented by green hollow square symbols, setting UBSS = 0.5, for all the datasets used in this work.}%the Ion, Spect, Abal$_1$ and Abal$_2$ datasets.}

\label{fig:evolved:solutions_nsgaii}
\end{figure*}

\begin{figure*}
  \centering
%  \resizebox{0.75\textwidth}{!}{
  \begin{tabular}{ccc}
    \scriptsize{Ion} & \scriptsize{Spect} &  \scriptsize{Yeast$_1$} \\
    \hspace{-0.42cm}  \includegraphics[width=0.360\textwidth]{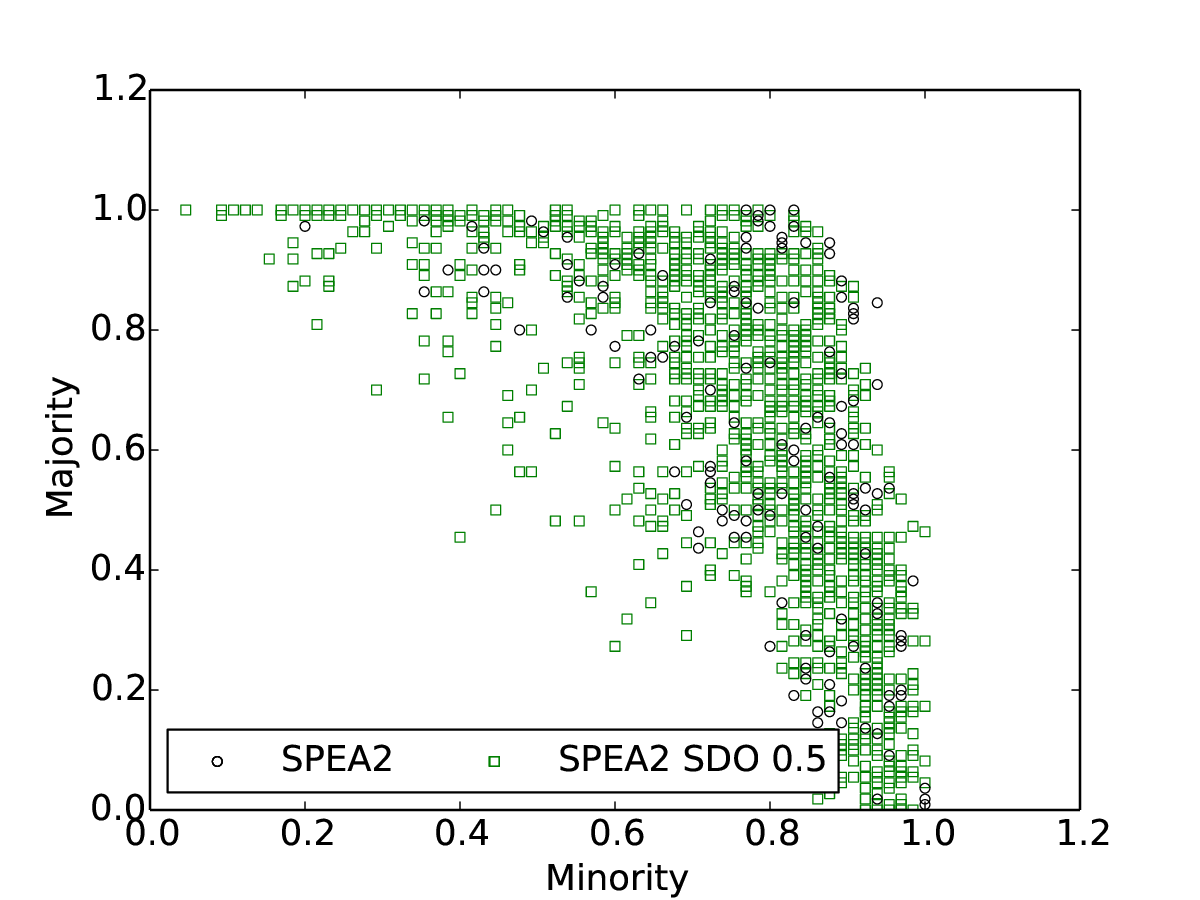}  & \hspace{-0.95cm}   \includegraphics[width=0.360\textwidth]{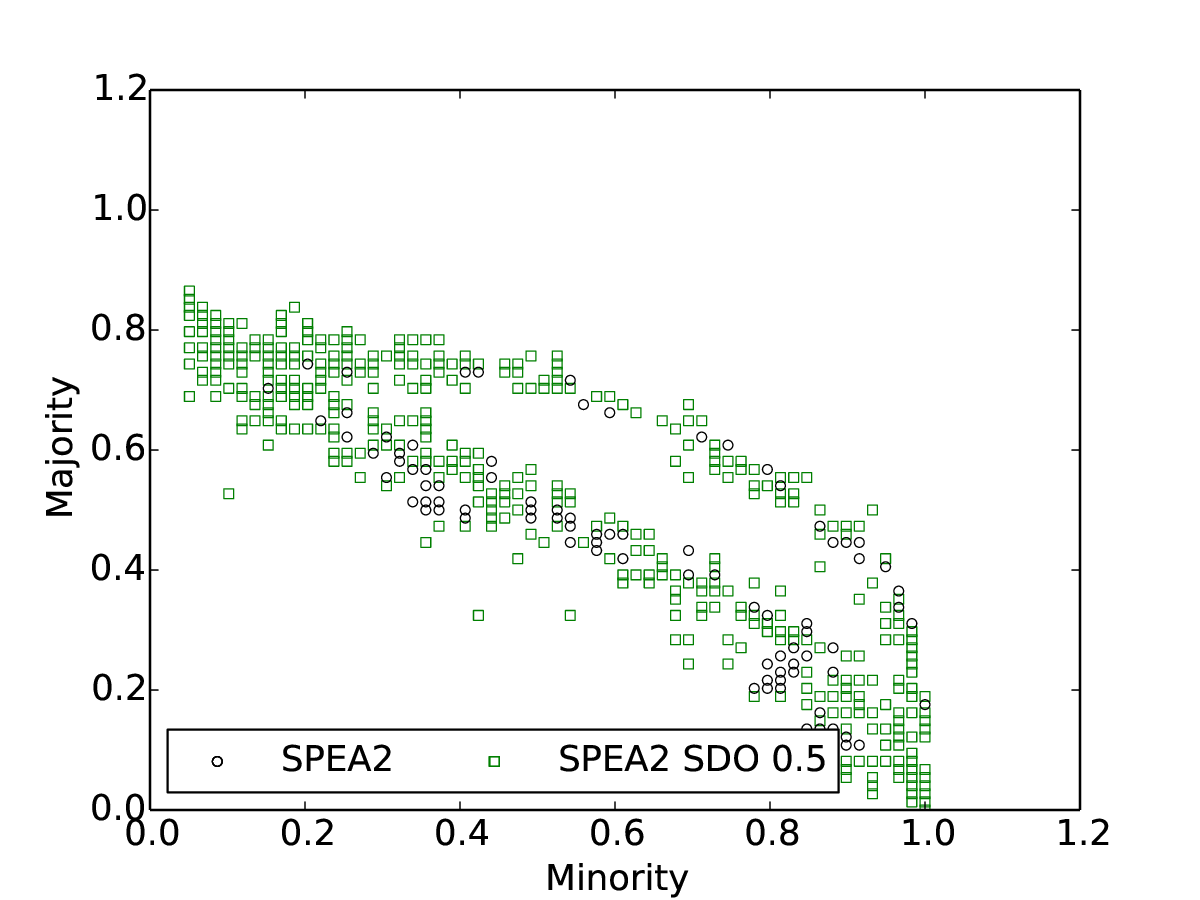} & \hspace{-0.95cm}  \includegraphics[width=0.36\textwidth]{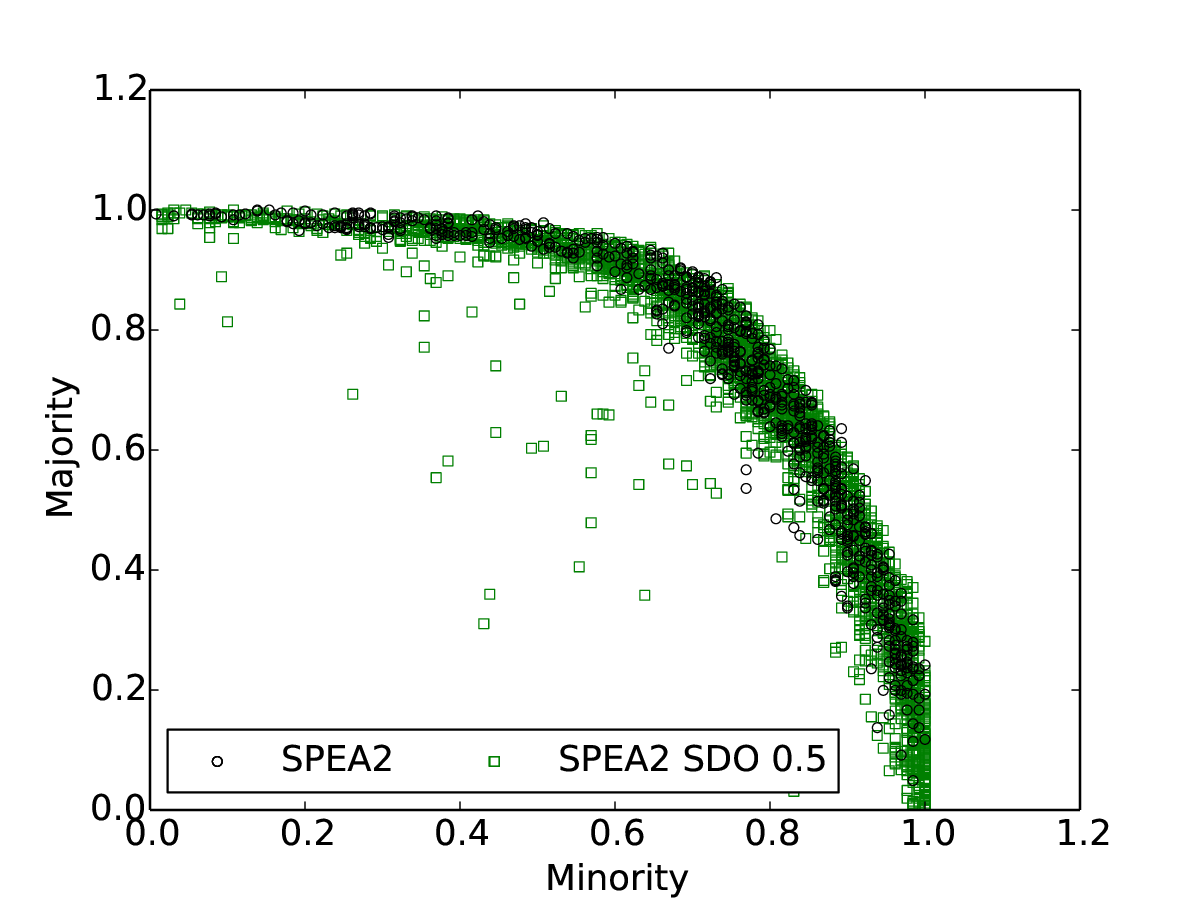} \\
    
    \scriptsize{Yeast$_2$} & \scriptsize{Abal$_1$} &  \scriptsize{Abal$_2$}\\
    
    \hspace{-0.42cm}   \includegraphics[width=0.360\textwidth]{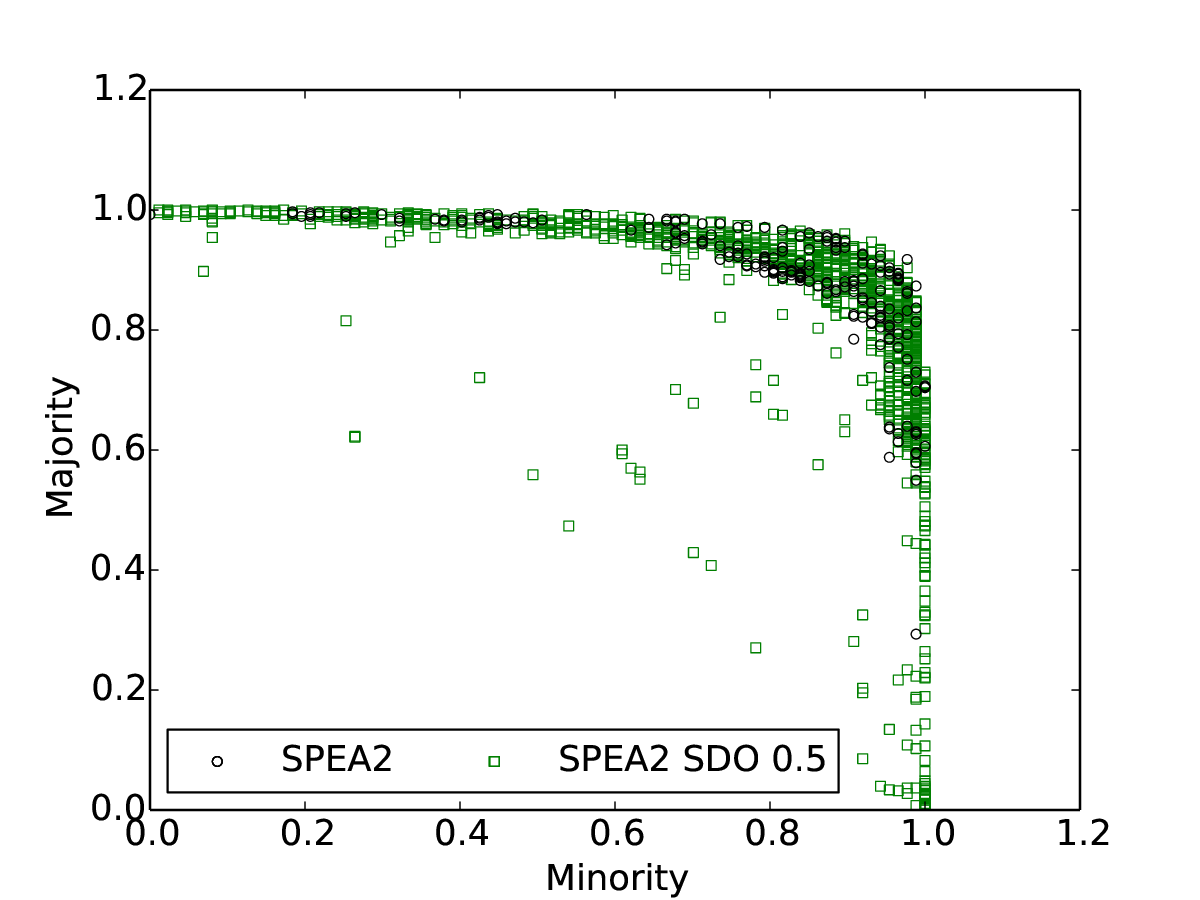} & \hspace{-0.95cm}  \includegraphics[width=0.360\textwidth]{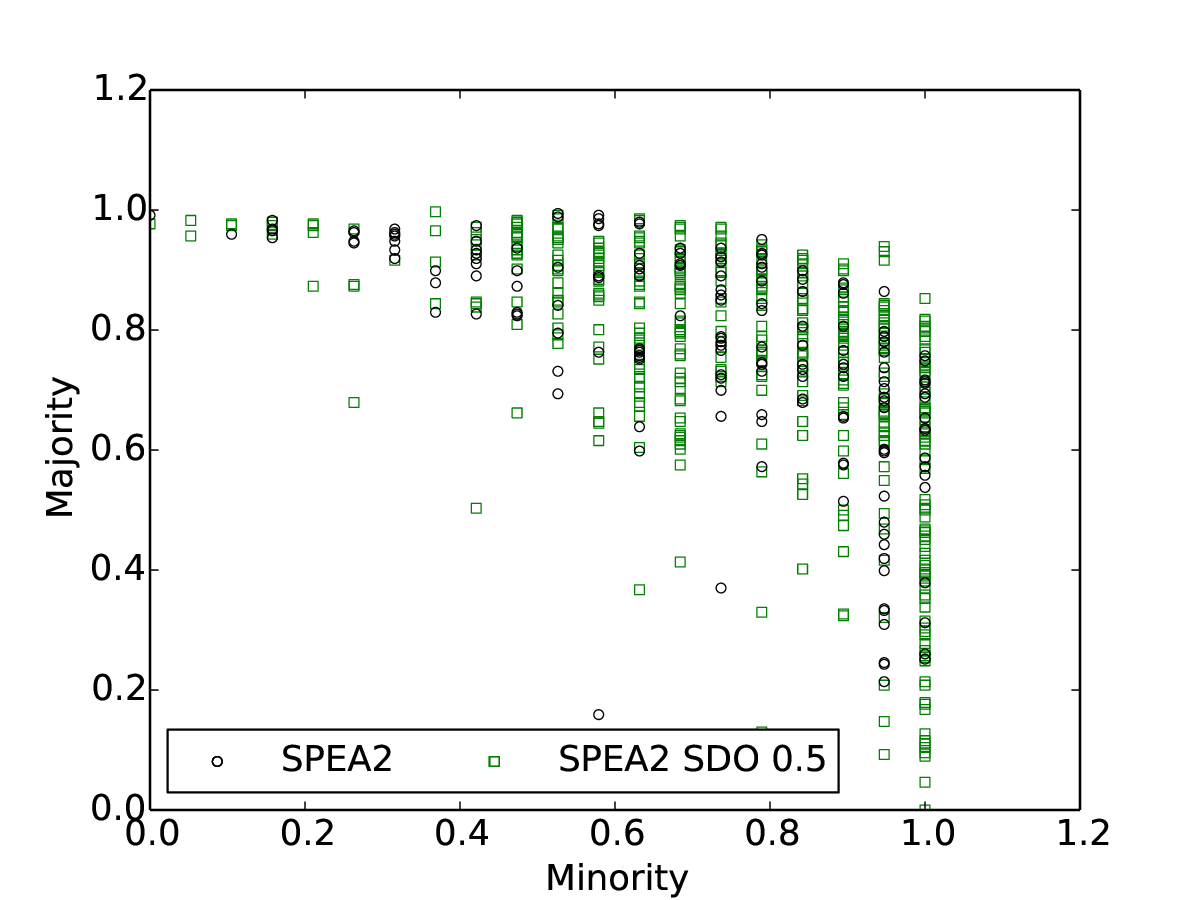}   & \hspace{-0.95cm}   \includegraphics[width=0.360\textwidth]{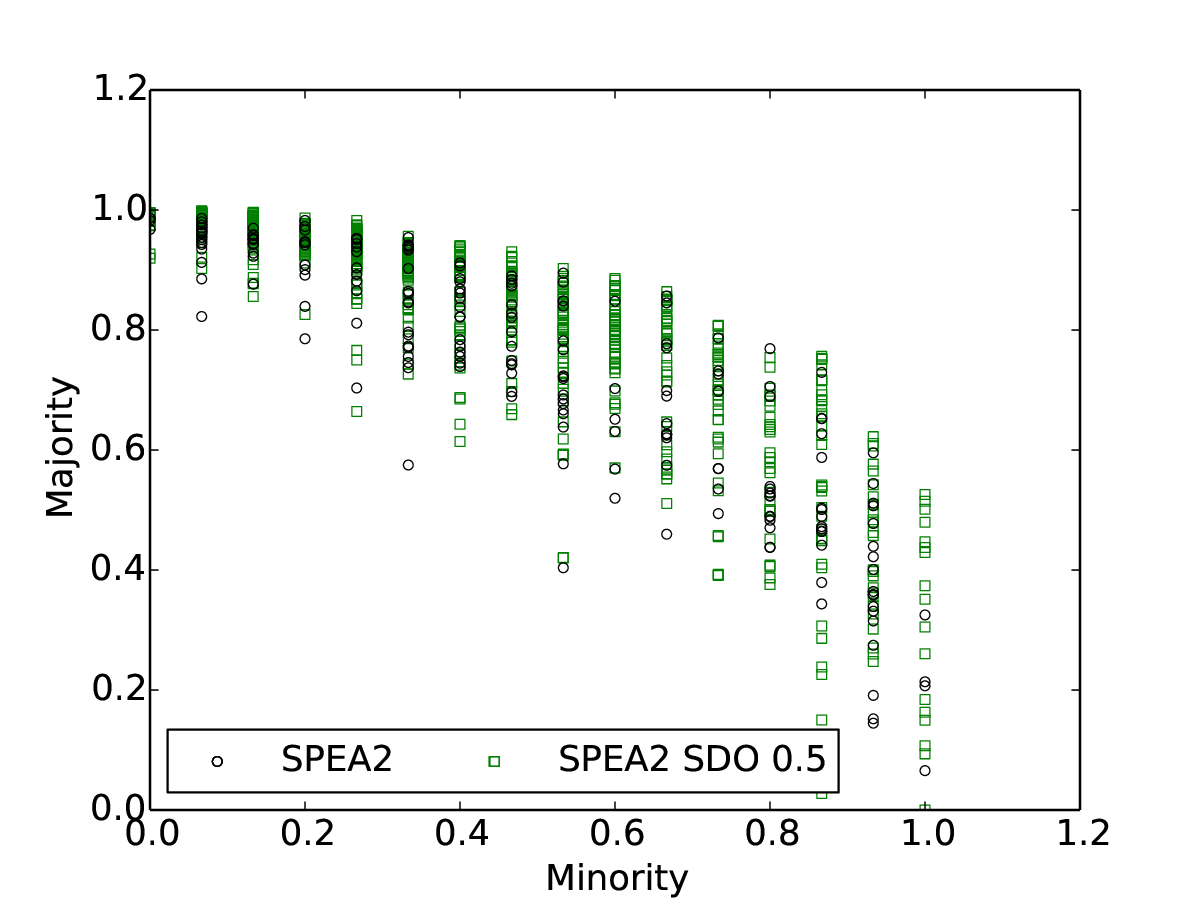} \\

  \end{tabular}
 % }
\caption{Evolved solutions that were exclusively found  by either \underline{SPEA2}, represented by black hollow circle symbols, and its SDO variant, represented by green hollow square symbols, setting UBSS = 0.5, for all the datasets used in this work.}%the Ion, Spect, Abal$_1$ and Abal$_2$ datasets.}

\label{fig:evolved:solutions_spea2}
\end{figure*}

From this analysis, it is clear that our proposed SDO produces more unique solutions compared to any of the canonical MO approaches used in this work, regardless of the dataset used. However, we cannot say whether these unique solutions found by our approach are better or worse compared to those obtained by NSGA-II or SPEA2. To address this, we provide further analysis of the approximated Pareto fronts in Section \ref{sec:sdo_discuss}.  Figures~\ref{fig:evolved:solutions_nsgaii} and~\ref{fig:evolved:solutions_spea2} show the evolved solutions that were exclusively found by (i) either NSGA-II or NSGA-II SDO, and (ii) either SPEA2 or SPEA2 SDO, respectively, for all the datasets used in this work. It is clear to see from Figures~\ref{fig:evolved:solutions_nsgaii} and~\ref{fig:evolved:solutions_spea2}, that the unique evolved solutions that were found by SDO (setting UBSS = 0.5 to be used in Eq.~(\ref{semantic:distance:one:value})), represented by green hollow square symbols, are more numerous and more spread than those produced by the canonical EMO approaches: NSGA-II (Figure~\ref{fig:evolved:solutions_nsgaii}) or SPEA2 (Figure~\ref{fig:evolved:solutions_spea2}), represented by black hollow circles. This is particularly clear for the Ion, Spect, Yeast$_2$, Abal$_1$ and Abal$_2$ datasets. The situation is less clear for the Yeast$_1$ dataset.

\subsection{Hypervolume Comparison}
\label{sec:frontHypervolume}
To further compare the three semantic-based methods and the two EMO methods employed in our study, we calculate the hypervolume~\cite{Coello:2006:EAS:1215640} of the approximated Pareto fronts, which is a commonly used performance metric for MO techniques. For bi-objective problems such as the unbalanced binary classification problems employed in this work, the hypervolume is computed as the sum of all the trapezoidal areas fitted under each point in objective space, where the set of points represent the candidate solutions. Moreover, the hypervolume explicitly captures the underlying goals of the MOGP, to concurrently
maximise the accuracies of both the majority class and the minority class. Additionally, the accumulated Pareto-optimal (PO) front, which is the set of non-dominated solutions after merging the Pareto-approximated fronts for all the independent runs, set at 50 in this study, is also computed.

\begin{table}
\caption{Average ($\pm$ standard deviation) hypervolume of evolved Pareto-approximated fronts and PO fronts  for NSGA-II and SPEA2 over 50 independent runs. No significant differences were found in any single dataset}
\centering
\resizebox{0.75\columnwidth}{!}{
\begin{tabular}{ccccc}\hline
\multirow{3}{*}{Dataset} & \multicolumn{2}{c}{NSGA-II}        & \multicolumn{2}{c}{SPEA2} \\
           & \multicolumn{2}{c}{Hypervolume}       & \multicolumn{2}{c}{Hypervolume} \\
    &  Average & PO Front     & Average & PO Front     \\ \hline

Ion &  0.766 $\pm$ 0.114 & 0.938 &  0.786 $\pm$ 0.094  & 0.948 \\

 Spect &  0.534 $\pm$ 0.024  & 0.647 & 0.544 $\pm$ 0.032 & 0.659  \\

Yeast$_1$ & 0.838 $\pm$ 0.011  & 0.876  & 0.838 $\pm$ 0.008 & 0.877 \\

 Yeast$_2$ & 0.950 $\pm$ 0.009  & 0.976  & 0.946 $\pm$ 0.015 &  0.978 \\

 Abal$_1$ & 0.847 $\pm$ 0.058 &  0.961 &  0.832 $\pm$ 0.078 &  0.960 \\

 Abal$_2$ &0.576 $\pm$ 0.122 & 0.842  & 0.544 $\pm$ 0.147 & 0.834 \\
\hline
\end{tabular}
}
\label{tab:hyperarea:nsgaii:spea2}
\end{table}

Tables~\ref{tab:hyperarea:nsgaii:spea2},~\ref{results:nsgaii:semantics} and~\ref{results:spea2:semantics} show the average hypervolume across 50 runs and the hypervolume of the accumulated PO front of these runs for each of the problems shown in Table~\ref{tab:datasets}. In Table~\ref{tab:hyperarea:nsgaii:spea2}, we show these hypervolume values for the two EMO approaches used in this work: NSGA-II and SPEA2. In the same manner, Tables~\ref{results:nsgaii:semantics} and~\ref{results:spea2:semantics} show these hypervolume values yield by the three semantic-based methods using NSGA-II and SPEA2, respectively. Furthermore, for Tables ~\ref{results:nsgaii:semantics} and~\ref{results:spea2:semantics}, these hypervolume values are given for each
of the 16 different configurations of each semantic methods (based on the upper and lower bounds).
Bold text in the tables indicates an average performance that is superior than the average performance of the baseline method (NSGA-II or SPEA2).
The results suggest that, on average, our proposed method SDO (upper part of Tables~\ref{results:nsgaii:semantics} and~\ref{results:spea2:semantics}) consistently outperforms the baselines, while this is not clearly
the case for SSC and SCD (middle and bottom of these tables, respectively).

\begin{table*}
  \caption{Average ($\pm$ standard deviation) hypervolume of evolved Pareto-approximated fronts and PO fronts for the NSGA-II semantic-based methods (SDO, SSC, SCD) over 50 independent runs.
  Bold indicates better performance compared to the baseline NSGA-II results reported in Table \ref{tab:hyperarea:nsgaii:spea2}.}
 
  \centering
   \resizebox{0.85\textwidth}{!}{
  \begin{tabular}{cccccccccc}\hline

                             &                            & \multicolumn{8}{c}{Hypervolume}  \\
    &                            & \multicolumn{4}{c}{Average} & \multicolumn{4}{c}{PO Front}  \\
     &          & \multicolumn{4}{c}{UBSS} & \multicolumn{4}{c}{UBSS} \\ 
    &    LBSS              & 0.25 & 0.5& 0.75 & 1.0 & 0.25 & 0.5& 0.75 & 1.0 \\ \hline

    \multicolumn{10}{c}{\textsf{NSGA-II SDO}} \\ \hline
    \multirow{4}{*}{Ion} & -- & \textbf{0.860 $\pm$ 0.033} & \textbf{0.869 $\pm$ 0.037}  & \textbf{{0.869 $\pm$ 0.033}} & \textbf{0.845 $\pm$ 0.057} & \textbf{0.948} & \textbf{0.958} & \textbf{0.962} & \textbf{{0.950}}  \\
     & 0.001                & {\textbf{0.817 $\pm$ 0.087}} & \textbf{0.819 $\pm$ 0.104} & \textbf{0.857 $\pm$ 0.057} & \textbf{0.861 $\pm$ 0.047} & \textbf{0.942} & \textbf{0.957} & \textbf{0.954} & \textbf{0.958} \\ 
     & 0.01     & \textbf{0.825 $\pm$ 0.084} & \textbf{0.843 $\pm$ 0.073} & \textbf{0.861 $\pm$ 0.045} & \textbf{0.861 $\pm$ 0.038} & \textbf{0.946} & \textbf{0.956} & \textbf{0.957} & \textbf{0.944} \\
      & 0.1     & \textbf{0.846 $\pm$ 0.070} & \textbf{0.848 $\pm$ 0.068} &\textbf{0.844 $\pm$ 0.075} & \textbf{0.864 $\pm$ 0.044} & \textbf{0.950} & \textbf{0.956} &\textbf{0.953} & \textbf{0.960} \\ \hline 
    
   \multirow{4}{*}{Spect} & -- & \textbf{0.591 $\pm$ 0.027}  & \textbf{0.593 $\pm$ 0.025}  & \textbf{0.594 $\pm$ 0.023} & \textbf{{0.600 $\pm$ 0.019}} &  \textbf{0.684} & \textbf{0.679} & \textbf{0.689} & \textbf{0.694} \\
                       & 0.001 & \textbf{0.562 $\pm$ 0.021} & {\textbf{0.558 $\pm$ 0.025}} & \textbf{0.561 $\pm$ 0.019}  & \textbf{0.560 $\pm$ 0.016} & \textbf{0.668} & \textbf{0.653} & \textbf{0.660} & {0.644} \\ 
                       & 0.01  & \textbf{0.564 $\pm$ 0.025} & \textbf{0.560 $\pm$ 0.023} & \textbf{0.566 $\pm$ 0.024} & \textbf{0.559 $\pm$ 0.016} & \textbf{0.672} & \textbf{0.650} & \textbf{0.669} & {0.643} \\ 
                       & 0.1  & \textbf{0.563 $\pm$ 0.022} & \textbf{0.563 $\pm$ 0.024} & \textbf{0.567 $\pm$ 0.018} & \textbf{0.561 $\pm$ 0.024} & \textbf{0.664} & \textbf{0.658} & \textbf{0.655} & \textbf{0.658}  \\  \hline

  \multirow{4}{*}{Yeast$_1$} & -- & \textbf{0.850 $\pm$ 0.006} & \textbf{0.849 $\pm$ 0.008} & \textbf{0.849 $\pm$ 0.006} & \textbf{0.849 $\pm$ 0.006} & \textbf{0.881} & \textbf{0.881} & \textbf{0.882} & \textbf{0.881} \\ 
   & 0.001 & {\textbf{0.845 $\pm$ 0.007}} & \textbf{0.847 $\pm$ 0.006} & \textbf{0.848 $\pm$ 0.004} & \textbf{0.848 $\pm$ 0.005} & \textbf{0.879} & \textbf{0.882} & \textbf{0.879} & \textbf{0.880} \\ 
     & 0.01     & \textbf{0.848 $\pm$ 0.006}  & \textbf{0.849 $\pm$ 0.005} & \textbf{0.848 $\pm$ 0.005} & \textbf{{0.850 $\pm$ 0.005}} & \textbf{0.881} & \textbf{0.881} & \textbf{0.879} & \textbf{0.881} \\ 
      & 0.1     & \textbf{0.847 $\pm$ 0.005} & \textbf{0.848 $\pm$ 0.005} & \textbf{0.848 $\pm$ 0.005} & \textbf{0.850 $\pm$ 0.005}& \textbf{0.878} & \textbf{0.879} & \textbf{0.879} & \textbf{0.883} \\ \hline

    \multirow{4}{*}{Yeast$_2$} & -- & \textbf{0.961 $\pm$ 0.007}  & \textbf{0.961 $\pm$ 0.007} & \textbf{0.960 $\pm$ 0.008} & \textbf{0.962 $\pm$ 0.007}& \textbf{0.978} & \textbf{0.979} & \textbf{0.979} & \textbf{0.979} \\
   & 0.001 & \textbf{0.959 $\pm$ 0.008} & \textbf{0.958 $\pm$ 0.007} & \textbf{0.961 $\pm$ 0.006} & \textbf{0.961 $\pm$ 0.006} & \textbf{0.981} & \textbf{0.978} & \textbf{0.979} & \textbf{0.978} \\
     & 0.01     & {\textbf{0.955 $\pm$ 0.009}} & \textbf{0.959 $\pm$ 0.007} & \textbf{0.960 $\pm$ 0.009} & \textbf{0.961 $\pm$ 0.007} & \textbf{0.979} & \textbf{0.980} & \textbf{0.979} & \textbf{0.978} \\
      & 0.1     & \textbf{0.958 $\pm$ 0.009}   & \textbf{0.960 $\pm$ 0.007} & \textbf{0.961 $\pm$ 0.007} & \textbf{{0.962 $\pm$ 0.006}} & \textbf{0.978} & \textbf{0.978} & \textbf{0.981} & \textbf{0.979} \\ \hline

    \multirow{4}{*}{Abal$_1$} & -- & \textbf{0.849 $\pm$ 0.081}  & \textbf{0.862 $\pm$ 0.087}  & 0.847 $\pm$ 0.089  & \textbf{0.849 $\pm$ 0.085}  & \textbf{0.964}  & \textbf{0.970}  & \textbf{0.966}  & \textbf{0.967} \\
   & 0.001 & \textbf{{0.892 $\pm$ 0.051}} &  \textbf{0.905 $\pm$ 0.036} & \textbf{0.907 $\pm$ 0.036} & \textbf{0.906 $\pm$ 0.034} & \textbf{0.970} & \textbf{0.968} & \textbf{0.969} & \textbf{0.971} \\ 
     & 0.01     & \textbf{0.908 $\pm$ 0.038} & \textbf{0.900 $\pm$ 0.056} & \textbf{{0.919 $\pm$ 0.022}} & \textbf{0.919 $\pm$ 0.026} & \textbf{0.969} & \textbf{0.973} & \textbf{0.970} & \textbf{0.972} \\
      & 0.1     & \textbf{0.910 $\pm$ 0.037} & \textbf{0.911 $\pm$ 0.046}  & \textbf{0.912 $\pm$ 0.049} & \textbf{0.916 $\pm$ 0.031} & \textbf{0.970} & \textbf{0.972} & \textbf{0.969} & \textbf{0.970} \\  \hline

      \multirow{4}{*}{Abal$_2$} & -- & \textbf{0.591 $\pm$ 0.102}   &  \textbf{0.623 $\pm$ 0.138}  & \textbf{0.634 $\pm$ 0.115}   & \textbf{0.617  $\pm$ 0.137}  & \textbf{0.862}   & \textbf{0.878}  & \textbf{0.881}  &\textbf{0.873}  \\
       & 0.001 &   \textbf{0.729 $\pm$ 0.070} & \textbf{0.722 $\pm$ 0.063} & \textbf{{0.709 $\pm$ 0.080}} & \textbf{0.735 $\pm$ 0.074} & \textbf{0.877}  & \textbf{0.870} & \textbf{0.879}  & \textbf{0.885} \\
       & 0.01     & \textbf{0.721 $\pm$ 0.067} & \textbf{0.725 $\pm$ 0.075} & \textbf{0.721 $\pm$ 0.074} & \textbf{0.723 $\pm$ 0.066} & \textbf{0.881}  & \textbf{0.879}  & \textbf{0.884} & \textbf{0.880}\\ 
      & 0.1     &  \textbf{0.724 $\pm$ 0.076}  & \textbf{0.739 $\pm$ 0.065} & \textbf{0.736 $\pm$ 0.063} & \textbf{{0.756 $\pm$ 0.065}} & \textbf{0.888}  & \textbf{0.883} & \textbf{0.886} &  \textbf{0.890}\\ \hline 

      \hline     
 \multicolumn{10}{c}{\textsf{NSGA-II SSC}} \\ \hline

          \multirow{4}{*}{Ion}  & -- & 0.761 $\pm$ 0.108  &  0.749 $\pm$ 0.161  & 0.763 $\pm$ 0.152  & 0.744 $\pm$ 0.137  & \textbf{0.941}  & 0.937  & \textbf{0.951}  & \textbf{0.949} \\ 
      & 0.001 & 0.765 $\pm$ 0.134  & {0.753 $\pm$ 0.124 } & {0.699 $\pm$ 0.188}  & \textbf{0.803 $\pm$ 0.103 } & \textbf{0.954}  & 0.935  & {0.928}  & \textbf{0.946}   \\ 
       & 0.01     &  0.760 $\pm$ 0.125  & {0.751 $\pm$ 0.123}   & 0.710 $\pm$ 0.161  & \textbf{0.802 $\pm$ 0.104}  & \textbf{0.947}  & 0.929  & 0.928   & \textbf{0.947}  \\
        & 0.1     &  \textbf{0.775 $\pm$ 0.095}  &  {0.738 $\pm$ 0.184}  & {0.746 $\pm$ 0.141}  & \textbf{0.778 $\pm$ 0.099}  & \textbf{0.957}  & \textbf{0.951}  & \textbf{0.945}   & 0.936  \\ \hline

\multirow{4}{*}{Spect} & -- & 0.525 $\pm$ 0.025  & 0.532 $\pm$ 0.029  & \textbf{0.537 $\pm$ 0.020}  & \textbf{0.535 $\pm$ 0.029}  & 0.633  & 0.634  & 0.634  & 0.634  \\
      & 0.001 & 0.530 $\pm$ 0.029  & \textbf{0.539 $\pm$ 0.030}  & \textbf{0.542 $\pm$ 0.023}  & \textbf{0.540 $\pm$ 0.025}  & \textbf{0.651}  & 0.635  & 0.638  & \textbf{0.654}  \\
     & 0.01     & 0.535 $\pm$ 0.029  & \textbf{0.537 $\pm$ 0.027}  & \textbf{0.541 $\pm$ 0.027}  & \textbf{0.540 $\pm$ 0.028}   & \textbf{0.655}  & 0.633  & \textbf{0.658}  & \textbf{0.651}  \\
      & 0.1     & 0.532 $\pm$ 0.029  &0.531 $\pm$ 0.026  & 0.534 $\pm$ 0.027  & 0.533 $\pm$ 0.022  & {0.632}  & 0.641  & 0.635  & 0.635   \\ \hline 
                  %& 0.1  & &  & & & & & &  \\ \hline

      \multirow{4}{*}{Yeast$_1$} & -- & 0.819 $\pm$ 0.041   &0.829 $\pm$ 0.023    & 0.835 $\pm$ 0.014  &0.834 $\pm$ 0.017   & {0.874}  & 0.875  &\textbf{0.878}  & \textbf{0.878}  \\ 
      & 0.001 & 0.825 $\pm$ 0.031   &0.834 $\pm$ 0.029  &0.834 $\pm$ 0.019  &0.826 $\pm$ 0.039  & \textbf{0.877}  & \textbf{0.877}  & \textbf{0.877}  & \textbf{0.877}  \\ 
     & 0.01     &  0.827 $\pm$ 0.027  &0.835 $\pm$ 0.016   & 0.836 $\pm$ 0.019  & 0.830 $\pm$ 0.030  & 0.874  & \textbf{0.877}  & \textbf{0.877}  & \textbf{0.879}  \\ 
      & 0.1     &  0.831 $\pm$ 0.027  & 0.828 $\pm$ 0.034  &0.831 $\pm$ 0.028  &0.835 $\pm$ 0.014  & \textbf{0.879}  & 0.876  & 0.876  & 0.875  \\ \hline

  \multirow{4}{*}{Yeast$_2$} & -- & 0.950 $\pm$ 0.013  &0.948 $\pm$ 0.010   & 0.945 $\pm$ 0.032  & 0.947 $\pm$ 0.009  & \textbf{0.978}   & \textbf{0.977}  & \textbf{0.978}  & \textbf{0.977}  \\
 & 0.001 & 0.946 $\pm$ 0.013  &0.944 $\pm$ 0.028  &  0.947 $\pm$ 0.013  & 0.950 $\pm$ 0.011  & {0.976}  & 0.976  & \textbf{0.977}  & \textbf{0.979}  \\
     & 0.01     & 0.947 $\pm$ 0.014  & 0.944 $\pm$ 0.024  & 0.946 $\pm$ 0.015  &0.949 $\pm$ 0.012  & \textbf{0.978}  & \textbf{0.978}  & \textbf{0.978}  & \textbf{0.978}   \\ 
      & 0.1     &0.948 $\pm$ 0.014   &0.948 $\pm$ 0.012  & 0.946 $\pm$ 0.009  & 0.947 $\pm$ 0.016  & \textbf{0.978}  & \textbf{0.978}  & \textbf{0.977}  & \textbf{0.977}  \\ \hline

        \multirow{4}{*}{Abal$_1$} & -- &  0.844 $\pm$ 0.084  & 0.839 $\pm$ 0.083  & 0.834 $\pm$ 0.070  &0.824 $\pm$ 0.099  & \textbf{0.963}   & \textbf{0.967}   & \textbf{0.962}  & \textbf{0.962}  \\
     & 0.001 & \textbf{0.851 $\pm$ 0.062}  & 0.812 $\pm$ 0.086  & 0.845 $\pm$ 0.077  &0.844 $\pm$ 0.079  & \textbf{0.964}  & {0.961}  & 0.959  & \textbf{0.967}  \\ 
     & 0.01     &  \textbf{0.850 $\pm$ 0.076}  & 0.833 $\pm$ 0.091  & 0.829 $\pm$ 0.096  & 0.836 $\pm$ 0.090  & \textbf{0.972}  & 0.957  & 0.959  & \textbf{0.963}  \\
      & 0.1     &  \textbf{0.869 $\pm$ 0.064}  &0.838 $\pm$ 0.083  &0.844 $\pm$ 0.075  & 0.834 $\pm$ 0.084  & \textbf{0.963}  & \textbf{0.965}  & \textbf{0.965}  & \textbf{0.962}  \\ \hline

              \multirow{4}{*}{Abal$_2$} & -- & 0.521 $\pm$ 0.121   & 0.532 $\pm$ 0.103  & 0.529 $\pm$ 0.128  & 0.511 $\pm$ 0.118  & 0.810  & 0.802  & 0.841  & 0.801 \\
      & 0.001 & 0.561 $\pm$ 0.082  & {0.534 $\pm$ 0.102}  & 0.542 $\pm$ 0.104  & {0.502 $\pm$ 0.161}  & 0.823  & {0.865}  & 0.829  & {0.820}  \\
     & 0.01     & 0.494 $\pm$ 0.147 & {0.536 $\pm$ 0.114}  & 0.533 $\pm$ 0.134  & 0.547 $\pm$ 0.123  & \textbf{0.844} & {0.826}  & 0.841  & \textbf{0.850}  \\
      & 0.1     & 0.513 $\pm$ 0.132 & 0.549 $\pm$ 0.120  & 0.514 $\pm$ 0.112 &0.532 $\pm$ 0.131  & {0.806} & 0.820  & {0.785} & 0.831  \\ \hline

\hline
      
\multicolumn{10}{c}{\textsf{NSGA-II SCD }} \\ \hline

    \multirow{4}{*}{Ion} & -- & \textbf{0.788 $\pm$ 0.114}  & \textbf{0.800 $\pm$ 0.109}  & \textbf{0.771 $\pm$ 0.145}   & \textbf{0.791 $\pm$ 0.111}  & \textbf{0.943}  & \textbf{0.946}  &0.932   & \textbf{0.952} \\
     & 0.001                & 0.542 $\pm$ 0.127 & \textbf{0.789 $\pm$ 0.110}  & \textbf{0.789 $\pm$ 0.110}  & 0.766 $\pm$ 0.120  & {0.876} & \textbf{0.955}  & \textbf{0.955}  & \textbf{0.940} \\
     & 0.01                 & \textbf{0.772 $\pm$ 0.119}  & \textbf{0.801 $\pm$ 0.099}  & \textbf{0.801 $\pm$ 0.099}  & 0.756 $\pm$ 0.158  & \textbf{0.946}  &  \textbf{0.941}  & \textbf{0.941}  & \textbf{0.941}  \\
      & 0.1                 & \textbf{0.770 $\pm$ 0.109}   & \textbf{0.771 $\pm$ 0.124}   & \textbf{0.771 $\pm$ 0.124}  & 0.756 $\pm$ 0.097  & \textbf{0.947}  & \textbf{0.951}  & \textbf{0.951}  & \textbf{0.940}  \\ \hline

          \multirow{4}{*}{Spect} & -- &\textbf{0.542 $\pm$ 0.022}  & \textbf{0.537 $\pm$ 0.027}  & \textbf{0.538 $\pm$ 0.037}  & \textbf{0.537 $\pm$ 0.037}  &0.642  & \textbf{0.658}  & \textbf{0.652} & 0.643  \\
      & 0.001 & \textbf{0.539 $\pm$ 0.032}  & \textbf{0.538 $\pm$ 0.026}  & \textbf{0.538 $\pm$ 0.027}  & \textbf{0.535 $\pm$ 0.036}  & \textbf{0.650}  & 0.643  &0.644  & \textbf{0.648}   \\
     & 0.01     & \textbf{0.537 $\pm$ 0.037}  & \textbf{0.538 $\pm$ 0.026}   & \textbf{0.538 $\pm$ 0.027}  & \textbf{0.535 $\pm$ 0.036}  & \textbf{0.650}   & 0.643 &0.644  & \textbf{0.648}  \\
      & 0.1     &  \textbf{0.542 $\pm$ 0.025} & \textbf{0.535 $\pm$ 0.027}  & \textbf{0.536 $\pm$ 0.024}  & \textbf{0.538 $\pm$ 0.024}  &\textbf{0.652}  &0.643  & {0.632} & 0.644  \\
      \hline

  \multirow{4}{*}{Yeast$_1$} & -- & 0.836 $\pm$ 0.014  & 0.826 $\pm$ 0.038  & 0.838 $\pm$ 0.008  & 0.837 $\pm$ 0.011  & \textbf{0.877}  &0.874  &0.876  &\textbf{0.877}  \\
   & 0.001 & 0.834 $\pm$ 0.026  &0.836 $\pm$ 0.015    &0.837 $\pm$ 0.016  &0.834 $\pm$ 0.016  & \textbf{0.877}   & {0.873}   & 0.875  & 0.876  \\ 
     & 0.01     & 0.834 $\pm$ 0.015  & 0.836 $\pm$ 0.011  & 0.834 $\pm$ 0.018  & 0.835 $\pm$ 0.015  &0.876  & \textbf{0.877}  & 0.875 & \textbf{0.878}  \\
      & 0.1     & 0.836 $\pm$ 0.014  & 0.826 $\pm$ 0.038  &0.838 $\pm$ 0.008  & 0.837 $\pm$ 0.011  & \textbf{0.879}  & 0.875  & \textbf{0.877}  &  0.876  \\ \hline

        \multirow{4}{*}{Yeast$_2$} & -- & 0.948 $\pm$ 0.013  & 0.950 $\pm$ 0.009   & 0.947 $\pm$ 0.011  & 0.946 $\pm$ 0.011  & \textbf{0.978}  & \textbf{0.977}  & 0.976  &  {0.976}  \\
      & 0.001 & 0.947 $\pm$ 0.016  &0.946 $\pm$ 0.016  &0.946 $\pm$ 0.016  &0.950 $\pm$ 0.012  &0.976  &0.976  &0.976  & \textbf{0.979}  \\
     & 0.01     & 0.950 $\pm$ 0.010  & 0.949 $\pm$ 0.009  & 0.949 $\pm$ 0.009  & 0.947 $\pm$ 0.011   & \textbf{0.979}   &\textbf{0.979}  &\textbf{0.979}  & \textbf{0.977}  \\
      & 0.1     & 0.948 $\pm$ 0.010  & 0.947 $\pm$ 0.009  & 0.947 $\pm$ 0.009  &\textbf{0.952 $\pm$ 0.008}  & \textbf{0.977}  &\textbf{0.978}  &\textbf{0.978}  & \textbf{0.978}  \\
      \hline

  \multirow{4}{*}{Abal$_1$} & -- & 0.841 $\pm$ 0.089   &0.838 $\pm$ 0.073  & 0.838 $\pm$ 0.080  & 0.847 $\pm$ 0.071  & \textbf{0.963}  & 0.960  & \textbf{0.964}  & 0.961 \\
   & 0.001 & 0.828 $\pm$ 0.074  &0.830 $\pm$ 0.103  & 0.830 $\pm$ 0.103  & 0.799 $\pm$ 0.114  & {0.958}  & \textbf{0.963}  & \textbf{0.963}  & 0.961  \\
     & 0.01     &  0.842 $\pm$ 0.082  &0.839 $\pm$ 0.083  & 0.839 $\pm$ 0.083  & 0.825 $\pm$ 0.070  &0.964   & \textbf{0.962}  & \textbf{0.962}  & \textbf{0.964}  \\
      & 0.1     &  \textbf{0.864 $\pm$ 0.062}  & 0.833 $\pm$ 0.076  & 0.833 $\pm$ 0.076  & 0.837 $\pm$ 0.071  & \textbf{0.962}  & 0.958  & 0.958  & \textbf{0.964}  \\ \hline

      \multirow{4}{*}{Abal$_2$} & -- &0.561 $\pm$ 0.120   & 0.542 $\pm$ 0.110  & 0.534 $\pm$ 0.150  & 0.570 $\pm$ 0.104  & 0.831  &0.824  &\textbf{0.862}   &0.830  \\
       & 0.001 &  0.533 $\pm$ 0.131  & \textbf{0.577 $\pm$ 0.074}  & \textbf{0.577 $\pm$ 0.074}  & \textbf{0.580 $\pm$ 0.092}  & 0.831  &0.815   & 0.815  & 0.823 \\
      & 0.01     & 0.572 $\pm$ 0.121  & \textbf{0.577 $\pm$ 0.097}  &0.577 $\pm$ 0.097  &0.565 $\pm$ 0.138  & 0.842  & 0.830  &0.830   & \textbf{0.845}   \\
      & 0.1     &  0.529 $\pm$ 0.127  &0.574 $\pm$ 0.085   & 0.574 $\pm$ 0.085  & 0.566 $\pm$ 0.097  & {0.811}  & 0.831  & 0.831  & 0.817 \\ \hline
\hline
  \end{tabular}
  }
  \label{results:nsgaii:semantics}
\end{table*}

\begin{table*}
  \caption{Average ($\pm$ standard deviation) hypervolume of evolved Pareto-approximated fronts and PO fronts for the SPEA2 semantic-based methods (SDO, SSC, SCD) over 50 independent runs.
  Bold indicates better performance compared to the baseline SPEA2 results reported in Table \ref{tab:hyperarea:nsgaii:spea2}.}
 
  \centering
   \resizebox{0.85\textwidth}{!}{
  \begin{tabular}{cccccccccc}\hline

                             &                            & \multicolumn{8}{c}{Hypervolume}  \\
    &                            & \multicolumn{4}{c}{Average} & \multicolumn{4}{c}{PO Front}  \\
     &          & \multicolumn{4}{c}{UBSS} & \multicolumn{4}{c}{UBSS} \\ 
    &    LBSS              & 0.25 & 0.5& 0.75 & 1.0 & 0.25 & 0.5& 0.75 & 1.0 \\ \hline

    \multicolumn{10}{c}{\textsf{SPEA2 SDO}} \\ \hline

        \multirow{4}{*}{Ion} & -- & \textbf{0.859 $\pm$ 0.031} & \textbf{0.869 $\pm$ 0.029}  & \textbf{0.862 $\pm$ 0.034} & \textbf{0.865 $\pm$ 0.047}  & \textbf{0.951} & \textbf{0.952} & \textbf{0.950} & \textbf{0.961} \\
     & 0.001                & \textbf{0.858 $\pm$ 0.041} & \textbf{0.852 $\pm$ 0.075} & \textbf{0.870 $\pm$ 0.055}&  \textbf{0.874 $\pm$ 0.055} & \textbf{0.946} & \textbf{0.955} & \textbf{0.952} & \textbf{0.956} \\
     & 0.01     &  {\textbf{0.837 $\pm$ 0.097}} & \textbf{0.851 $\pm$ 0.077} & \textbf{{0.875 $\pm$ 0.032}} & \textbf{0.863 $\pm$ 0.049} & \textbf{0.956} & \textbf{0.951} & \textbf{0.953} & \textbf{0.959} \\
    & 0.1     &   \textbf{0.852 $\pm$ 0.071} & \textbf{0.856 $\pm$ 0.053} & \textbf{0.873 $\pm$ 0.035} & \textbf{0.862 $\pm$ 0.038}  & {0.947} & \textbf{0.949} & \textbf{0.952} & \textbf{0.950} \\
    \hline
    %% \multirow{4}{*}{Ion} & -- &  & & & & 0.940 & 0.942 & 0.953  & 0.942\\ 
    %%  & 0.001                & &  & & & 0.943 & 0.940 & 0.957 & 0.937  \\
    %%  & 0.01                 & &  & & & 0.937 & 0.939 &0.962 &0.946  \\
    %%   & 0.1                 & &  & & & 0.940 & 0.949 & 0.953 &  0.936 \\ \hline

  \multirow{4}{*}{Spect} & -- & \textbf{0.591 $\pm$ 0.020} & \textbf{{0.599 $\pm$ 0.021}} & \textbf{0.597 $\pm$ 0.018} & \textbf{0.595 $\pm$ 0.022} & \textbf{0.678} & \textbf{0.688} & \textbf{0.686} & \textbf{0.695} \\
                       &0.001& \textbf{0.569 $\pm$ 0.021} & \textbf{0.565 $\pm$ 0.024} & \textbf{0.566 $\pm$ 0.023} & \textbf{0.563 $\pm$ 0.023} & \textbf{0.668} & \textbf{0.666} & \textbf{0.672} & {0.658} \\
                       & 0.01  & \textbf{0.568 $\pm$ 0.023}  & \textbf{0.567 $\pm$ 0.024} & \textbf{0.564 $\pm$ 0.025} & \textbf{0.563 $\pm$ 0.023} &  \textbf{0.666} & \textbf{0.674} & \textbf{0.664} & {0.658}  \\
  & 0.1  & \textbf{0.566 $\pm$ 0.023} & {\textbf{0.560 $\pm$ 0.020}} & \textbf{0.567 $\pm$ 0.027} & \textbf{0.561 $\pm$ 0.022} & \textbf{0.666} & {0.654} & \textbf{0.673} &  {0.658} \\
  \hline

  \multirow{4}{*}{Yeast$_1$} & -- & \textbf{0.850 $\pm$ 0.007}  & \textbf{0.850 $\pm$ 0.006} & \textbf{0.849 $\pm$ 0.008} & \textbf{0.849 $\pm$ 0.004} & \textbf{0.882} &  \textbf{0.881} & \textbf{0.881} & \textbf{0.881} \\
   & 0.001 &\textbf{0.848 $\pm$ 0.006} & \textbf{0.847 $\pm$ 0.007} & \textbf{0.848 $\pm$ 0.004} & \textbf{0.850 $\pm$ 0.006} & \textbf{0.880} & \textbf{0.883} & \textbf{0.880} & \textbf{0.883} \\
     & 0.01     & \textbf{0.848 $\pm$ 0.006} &\textbf{0.847 $\pm$ 0.006} & \textbf{{0.850 $\pm$ 0.005}} & \textbf{0.850 $\pm$ 0.005} & \textbf{0.881} & \textbf{0.880} & \textbf{0.882} & \textbf{0.879} \\
  & 0.1     &  {\textbf{0.847 $\pm$ 0.005}} & \textbf{0.849 $\pm$ 0.006} & \textbf{0.848 $\pm$ 0.005} & \textbf{0.849 $\pm$ 0.006} & \textbf{0.879} & \textbf{0.882} & \textbf{0.880} & \textbf{0.882} \\
  \hline

  %% \multirow{4}{*}{Yeast$_1$} & -- & & & & & & & & \\
  %%  & 0.001 & &  & & & & & &  \\ & & & & & & & \\
  %%    & 0.01     & &  & & & & & &  \\
  %%     & 0.1     & &  & & & & & &  \\ \hline

    \multirow{4}{*}{Yeast$_2$} & -- & \textbf{0.962 $\pm$ 0.007}  & \textbf{0.962 $\pm$ 0.006} & \textbf{0.962 $\pm$ 0.006} & \textbf{0.963 $\pm$ 0.008} & \textbf{0.979} & \textbf{0.979} & \textbf{0.979} & {0.977} \\
   & 0.001 & {\textbf{0.958 $\pm$ 0.008}} & \textbf{0.960 $\pm$ 0.007} & \textbf{0.960 $\pm$ 0.005} & \textbf{0.960 $\pm$ 0.005} & \textbf{0.980} & \textbf{0.979} & \textbf{0.979} & {0.977} \\
     & 0.01     & \textbf{0.959 $\pm$ 0.008} & \textbf{0.961 $\pm$ 0.007} & \textbf{0.961 $\pm$ 0.005} & \textbf{0.962 $\pm$ 0.007} & \textbf{0.979} & \textbf{0.980} & {0.978} & {0.978}  \\
  & 0.1     & \textbf{0.961 $\pm$ 0.007} & \textbf{0.961 $\pm$ 0.007} & \textbf{0.960 $\pm$ 0.007} & \textbf{{0.964 $\pm$ 0.007}} & \textbf{0.980} & \textbf{0.979} & \textbf{0.979} & \textbf{0.980} \\
  \hline

  %% \multirow{4}{*}{Yeast$_2$} & -- &  & & & & & & & \\
  %%  & 0.001 &  & & & & & & & \\
  %%    & 0.01     &  & & & & & & & \\
  %% & 0.1     &  & & & & & & & \\ \hline

      \multirow{4}{*}{Abal$_1$} & -- & \textbf{0.875 $\pm$ 0.059} & {\textbf{0.868 $\pm$ 0.081}} & \textbf{0.875 $\pm$ 0.059} & \textbf{0.873 $\pm$ 0.069} & \textbf{0.965} & \textbf{0.974} & \textbf{0.968} & \textbf{0.972} \\
   & 0.001 & \textbf{0.895 $\pm$ 0.061} & \textbf{0.911 $\pm$ 0.031} & \textbf{0.905 $\pm$ 0.044} & \textbf{0.903 $\pm$ 0.036} & \textbf{0.974} & \textbf{0.973} & \textbf{0.972} & \textbf{0.972} \\
     & 0.01     &  \textbf{0.903 $\pm$ 0.038} & \textbf{0.906 $\pm$ 0.042} & \textbf{0.901 $\pm$ 0.048} &\textbf{0.910 $\pm$ 0.039} & \textbf{0.966} & \textbf{0.969} & \textbf{0.972} & \textbf{0.974} \\
  & 0.1     & \textbf{0.888 $\pm$ 0.067}  & \textbf{{0.918 $\pm$ 0.032}} & \textbf{0.910 $\pm$ 0.046} & \textbf{0.916 $\pm$ 0.027} & \textbf{0.974} & \textbf{0.970} & \textbf{0.968} & \textbf{0.967} \\
  \hline

  %% \multirow{4}{*}{Abal$_1$} & -- & 0.888 $\pm$ 0.065 & 0.893 $\pm$ 0.077 &0.894 $\pm$ 0.056  & 0.896 $\pm$ 0.035& 0.968 & 0.965 & 0.971 & 0.962 \\
  %%  & 0.001 & 0.900 $\pm$ 0.044  & 0.904 $\pm$ 0.041& 0.909 $\pm$ 0.025 & 0.915 $\pm$ 0.025 & 0.964 & 0.974 & 0.966 & 0.969 \\
  %%    & 0.01     &  0.907 $\pm$ 0.035 & 0.901 $\pm$ 0.058 & 0.912 $\pm$ 0.024 & 0.919 $\pm$ 0.031 & 0.968 & 0.966 & 0.965 & 0.969 \\
  %%     & 0.1     &  0.908 $\pm$ 0.033 & 0.907 $\pm$ 0.051 & 0.911 $\pm$ 0.030 & 0.909 $\pm$ 0.027 & 0.972 & 0.971 & 0.968 & 0.964\\ \hline

        \multirow{4}{*}{Abal$_2$} & -- & {\textbf{0.620 $\pm$ 0.148}}  & \textbf{0.633 $\pm$ 0.124}  & \textbf{0.651 $\pm$ 0.146} & \textbf{0.630 $\pm$ 0.138} & \textbf{0.874} & \textbf{0.861} & \textbf{0.879} & \textbf{0.876}\\
 & 0.001 &  \textbf{0.717 $\pm$ 0.069} & \textbf{0.709 $\pm$ 0.079} & \textbf{0.722 $\pm$ 0.083} & \textbf{0.733 $\pm$ 0.075} & \textbf{0.868} &\textbf{0.883} & \textbf{0.886} & \textbf{0.891} \\
       & 0.01     &\textbf{0.706 $\pm$ 0.084} & \textbf{0.720 $\pm$ 0.067} & \textbf{0.726 $\pm$ 0.067} & \textbf{0.747 $\pm$ 0.070} & \textbf{0.884} & \textbf{0.880} & \textbf{0.877} & \textbf{0.887} \\
      & 0.1     & \textbf{0.732 $\pm$ 0.064} &  \textbf{0.733 $\pm$ 0.066} & \textbf{{0.749 $\pm$ 0.063}} & \textbf{0.737 $\pm$ 0.081} & \textbf{0.880} & \textbf{0.876} & \textbf{0.883} & \textbf{0.877}\\ \hline 

      %% \multirow{4}{*}{Abal$_2$} & -- & & & & & & & & \\
      %%  & 0.001 &  & & & & & & & \\
      %% & 0.01     & & & & & & & & \\
      %% & 0.1     &   & & & & & & & \\ \hline

      %\multicolumn{2}{c}{Better () / Worse (-) }  & \multirow{1}{*}{24 / 0} & \multirow{1}{*}{ 24 / 0} & \multirow{1}{*}{24 / 0} & \multirow{1}{*}{24 / 0} & \multirow{1}{*}{22 / 2}   & \multirow{1}{*} {23 / 1} & \multirow{1}{*}{ 23 / 0} & \multirow{1}{*}{18 / 5}  \\%424\\

            % \multicolumn{2}{c}{Eq. ($\equiv$) / NS (x)}  & \multirow{1}{*}{0 / 0}  & \multirow{1}{*}{ 0/ 0} & \multirow{1}{*}{0 / 0}  & \multirow{1}{*}{0 / 0}  & \multirow{1}{*}{0 / -}    & \multirow{1}{*}{0 / -}  &  \multirow{1}{*}{1 / -} & \multirow{1}{*}{1 / -}  \\

\hline

      \hline     
 \multicolumn{10}{c}{\textsf{SPEA2 SSC}} \\ \hline

    %% \multirow{4}{*}{Ion} & -- & 0.846 $\pm$ 0.032 &0.834 $\pm$ 0.061 &0.839 $\pm$ 0.051 & 0.840 $\pm$ 0.037 & 0.944 &0.941  & 0.941 & 0.944\\ 
    %%  & 0.001                &0.844 $\pm$ 0.060 &0.835 $\pm$ 0.081  & 0.832 $\pm$ 0.049 & 0.842 $\pm$ 0.034 & 0.949 & 0.940 & 0.942 & 0.935  \\
    %%  & 0.01                 & 0.845 $\pm$ 0.060 & 0.835 $\pm$ 0.081  &0.832 $\pm$ 0.049 & 0.842 $\pm$ 0.034& 0.949 & 0.939 & 0.942 & 0.935 \\
    %%   & 0.1                 & 0.825 $\pm$ 0.081 & 0.829 $\pm$ 0.060  &0.831 $\pm$ 0.069 &0.848 $\pm$ 0.032 & 0.948 & 0.929 & 0.944 & 0.938 \\ \hline

       \multirow{4}{*}{Ion}  & -- &  0.724 $\pm$ 0.157  & 0.767 $\pm$ 0.081  &0.743 $\pm$ 0.120  & 0.764 $\pm$ 0.100  & 0.935  & 0.924  & 0.939  & 0.939 \\
      & 0.001 &  {0.747 $\pm$ 0.173}   & {0.767 $\pm$ 0.121}  & {0.755 $\pm$ 0.155}  & \textbf{0.790 $\pm$ 0.101}  & \textbf{0.951}  & {0.936}  & {0.934}  & {0.947}  \\
     & 0.01     & {0.741 $\pm$ 0.172}  & {0.765 $\pm$ 0.117}   & {0.757 $\pm$ 0.147}  & \textbf{0.790 $\pm$ 0.101}  & \textbf{0.955}  & {0.942}  & {0.940}  & {0.947}  \\
      & 0.1     & \textbf{0.787 $\pm$ 0.106}  & {0.782 $\pm$ 0.108}  & {0.778 $\pm$ 0.124}  & \textbf{0.787 $\pm$ 0.119}  &  {0.939}  & {0.942}  & \textbf{0.956}  & \textbf{0.961}  \\
      \hline

      \multirow{4}{*}{Spect} & -- &  0.521 $\pm$ 0.045  & 0.536 $\pm$ 0.021  & 0.543 $\pm$ 0.028  & 0.533 $\pm$ 0.027  & 0.639  & 0.648  & 0.657  & 0.650  \\
      & 0.001 &0.533 $\pm$ 0.028   &0.536 $\pm$ 0.022  & 0.530 $\pm$ 0.035  & 0.536 $\pm$ 0.021  & 0.644  & 0.659  & 0.634  & 0.634  \\ 
     & 0.01     & 0.530 $\pm$ 0.027  & 0.534 $\pm$ 0.029  &0.535 $\pm$ 0.023  &0.538 $\pm$ 0.028  & 0.648  & 0.644  & 0.636  & \textbf{0.660}  \\
      & 0.1     & 0.537 $\pm$ 0.021  & 0.542 $\pm$ 0.025  & 0.536 $\pm$ 0.034  &0.533 $\pm$ 0.028  & 0.640  & 0.650  & 0.641  & 0.645  \\
      \hline

   %% \multirow{4}{*}{Spect} & -- & & & & & & & & \\ 
   %%                     & 0.001 & &  & & & & & &  \\
   %%                     & 0.01  & &  & & & & & &  \\
   %%                     & 0.1  & &  & & & & & &  \\ \hline

         \multirow{4}{*}{Yeast$_1$} & -- &  0.824 $\pm$ 0.030  &0.824 $\pm$ 0.042   & 0.831 $\pm$ 0.020  & 0.828 $\pm$ 0.030  & 0.877  &0.876  &0.877   &0.874  \\
      & 0.001 &  {0.826 $\pm$ 0.029}  & 0.824 $\pm$ 0.062  &0.828 $\pm$ 0.025  &0.833 $\pm$ 0.017  & {0.877}  & 0.875  & 0.876  & 0.877  \\
     & 0.01     &  0.830 $\pm$ 0.020  & 0.829 $\pm$ 0.033  & 0.832 $\pm$ 0.021  & 0.832 $\pm$ 0.020  & 0.874  & 0.876  & 0.876  & 0.876  \\
      & 0.1     & 0.828 $\pm$ 0.032  &0.836 $\pm$ 0.015  &0.830 $\pm$ 0.028  &0.836 $\pm$ 0.014  & 0.875  & 0.877  & 0.877  & 0.876  \\
      \hline

  %% \multirow{4}{*}{Yeast$_1$} & -- & & & & & & & & \\
  %%  & 0.001 & &  & & & & & &  \\ & & & & & & & \\
  %%    & 0.01     & &  & & & & & &  \\
  %%     & 0.1     & &  & & & & & &  \\ \hline

            \multirow{4}{*}{Yeast$_2$} & -- & \textbf{0.950 $\pm$ 0.010}  & \textbf{0.947 $\pm$ 0.011}  & \textbf{0.950 $\pm$ 0.010}  & \textbf{0.951 $\pm$ 0.010}  &0.977   &0.976   &0.978  & \textbf{0.979} \\
      & 0.001 & \textbf{0.947 $\pm$ 0.015}  & \textbf{0.947 $\pm$ 0.010}  & \textbf{0.948 $\pm$ 0.011}  & \textbf{0.948 $\pm$ 0.010}  & 0.978  & 0.977  & 0.978  &0.976  \\
     & 0.01     & \textbf{0.948 $\pm$ 0.012}  &\textbf{0.948 $\pm$ 0.013}  &0.943 $\pm$ 0.022  & \textbf{0.950 $\pm$ 0.010}  & 0.978  & \textbf{0.979}  & 0.977  & 0.978  \\
      & 0.1     & 0.944 $\pm$ 0.024  &0.943 $\pm$ 0.017  &\textbf{0.947 $\pm$ 0.010}  & 0.945 $\pm$ 0.015  & 0.976  & 0.977  & 0.975  & 0.975  \\
      \hline

  %% \multirow{4}{*}{Yeast$_2$} & -- &  & & & & & & & \\
  %%  & 0.001 &  & & & & & & & \\
  %%    & 0.01     &  & & & & & & & \\
  %%     & 0.1     &  & & & & & & & \\ \hline

            \multirow{4}{*}{Abal$_1$} & -- &0.831 $\pm$ 0.071   & \textbf{0.856 $\pm$ 0.088}  & 0.822 $\pm$ 0.080  & \textbf{0.851 $\pm$ 0.061}  & 0.960  & 0.960  & 0.966  & 0.961 \\
      & 0.001 &0.812 $\pm$ 0.094   & \textbf{0.854 $\pm$ 0.082}  & \textbf{0.836 $\pm$ 0.076}  &\textbf{0.847 $\pm$ 0.065}  & 0.963  & 0.965  & 0.966  & 0.963  \\
     & 0.01     &  0.819 $\pm$ 0.098  & 0.824 $\pm$ 0.106  & \textbf{0.841 $\pm$ 0.070}  & \textbf{0.851 $\pm$ 0.063}  & \textbf{0.969}  & \textbf{0.965}  & \textbf{0.964}  & \textbf{0.962}  \\
      & 0.1     & \textbf{0.844 $\pm$ 0.083}  & \textbf{0.833 $\pm$ 0.088}  & \textbf{0.853 $\pm$ 0.095}  & \textbf{0.837 $\pm$ 0.090}  & \textbf{0.965}  & \textbf{0.967}  & \textbf{0.963}  & \textbf{0.965}  \\
            \hline

                  \multirow{4}{*}{Abal$_2$} & -- & \textbf{0.548 $\pm$ 0.120}  & 0.500 $\pm$ 0.139  & 0.515 $\pm$ 0.137  & 0.532 $\pm$ 0.107  & 0.819  & 0.790   &0.815  & 0.802 \\
      & 0.001 & 0.515 $\pm$ 0.135  &0.518 $\pm$ 0.125  & 0.541 $\pm$ 0.111  &0.521 $\pm$ 0.127  & 0.812  & 0.829  & \textbf{0.836}  & 0.807  \\
     & 0.01     & 0.537 $\pm$ 0.105  & 0.500 $\pm$ 0.163  & 0.527 $\pm$ 0.152  & {0.521 $\pm$ 0.095}  & 0.820  & 0.817  & 0.816  & {0.816}  \\
      & 0.1     & \textbf{0.561 $\pm$ 0.111}  & \textbf{0.556 $\pm$ 0.094}  & 0.516 $\pm$ 0.142  & \textbf{0.558 $\pm$ 0.098}  & \textbf{0.840}  & \textbf{0.838}  & \textbf{0.838}  & \textbf{0.838}  \\
      \hline

      %\multicolumn{2}{c}{Better () / Worse (-) }  & \multirow{1}{*}{0 / 0} & \multirow{1}{*}{ 0 / 0} & \multirow{1}{*}{0 / 0} & \multirow{1}{*}{0 / 0} & \multirow{1}{*}{6 / 13}   & \multirow{1}{*} {5 / 16} & \multirow{1}{*}{ 7 / 13} & \multirow{1}{*}{8 / 14}  \\%424\\

             %\multicolumn{2}{c}{Eq. ($\equiv$) / NSS }  & \multirow{1}{*}{0 / 24}  & \multirow{1}{*}{ 0/ 24} & \multirow{1}{*}{0 / 24}  & \multirow{1}{*}{0 / 24}  & \multirow{1}{*}{5 / -}    & \multirow{1}{*}{3 / -}  &  \multirow{1}{*}{4 / -} & \multirow{1}{*}{2 / -}  \\

\hline

      %% \multirow{4}{*}{Abal$_2$} & -- & & & & & & & & \\
      %%  & 0.001 &  & & & & & & & \\
      %% & 0.01     & & & & & & & & \\
      %% & 0.1     &   & & & & & & & \\ \hline

%      \multicolumn{2}{c}{Better() / Worse(-) / Equal(=)} & \\
 %     \multicolumn{2}{c}{w.r.t. Canonical SPEA2} & \\
\hline
      
\multicolumn{10}{c}{\textsf{SPEA2 SCD}} \\ \hline

   \multirow{4}{*}{Ion} & -- & \textbf{0.804 $\pm$ 0.100}  & 0.785 $\pm$ 0.138  & 0.785 $\pm$ 0.116  & \textbf{0.819 $\pm$ 0.073}  &  \textbf{0.956}  & \textbf{0.951}  & 0.948  & 0.944 \\
                       & 0.001 & \textbf{0.790 $\pm$ 0.093}  & 0.786 $\pm$ 0.103  & 0.786 $\pm$ 0.103  & \textbf{0.790 $\pm$ 0.088}  & \textbf{0.950}  &0.942  & 0.942  &\textbf{ 0.952}  \\ 
                       & 0.01  & 0.765 $\pm$ 0.156  &  \textbf{0.803 $\pm$ 0.074}  & \textbf{0.803 $\pm$ 0.074}  & \textbf{0.794 $\pm$ 0.098}  & \textbf{0.962}  & 0.938  & 0.938  &  \textbf{0.949}  \\
                       & 0.1  & \textbf{0.806 $\pm$ 0.110}  & \textbf{0.794 $\pm$ 0.096}   & \textbf{0.794 $\pm$ 0.096}  & 0.773 $\pm$ 0.142  &  \textbf{0.956}  & 0.941   & 0.941  &\textbf{ 0.955}  \\ \hline

         \multirow{4}{*}{Spect} & -- &0.541 $\pm$ 0.027  & \textbf{0.547 $\pm$ 0.020}  & \textbf{0.548 $\pm$ 0.023}  &0.533 $\pm$ 0.027  & 0.656  & 0.652  & \textbf{0.663}  & 0.632  \\
      & 0.001 & \textbf{0.546 $\pm$ 0.024}  &0.544 $\pm$ 0.024  & 0.543 $\pm$ 0.026  & 0.540 $\pm$ 0.027  &0.645   &0.659    & 0.655   & 0.651 \\ 
     & 0.01     & \textbf{0.545 $\pm$ 0.023}  &0.544 $\pm$ 0.024  &0.543 $\pm$ 0.026  &0.540 $\pm$ 0.027  & 0.647  & 0.659  & 0.654  & 0.651 \\
      & 0.1     & \textbf{0.546 $\pm$ 0.018}  &0.543 $\pm$ 0.025  &0.542 $\pm$ 0.025  &0.541 $\pm$ 0.025  &  0.646  &  0.653   & 0.657    & 0.655  \\
      \hline

   %% \multirow{4}{*}{Spect} & -- & & & & & & & & \\ 
   %%                     & 0.001 & &  & & & & & &  \\
   %%                     & 0.01  & &  & & & & & &  \\
   %%                     & 0.1  & &  & & & & & &  \\ \hline

     %%     \multirow{4}{*}{Yeast$_1$} & -- & 0.832 $\pm$ 0.021& 0.836 $\pm$ 0.015 & 0.836 $\pm$ 0.015 & 0.839 $\pm$ 0.009 & 0.876 & 0.875 & 0.878 & 0.876 \\
     %%  & 0.001 &0.838 $\pm$ 0.011  & 0.850 $\pm$ 0.006 & 0.839 $\pm$ 0.008 & 0.829 $\pm$ 0.045 & 0.878 & 0.878 & 0.878 & 0.876 \\
     %% & 0.01     &  0.829 $\pm$ 0.037 &  0.836 $\pm$ 0.011 & 0.836 $\pm$ 0.011 &0.837 $\pm$ 0.012 & 0.878 & 0.877 & 0.877 & 0.876 \\
     %%  & 0.1     & 0.833 $\pm$ 0.014 & 0.832 $\pm$ 0.024 &0.832 $\pm$ 0.024 & 0.832 $\pm$ 0.024& 0.875 & 0.876 & 0.876 & 0.876 \\
     %%  \hline

  \multirow{4}{*}{Yeast$_1$} & -- & 0.835 $\pm$ 0.015  & 0.836 $\pm$ 0.011  & 0.836 $\pm$ 0.014  & 0.837 $\pm$ 0.016 & \textbf{0.878}  & 0.876  & \textbf{0.879}  &  \textbf{0.880}  \\
   & 0.001 & 0.835 $\pm$ 0.020  & 0.835 $\pm$ 0.014  &0.834 $\pm$ 0.014  &0.836 $\pm$ 0.020  & 0.874  & \textbf{0.878}  & 0.876  & \textbf{0.878}  \\ 
     & 0.01     & 0.838 $\pm$ 0.008  & 0.832 $\pm$ 0.023  &0.832 $\pm$ 0.017  & 0.836 $\pm$ 0.011  &0.876   & \textbf{0.879}  & 0.875  & 0.874  \\
      & 0.1     & 0.835 $\pm$ 0.015  & 0.825 $\pm$ 0.061   &0.834 $\pm$ 0.023   & 0.835 $\pm$ 0.012  & 0.877  & 0.875  & \textbf{0.879}  & 0.876  \\ \hline

  \multirow{4}{*}{Yeast$_2$} & -- &  \textbf{0.949 $\pm$ 0.008}  & 0.946 $\pm$ 0.015  & \textbf{0.948 $\pm$ 0.009}  & \textbf{0.948 $\pm$ 0.009}  &0.977  &0.978   & 0.976  & 0.978 \\
   & 0.001 & \textbf{0.948 $\pm$ 0.012}  & \textbf{0.948 $\pm$ 0.009}  & \textbf{0.948 $\pm$ 0.009}  & \textbf{0.947 $\pm$ 0.011}  & 0.978  & 0.977  & 0.977  &0.977  \\
     & 0.01     & \textbf{0.948 $\pm$ 0.010}  & \textbf{0.948 $\pm$ 0.011}  & \textbf{0.948 $\pm$ 0.011}  & \textbf{0.948 $\pm$ 0.015}  &0.975  & 0.977  & 0.977  & 0.977 \\
  & 0.1     &  \textbf{0.949 $\pm$ 0.010}  & \textbf{0.950 $\pm$ 0.012}  & \textbf{0.950 $\pm$ 0.012}  & 0.945 $\pm$ 0.036  &0.977  &0.978   & 0.978  & 0.977  \\ \hline

    \multirow{4}{*}{Abal$_1$} & -- & 0.829 $\pm$ 0.086  & \textbf{0.834 $\pm$ 0.072}  & \textbf{0.840 $\pm$ 0.083}  & \textbf{0.835 $\pm$ 0.074}  & \textbf{0.961}  &0.955  &0.960  & \textbf{0.968}  \\
   & 0.001 &   0.829 $\pm$ 0.072  & \textbf{0.834 $\pm$ 0.069}  & \textbf{0.834 $\pm$ 0.069}   &0.819 $\pm$ 0.088   & 0.959  & 0.957  &0.957   &0.960  \\
     & 0.01     &  \textbf{0.834 $\pm$ 0.084}  & 0.800 $\pm$ 0.078  & 0.800 $\pm$ 0.078  & 0.816 $\pm$ 0.097  &0.962   &0.960  &0.960  &0.952  \\
      & 0.1     &  \textbf{0.841 $\pm$ 0.085}  & 0.827 $\pm$ 0.071  &0.827 $\pm$ 0.071  & \textbf{0.842 $\pm$ 0.067}  & \textbf{0.965}  & 0.955  & 0.955  & \textbf{0.964}  \\ \hline

      \multirow{4}{*}{Abal$_2$} & -- &\textbf{0.572 $\pm$ 0.111}  &\textbf{0.577 $\pm$ 0.103}  & \textbf{0.562 $\pm$ 0.127}  & \textbf{0.560 $\pm$ 0.094}  &0.828  &\textbf{0.837}   &\textbf{0.856}   &\textbf{0.839}  \\
       & 0.001 & \textbf{0.557 $\pm$ 0.094}   & \textbf{0.578 $\pm$ 0.103}  & \textbf{0.578 $\pm$ 0.093}  & \textbf{0.554 $\pm$ 0.116}  & \textbf{0.838}  & \textbf{0.852}  & \textbf{0.852}  & \textbf{0.853}  \\
      & 0.01     & \textbf{0.569 $\pm$ 0.117}  & \textbf{0.576 $\pm$ 0.105}  & \textbf{0.576 $\pm$ 0.105}  & \textbf{0.580 $\pm$ 0.070}  & \textbf{0.838}  & \textbf{0.837}  & \textbf{0.837}  & 0.834 \\
      & 0.1     &  0.535 $\pm$ 0.128   & \textbf{0.567 $\pm$ 0.118}   & \textbf{0.567 $\pm$ 0.118}  & \textbf{0.586 $\pm$ 0.077}  &0.816  & 0.823  & 0.823  & \textbf{0.842}  \\ \hline

      %\multicolumn{2}{c}{Better () / Worse (-) }  & \multirow{1}{*}{0 / 0} & \multirow{1}{*}{ 0 / 0} & \multirow{1}{*}{0 / 0} & \multirow{1}{*}{0 / 0} & \multirow{1}{*}{10 / 12}   & \multirow{1}{*} {6 / 13} & \multirow{1}{*}{ 6 / 14} & \multirow{1}{*}{10 / 11}  \\%424+\\

             %\multicolumn{2}{c}{Eq. ($\equiv$) / NS (x)}  & \multirow{1}{*}{0 / 24}  & \multirow{1}{*}{ 0/ 24} & \multirow{1}{*}{0 / 24}  & \multirow{1}{*}{0 / 24}  & \multirow{1}{*}{2 / -}    & \multirow{1}{*}{5 / -}  &  \multirow{1}{*}{4 / -} & \multirow{1}{*}{3 / -}  \\

\hline

%      \multicolumn{2}{c}{Better(+) / Worse(-) / Equal(=)} & \\
%      \multicolumn{2}{c}{w.r.t. Canonical SPEA2} & \\
\hline

  \end{tabular}
  }
  \label{results:spea2:semantics}
\end{table*}

The following statistical analysis was performed.
For each baseline method, NSGA-II and SPEA2, we perform three blocks of comparisons, one for each
semantic variant, namely SDO, SSC and SCD (six blocks in total). In each block we perform a multi-group test with $N$ configurations, where $N=16$ represents each of the configurations tested based on different UBSS and LBSS bound values which are compared with one of the baseline methods (NSGA or SPEA2). We use the Friedman test in our analysis, comparing based on all 6 problems and considering 50 replicates (runs) for each problem.
The null hypothesis is that the median performance of all groups in the same block is the same, and we reject the null hypothesis at the $\alpha=0.01$ significance level.
For both SSC and SCD, the null hypothesis of the multi-group tests was not rejected considering each of the baselines, with p-values above the significance level of the test.
However, the null hypothesis was rejected in the multi-group tests for the SDO configurations relative to both baselines.
In these cases, a post-hoc test was conducted, once again using the Friedman test to perform $N$
pairwise tests between each SDO configuration and the baseline method (NSGA-II and SPEA2).
The Bonferroni-Dunn correction of the p-value was performed to account for the family-wise error of performing
multiple comparisons.
Once again, in all pairwise comparisons the null hypothesis was rejected at the $\alpha=0.01$ significance level,
comparing each SDO configuration and the respective baseline.
Based on these results, we  found that the SDO method outperforms each of the baseline methods,
while SSC and SCD did not.
Moreover, we can see that SDO performance is robust to how it is parameterised, relative to the LBSS and UBSS values.

\begin{figure*}
  \centering
  \begin{tabular}{ccc}
     \scriptsize{Spect} & \scriptsize{Abal$_1$} & \scriptsize{Abal$_2$} \\
     
     \hspace{-0.82cm}  \includegraphics[width=0.370\textwidth]{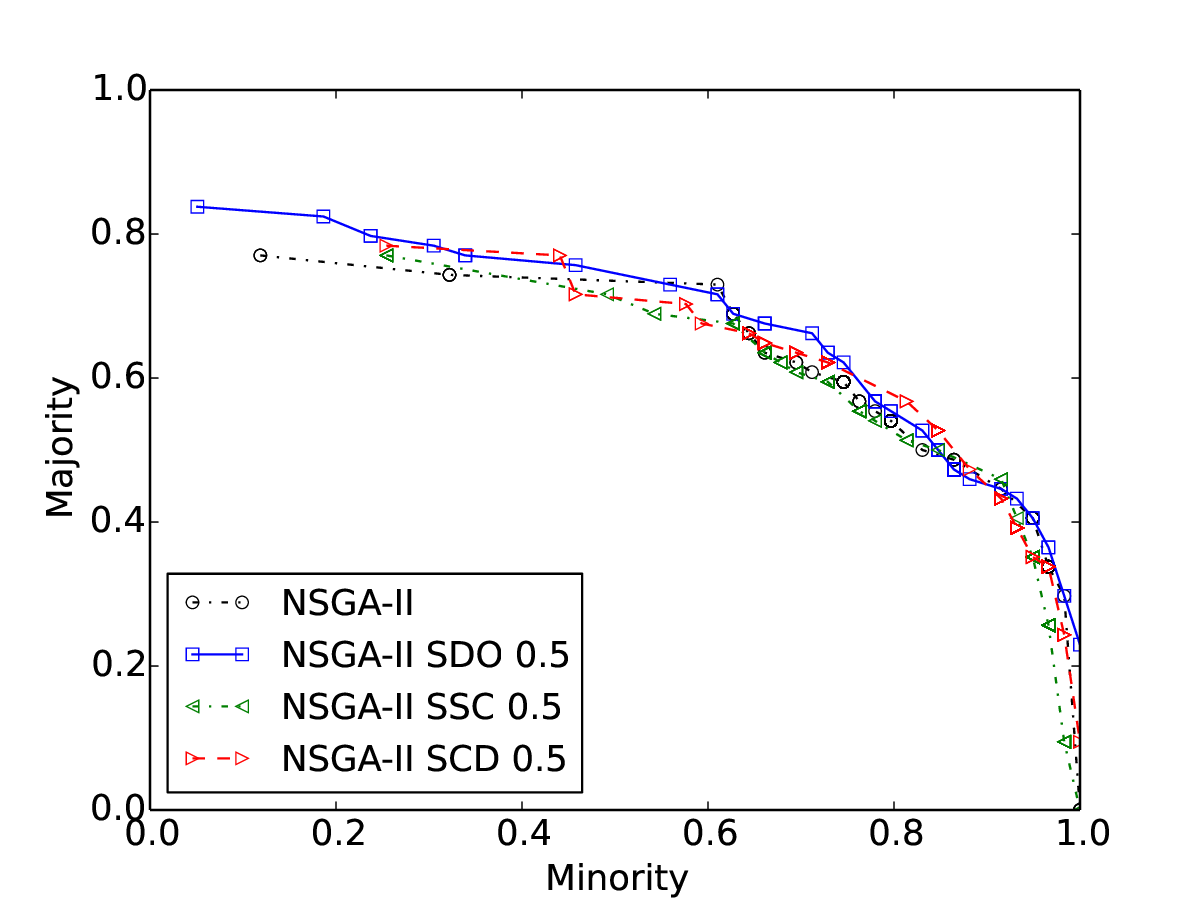}   & \hspace{-0.95cm}  \includegraphics[width=0.370\textwidth]{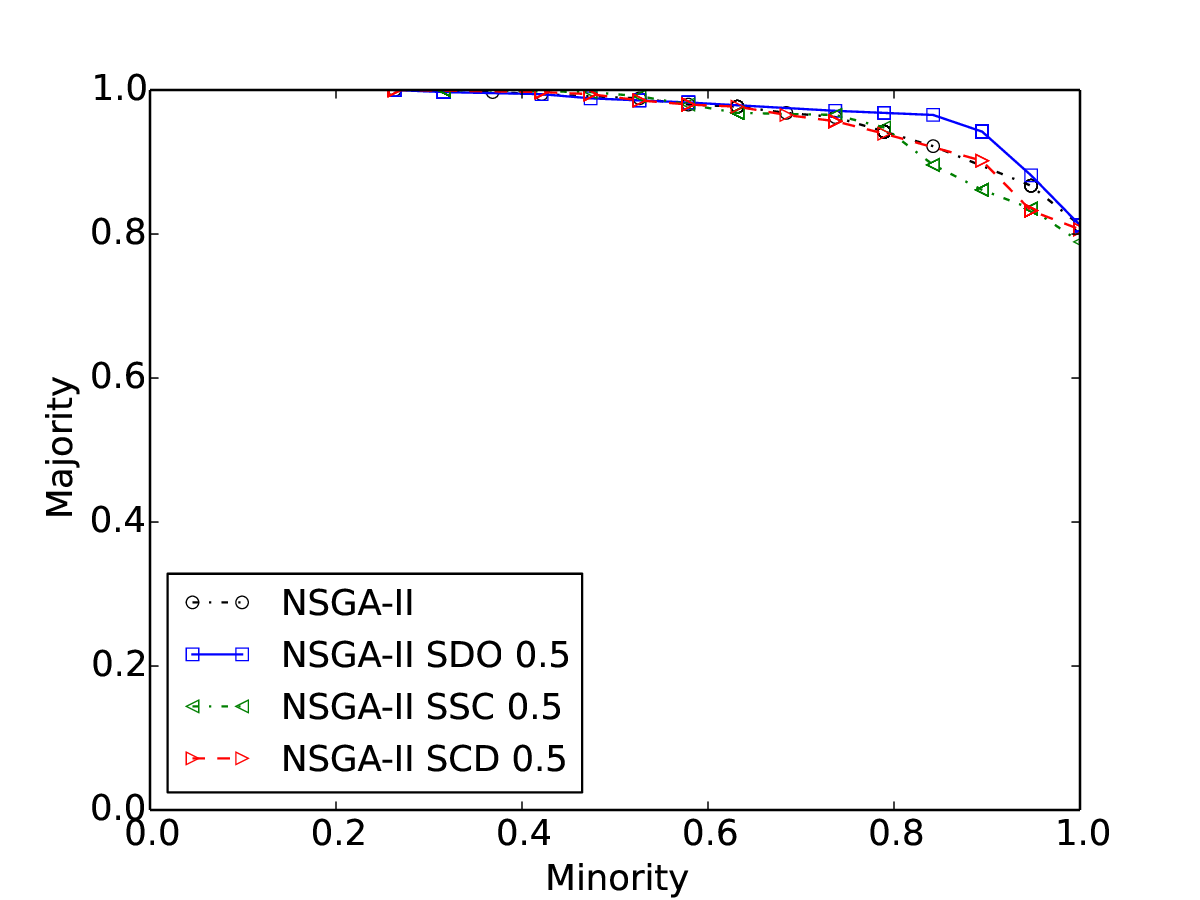} & \hspace{-0.95cm}  \includegraphics[width=0.370\textwidth]{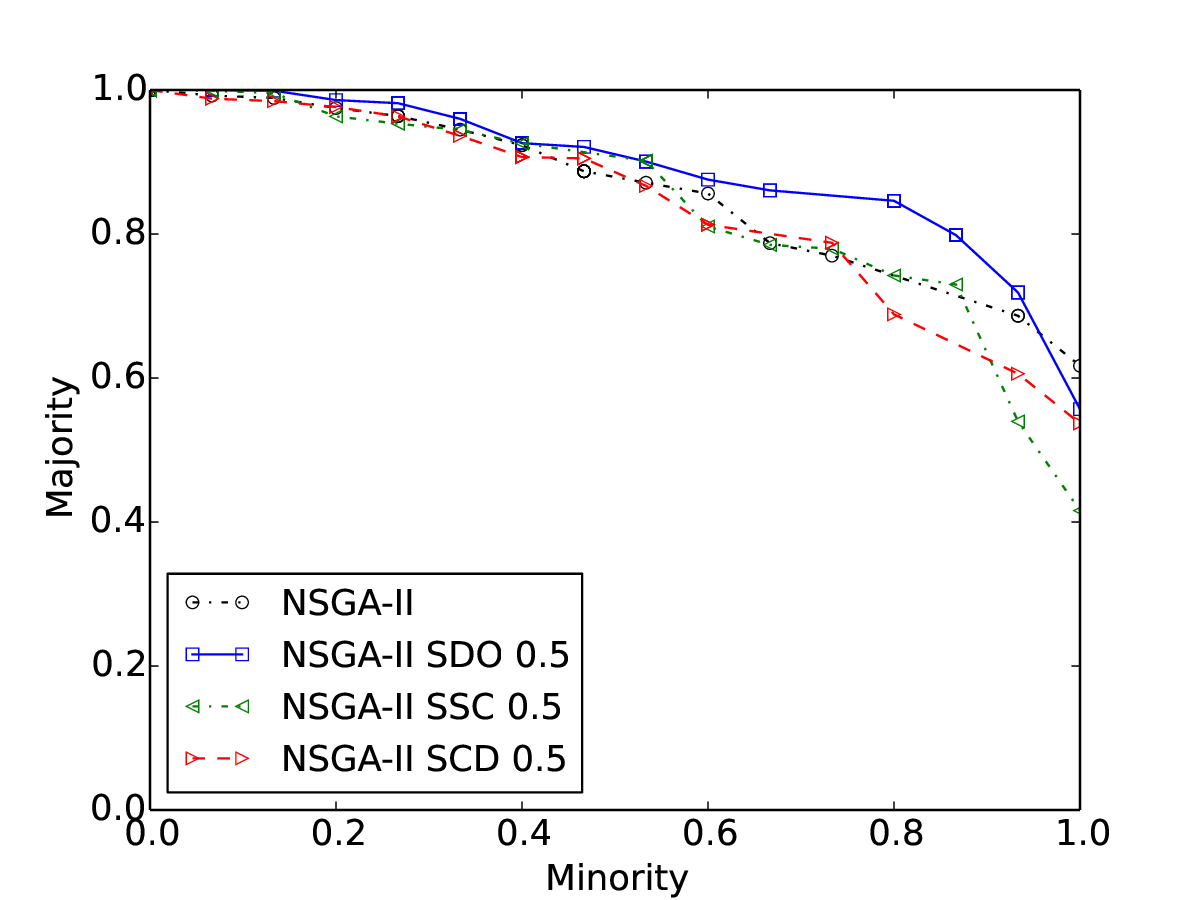} \\

     \hspace{-0.82cm}  \includegraphics[width=0.370\textwidth]{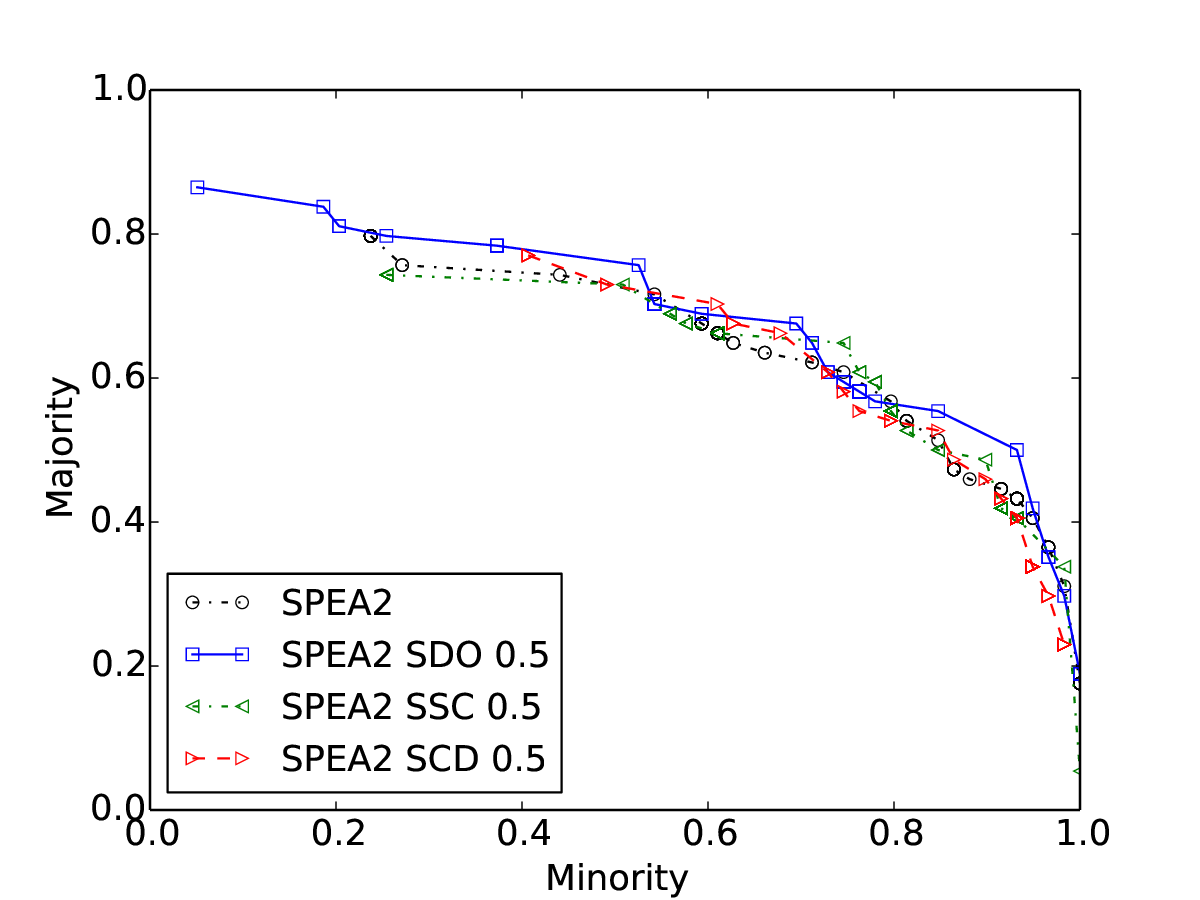}   & \hspace{-0.95cm}  \includegraphics[width=0.370\textwidth]{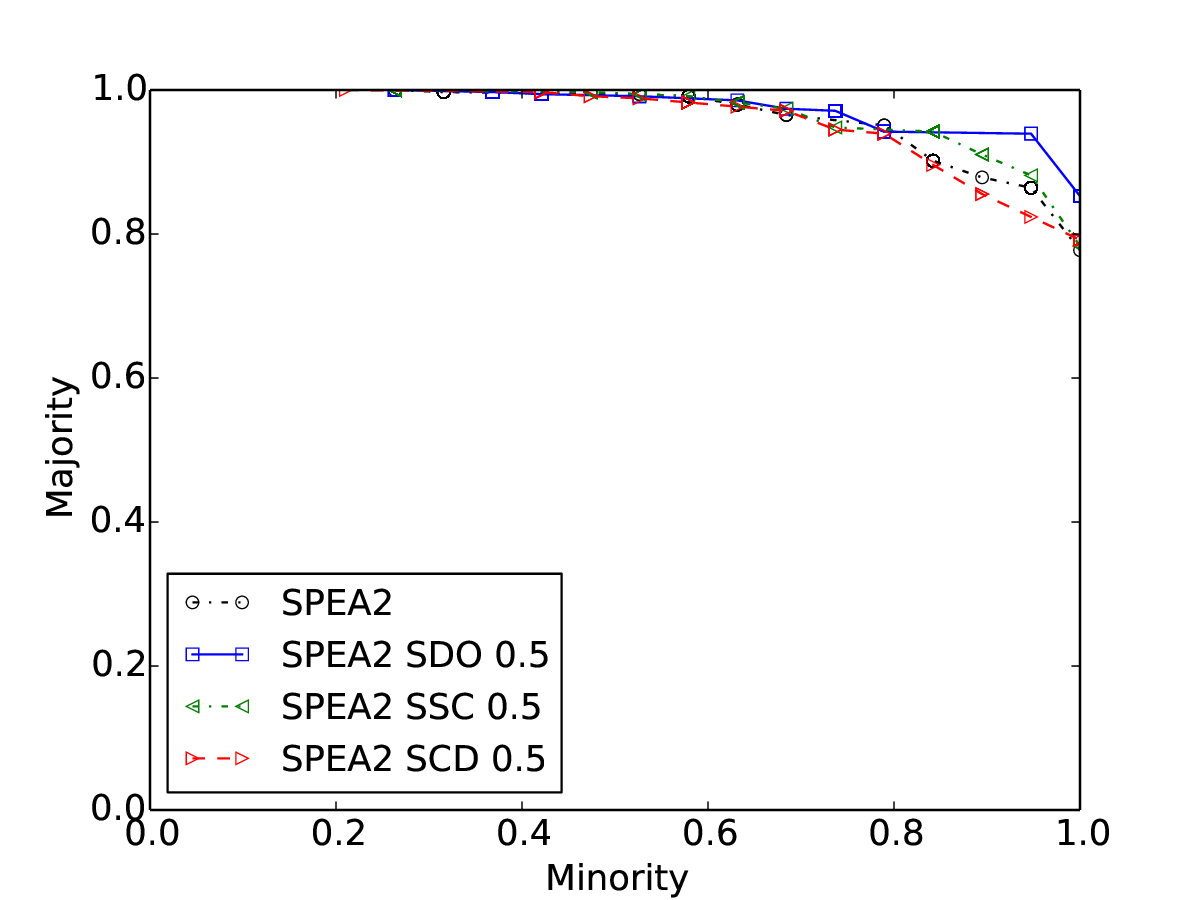} & \hspace{-0.95cm}  \includegraphics[width=0.370\textwidth]{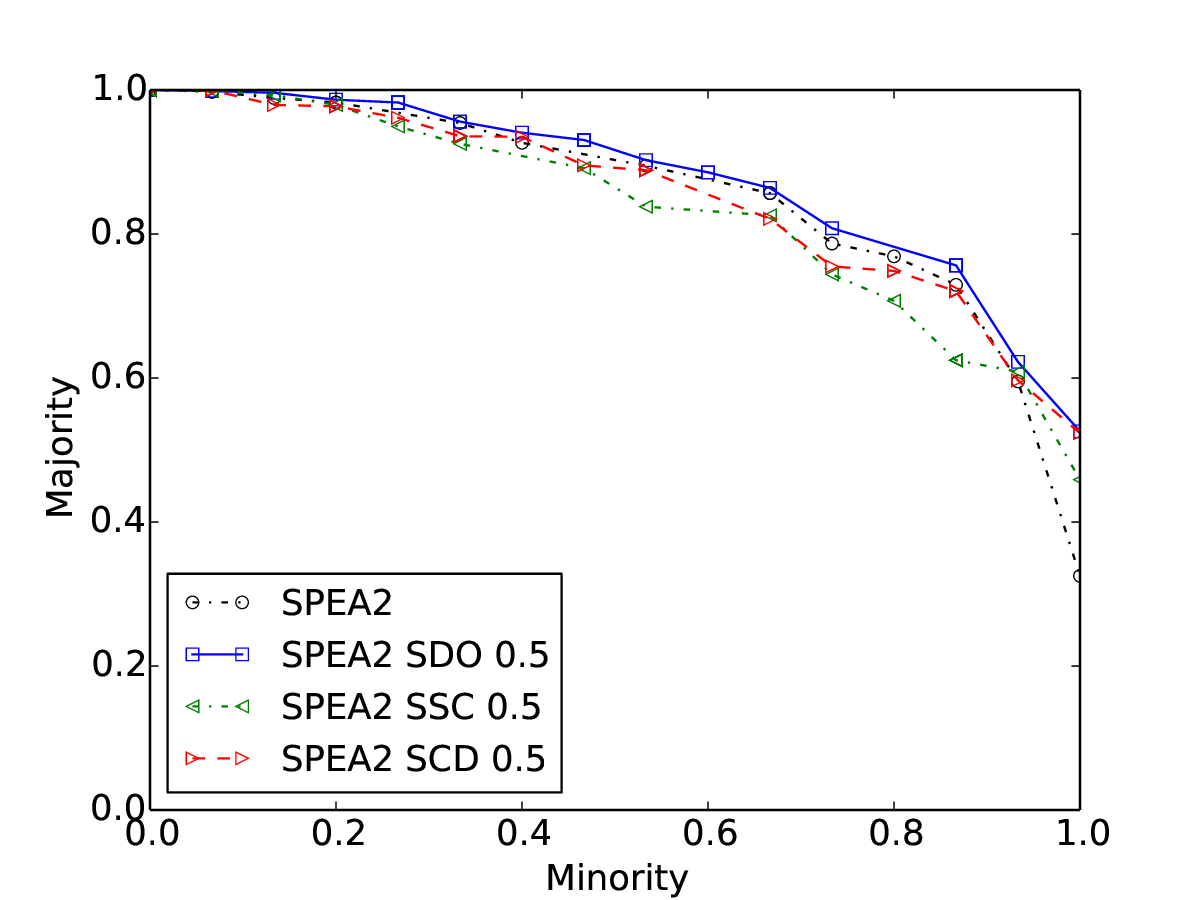}

\end{tabular}
\caption{Pareto fronts for  Spect, Abal$_1$, Abal$_2$ datasets using NSGA-II and its semantic-based variants (top) and  using SPEA2 and its variants (bottom). Semantic-based variants set UBSS = 0.5.} 
\label{fig:pareto}
\end{figure*}

From this statistical analysis, it is evident that SDO behaves, in average, better compared to any of the approaches used in this work (two canonical EMO approaches and the two other semantic-based methods).
To illustrate the performance of the proposed semantic methods, Figure~\ref{fig:pareto} shows the Pareto fronts obtained by each approach, setting UBSS at 0.5, as well as the Pareto fronts
obtained by NSGA-II and SPEA2. The figure shows the fronts for the Spect, Abal$_1$ and Abal$_2$ datasets, as representative examples.
From this figure, it is easy to see how SDO, represented by  blue hollow squares connected by a solid line, achieves better coverage of the objective space compared to the other approaches.
This is particularly clear for the Abal datasets, using either NSGA-II (top) or SPEA2 (bottom) as well as for the Spect dataset when using SPEA2 and less clear when using NSGA-II.

%% \begin{figure*}
%%   \centering
%%   \begin{tabular}{cccc}
%%     \scriptsize{Ion} & \scriptsize{Spect} & \scriptsize{Abal$_1$} & \scriptsize{Abal$_2$} \\
%%     \hspace{-0.82cm}  \includegraphics[width=0.280\textwidth]{figures/mo_ion_nsgaii} & \hspace{-0.95cm}  \includegraphics[width=0.280\textwidth]{figures/mo_spect_nsgaii}   & \hspace{-0.95cm}  \includegraphics[width=0.280\textwidth]{figures/mo_abalone_9_18_nsgaii} & \hspace{-0.95cm}  \includegraphics[width=0.280\textwidth]{figures/mo_abalone_9_other_nsgaii} \\

%%         \hspace{-0.82cm}  \includegraphics[width=0.280\textwidth]{figures/mo_ion_spea2} & \hspace{-0.95cm}  \includegraphics[width=0.280\textwidth]{figures/mo_spect_spea2}   & \hspace{-0.95cm}  \includegraphics[width=0.280\textwidth]{figures/mo_abalone_9_18_spea2} & \hspace{-0.95cm}  \includegraphics[width=0.280\textwidth]{figures/mo_abalone_9_other_spea2} 

%%  %% \scriptsize{Yeast$_1$} & \scriptsize{Yeast$_2$}&        \scriptsize{Abal$_1$}  \\
%%  %% \hspace{-0.820cm}  \includegraphics[width=0.355\textwidth]{figures/mo_mit_nsgaii_spea2} &        \hspace{-0.60cm}  \includegraphics[width=0.355\textwidth]{figures/mo_m3_nsgaii_spea2} & \hspace{-0.60cm}  \includegraphics[width=0.355\textwidth]{figures/mo_abalone_9_18_nsgaii_spea2}  \\
%% \end{tabular}
%% \caption{Pareto-optimal fronts for Ion, Spect, Abal$_1$, Abal$_2$ data sets using NSGA-II, NSGA-II SDO (UBSS = 0.5), SPEA2 and SPEA2 SDO (UBSS = 0.5).} 
%% \label{fig:pareto}
%% \end{figure*}

%%

\begin{table*}
  \caption{Average number of distinct solutions ($\pm$ std. deviation) occurring at the first front of generations 1, 10, 20, 20, 40 and 50 for canonical NSGA-II and NSGA-II SDO methods.}
 
  \centering
   \resizebox{1.0\textwidth}{!}{
  \begin{tabular}{c|cccccc}\hline

   & \multicolumn{6}{c}{\textsf{Generation}} \\\hline \hline
  Data & 1 & 10 & 20  & 30 & 40 & 50 \\
  set & mean $\pm$ std & mean $\pm$ std & mean $\pm$ std & mean $\pm$ std & mean $\pm$ std  & mean $\pm$ std \\ \hline
  & \multicolumn{6}{c}{\textsf{NSGA-II}} \\ \hline
 
Ion	&	11.32	$\pm$	2.3	&	9.98	$\pm$	3.57	&	8.42	$\pm$	3.44	&	8.38	$\pm$	2.91	&	8.68	$\pm$	2.95	&	8.28	$\pm$	3.14	\\	
Spect	&	13.78	$\pm$	2.14	&	21.36	$\pm$	2.23	&	21.22	$\pm$	2.09	&	19.84	$\pm$	2.19	&	19.12	$\pm$	2.4	&	17.34	$\pm$	2.67	\\	
Yeast$_1$	&	15.84	$\pm$	2.21	&	40.66	$\pm$	11.6	&	47.44	$\pm$	13.19	&	51.72	$\pm$	9.94	&	50.28	$\pm$	8.85	&	46.5	$\pm$	10.63	\\	
Yeast$_2$	&	8.56	$\pm$	1.7	&	15.18	$\pm$	6.71	&	20.24	$\pm$	6.87	&	24.34	$\pm$	4.76	&	22.46	$\pm$	4.25	&	20.84	$\pm$	4.63	\\	
Abal$_1$	&	8.7	$\pm$	1.58	&	10.98	$\pm$	1.8	&	10.56	$\pm$	2.09	&	10.02	$\pm$	1.87	&	9.58	$\pm$	1.47	&	9.12	$\pm$	1.77	\\	
Abal$_2$	&	8.76	$\pm$	1.51	&	6.68	$\pm$	1.77	&	6.18	$\pm$	1.44	&	5.7	$\pm$	1.83	&	5.98	$\pm$	1.73	&	7.08	$\pm$	2.59	\\

  \hline
  & \multicolumn{6}{c}{\textsf{NSGA-II SDO 0.5}} \\ \hline
  
Ion	&	11.32	$\pm$	2.3	&	12.86	$\pm$	2.01	&	12.2	$\pm$	2.02	&	11.3	$\pm$	1.58	&	11.34	$\pm$	1.93	&	10.16	$\pm$	1.5	\\
Spect	&	13.78	$\pm$	2.14	&	21.96	$\pm$	2.13	&	21.42	$\pm$	2	&	20.84	$\pm$	2.28	&	19.78	$\pm$	2.12	&	19.2	$\pm$	1.86	\\
Yeast$_1$	&	15.84	$\pm$	2.21	&	46.24	$\pm$	5.82	&	54.98	$\pm$	4.65	&	54.9	$\pm$	5.34	&	54.56	$\pm$	7.1	&	52.3	$\pm$	6.99	\\
Yeast$_2$	&	8.56	$\pm$	1.7	&	18.48	$\pm$	3.51	&	23.82	$\pm$	2.96	&	25.18	$\pm$	2.57	&	25.22	$\pm$	2.94	&	24.4	$\pm$	3.2	\\
Abal$_1$	&	8.7	$\pm$	1.58	&	11.46	$\pm$	1.5	&	11.2	$\pm$	1.7	&	11.04	$\pm$	1.28	&	10.42	$\pm$	1.47	&	10.14	$\pm$	1.57	\\
Abal$_2$	&	8.76	$\pm$	1.51	&	13.86	$\pm$	1.85	&	13.86	$\pm$	2.89	&	10.88	$\pm$	4.02	&	9.54	$\pm$	4.3	&	9.54	$\pm$	4.79	\\

    \hline
   \end{tabular}
  }

\label{tab:uniq:nsgaii}
%\end{table*}

%\begin{table*}
  \caption{Average number of distinct solutions ($\pm$ std. deviation) occurring at the first front of generations 1, 10, 20, 20, 40 and 50 for canonical SPEA2 and SPEA2 SDO methods.}
 
  \centering
   \resizebox{1.0\textwidth}{!}{
  \begin{tabular}{c|cccccc}\hline

   & \multicolumn{6}{c}{\textsf{Generation}} \\\hline \hline
  Data & 1 & 10 & 20  & 30 & 40 & 50 \\
  set & mean $\pm$ std & mean $\pm$ std & mean $\pm$ std & mean $\pm$ std & mean $\pm$ std  & mean $\pm$ std \\ \hline
 &  \multicolumn{6}{c}{\textsf{SPEA2}} \\ \hline
 
Ion	&	11.32	$\pm$	2.3	&	10.6	$\pm$	3.45	&	9.94	$\pm$	3.32	&	9.42	$\pm$	2.89	&	8.78	$\pm$	2.94	&	8.46	$\pm$	2.72	\\
Spect	&	13.78	$\pm$	2.14	&	21.6	$\pm$	2.93	&	20.88	$\pm$	2.31	&	19.56	$\pm$	2.05	&	18.76	$\pm$	2.21	&	17.92	$\pm$	2.35	\\
Yeast$_1$	&	15.84	$\pm$	2.21	&	41.82	$\pm$	12.01	&	48.42	$\pm$	11.65	&	46.18	$\pm$	14.32	&	46.14	$\pm$	15	&	47.12	$\pm$	11.44	\\
Yeast$_2$	&	8.56	$\pm$	1.7	&	15.58	$\pm$	5.96	&	21.7	$\pm$	5.73	&	23.06	$\pm$	4.65	&	22.32	$\pm$	3.85	&	20.7	$\pm$	5.32	\\
Abal$_1$	&	8.7	$\pm$	1.58	&	11	$\pm$	1.88	&	10.86	$\pm$	1.96	&	10.14	$\pm$	1.73	&	9.38	$\pm$	1.87	&	9.08	$\pm$	1.59	\\
Abal$_2$	&	8.76	$\pm$	1.51	&	6.8	$\pm$	2.02	&	6.76	$\pm$	2.14	&	6.98	$\pm$	3.08	&	6.9	$\pm$	2.73	&	6.3	$\pm$	2.32	\\

  \hline
  & \multicolumn{6}{c}{\textsf{SPEA2 SDO 0.5}} \\ \hline
 Ion	&	11.32	$\pm$	2.3	&	13.24	$\pm$	2.58	&	12.3	$\pm$	2.22	&	11.2	$\pm$	2.17	&	11.02	$\pm$	1.95	&	10.18	$\pm$	1.78	\\
Spect	&	13.78	$\pm$	2.14	&	22.1	$\pm$	2.31	&	21.44	$\pm$	2.52	&	21.34	$\pm$	2.06	&	20.46	$\pm$	1.89	&	20.12	$\pm$	1.85	\\
Yeast$_1$	&	15.84	$\pm$	2.21	&	46.52	$\pm$	4.83	&	55.52	$\pm$	4.48	&	56.96	$\pm$	5.11	&	55.78	$\pm$	5.1	&	54.48	$\pm$	5.64	\\
Yeast$_2$	&	8.56	$\pm$	1.7	&	18.14	$\pm$	3.34	&	23.94	$\pm$	3.11	&	25.66	$\pm$	2.38	&	25.6	$\pm$	2.47	&	25.36	$\pm$	2.88	\\
Abal$_1$	&	8.7	$\pm$	1.58	&	11.28	$\pm$	1.41	&	11.44	$\pm$	1.28	&	10.98	$\pm$	1.29	&	10.16	$\pm$	1.61	&	9.92	$\pm$	1.54	\\
Abal$_2$	&	8.76	$\pm$	1.51	&	14.22	$\pm$	1.87	&	13.28	$\pm$	3.15	&	11.08	$\pm$	4.52	&	10.5	$\pm$	4.52	&	9.62	$\pm$	4.53	\\

    \hline
   \end{tabular}
  }
\label{tab:uniq:spea2}
\end{table*}

\subsection{Discussion on results and why SDO works}
\label{sec:sdo_discuss}

Table~\ref{tab:uniq:nsgaii} and~\ref{tab:uniq:spea2} show the average number of distinct solutions, which are defined as solutions that exclude duplicates, across all 50 independent runs, in the first approximated Pareto front for Generations 1, 10, 20, 30, 40 and 50 for the canonical EMO methods (top of Tables~\ref{tab:uniq:nsgaii} and~\ref{tab:uniq:spea2}) and the and SDO methods (bottom part of these tables). For simplicity in our analysis, we use a single upper bound of 0.5 for the SDO method. By selecting the pivot as an individual from the sparsest region of the first front, we hope to attract new individuals to the surrounding region of sparsity, the concept of which has been discussed in greater detail in Section~\ref{sec:sdo} and depicted in Figure~\ref{fig:SDO_aproach}. As a consequence of attracting new individuals to these sparse regions, it is expected that the number of  solutions should increase in the first approximated Pareto front for subsequent generations. This is precisely what it is observed in Tables~\ref{tab:uniq:nsgaii} and~\ref{tab:uniq:spea2}, although it is fair to say that the increase in the numbers of individuals in the best Pareto front is minimal going, for instance, from 8.28 distinct solutions in the last generation when using the Ion dataset and NSGA-II to 10.16 distinct solutions when using NSGA-II SDO (top-right column of Table~\ref{tab:uniq:nsgaii}). Even when this increase is small, the trend is consistent for all the datasets used in this work, regardless of using NSGA-II SDO or SPEA2 SDO. This small increase in the number of distinct solutions partially explain why our proposed semantic-based method yields better results compared with their canonical EMO algorithms. As we have articulated in Section~\ref{sec:related}, multiple studies have reported the benefits of promoting semantics in evolutionary search leading to have diversity. If the same is true for our proposed semantic-based approach, dubbed ``Semantic-based Distance as an additional criteriOn" (SDO for short), then we believe that a reduction in the number of duplicated solutions should be observed. To verify and help us further understand why our method yields better results, we focus on this next.

%\textcolor{blue}{If we look at Tables~\ref{tab:freq:nsgaii} and~\ref{tab:freq:spea2}, when comparing the final approximated Pareto fronts of each method, we see a slight increase in the number of unique solution for the semantic method in contrast to the canonical method, with a typical range of 1 to 5 additional individuals being observed in the final generation. This slight increase may not be enough to warrant an increase in performance alone.}

 Tables~\ref{tab:freq:nsgaii} and~\ref{tab:freq:spea2} show the number of duplicated solutions, averaged over 50 independent runs, in multiple generations (1, 10, 20, 30, 40 and 50). We can see that the  number of duplicates substantially decreases for SDO in each generation for multiple datasets. For example, a fourth of the number of duplicated solutions are reported when using the Ion dataset and NSGA-II SDO compared to NSGA-II (right column in Table~\ref{tab:freq:nsgaii}). The same is observed in other datasets such as the Spect and Yeast$_2$, where the number of duplicates for the semantic-based method is half compared to the NSGA-II.

\begin{table*}[tbh!]
  \caption{Average number of duplicates ($\pm$ std. deviation) occurring at the first Pareto front of generations 1, 10, 20, 20, 40 and 50 for canonical NSGA-II and NSGA-II SDO methods.}
 
  \centering
   \resizebox{1.0\textwidth}{!}{
  \begin{tabular}{c|cccccc}\hline

   & \multicolumn{6}{c}{\textsf{Generation}} \\\hline \hline
  Data & 1 & 10 & 20  & 30 & 40 & 50 \\
  set & mean $\pm$ std & mean $\pm$ std & mean $\pm$ std & mean $\pm$ std & mean $\pm$ std  & mean $\pm$ std \\ \hline
  & \multicolumn{6}{c}{\textsf{NSGA-II}} \\ \hline
  Ion	&	1.73	$\pm$	2.75	&	29.49	$\pm$	104.89	&	37.08	$\pm$	117.41	&	36.18	$\pm$	113.71	&	35.36	$\pm$	106.18	&	43.58	$\pm$	114.89	\\					
Spect	&	1.10	$\pm$	0.55	&	2.51	$\pm$	3.53	&	4.31	$\pm$	10.14	&	7.09	$\pm$	20.36	&	11.30	$\pm$	34.19	&	15.11	$\pm$	44.10	\\					
Yeast$_1$	&	2.00	$\pm$	3.72	&	5.09	$\pm$	32.11	&	6.07	$\pm$	31.44	&	6.73	$\pm$	23.29	&	8.47	$\pm$	26.10	&	9.77	$\pm$	31.91	\\					
Yeast$_2$	&	3.03	$\pm$	5.67	&	12.49	$\pm$	68.64	&	6.88	$\pm$	37.38	&	8.62	$\pm$	31.64	&	14.65	$\pm$	48.69	&	19.27	$\pm$	61.88	\\					
Abal$_1$	&	1.15	$\pm$	0.46	&	12.23	$\pm$	56.19	&	25.61	$\pm$	78.06	&	33.68	$\pm$	88.88	&	43.86	$\pm$	106.38	&	46.55	$\pm$	111.15	\\					
Abal$_2$	&	5.39	$\pm$	12.21	&	74.85	$\pm$	174.41	&	79.33	$\pm$	180.00	&	87.72	$\pm$	187.83	&	80.33	$\pm$	181.22	&	64.54	$\pm$	164.32	\\	\hline

  & \multicolumn{6}{c}{\textsf{NSGA-II SDO 0.5}} \\ \hline
 
  Ion	&	1.73	$\pm$	2.75	&	5.70	$\pm$	25.70	&	6.53	$\pm$	33.30	&	9.13	$\pm$	36.99	&	10.40	$\pm$	37.92	&	11.49	$\pm$	40.81	\\
Spect	&	1.10	$\pm$	0.55	&	2.33	$\pm$	2.64	&	3.47	$\pm$	5.48	&	5.24	$\pm$	11.63	&	6.59	$\pm$	12.26	&	7.53	$\pm$	14.85	\\
Yeast$_1$	&	2.00	$\pm$	3.72	&	2.29	$\pm$	3.56	&	3.30	$\pm$	5.37	&	4.70	$\pm$	12.37	&	5.86	$\pm$	16.10	&	6.89	$\pm$	19.75	\\
Yeast$_2$	&	3.03	$\pm$	5.67	&	2.07	$\pm$	3.32	&	3.35	$\pm$	9.78	&	5.45	$\pm$	17.62	&	8.12	$\pm$	23.76	&	10.98	$\pm$	27.64	\\
Abal$_1$	&	1.15	$\pm$	0.46	&	4.46	$\pm$	14.61	&	8.73	$\pm$	36.10	&	13.59	$\pm$	39.66	&	23.00	$\pm$	59.09	&	29.88	$\pm$	74.32	\\
Abal$_2$	&	5.39	$\pm$	12.21	&	9.87	$\pm$	39.35	&	19.20	$\pm$	78.21	&	34.11	$\pm$	114.74	&	39.46	$\pm$	123.19	&	39.61	$\pm$	120.37	\\ \hline

   \end{tabular}
  }
\label{tab:freq:nsgaii}
%\end{table*}

%\begin{table*}
  \caption{Average number of duplicates ($\pm$ std. deviation) occurring at the first Pareto front of generations 1, 10, 20, 20, 40 and 50 for canonical SPEA2 and SPEA2 SDO methods.}
 
  \centering
   \resizebox{1.0\textwidth}{!}{
  \begin{tabular}{c|cccccc}\hline

   & \multicolumn{6}{c}{\textsf{Generation}} \\\hline \hline
  Data & 1 & 10 & 20  & 30 & 40 & 50 \\
  set & mean $\pm$ std & mean $\pm$ std & mean $\pm$ std & mean $\pm$ std & mean $\pm$ std  & mean $\pm$ std \\ \hline
  & \multicolumn{6}{c}{\textsf{SPEA2}} \\ \hline
 		
Ion	&	1.73	$\pm$	2.75	&	24.26	$\pm$	90.64	&	28.76	$\pm$	92.82	&	33.69	$\pm$	101.81	&	38.58	$\pm$	109.38	&	42.00	$\pm$	109.99	\\					
Spect	&	1.10	$\pm$	0.55	&	2.23	$\pm$	2.84	&	4.14	$\pm$	8.53	&	7.13	$\pm$	25.50	&	10.93	$\pm$	35.49	&	17.01	$\pm$	46.47	\\					
Yeast$_1$	&	2.00	$\pm$	3.72	&	5.13	$\pm$	32.33	&	6.31	$\pm$	30.91	&	8.29	$\pm$	37.45	&	8.69	$\pm$	32.28	&	8.51	$\pm$	27.74	\\					
Yeast$_2$	&	3.03	$\pm$	5.67	&	8.24	$\pm$	54.33	&	6.07	$\pm$	29.75	&	11.18	$\pm$	41.72	&	16.48	$\pm$	54.99	&	19.68	$\pm$	64.16	\\					
Abal$_1$	&	1.15	$\pm$	0.46	&	9.39	$\pm$	45.68	&	16.60	$\pm$	67.21	&	35.25	$\pm$	93.86	&	46.58	$\pm$	110.93	&	48.44	$\pm$	116.32	\\					
Abal$_2$	&	5.39	$\pm$	12.21	&	72.51	$\pm$	172.07	&	68.04	$\pm$	167.64	&	68.05	$\pm$	167.39	&	70.25	$\pm$	169.38	&	76.40	$\pm$	176.26	\\	\hline

  & \multicolumn{6}{c}{\textsf{SPEA2 SDO 0.5}} \\ \hline
 
 Ion	&	1.73	$\pm$	2.75	&	4.60	$\pm$	20.22	&	8.21	$\pm$	29.47	&	10.38	$\pm$	38.35	&	13.95	$\pm$	48.19	&	15.54	$\pm$	53.57	\\
Spect	&	1.10	$\pm$	0.55	&	2.10	$\pm$	2.26	&	3.33	$\pm$	4.11	&	4.64	$\pm$	6.97	&	5.85	$\pm$	8.66	&	7.25	$\pm$	10.74	\\
Yeast$_1$	&	2.00	$\pm$	3.72	&	2.17	$\pm$	3.16	&	3.32	$\pm$	5.09	&	4.79	$\pm$	11.62	&	5.21	$\pm$	10.52	&	6.52	$\pm$	16.93	\\
Yeast$_2$	&	3.03	$\pm$	5.67	&	2.34	$\pm$	6.28	&	2.91	$\pm$	4.72	&	4.32	$\pm$	6.68	&	6.68	$\pm$	15.05	&	8.01	$\pm$	17.53	\\
Abal$_1$	&	1.15	$\pm$	0.46	&	4.43	$\pm$	15.34	&	8.32	$\pm$	28.41	&	14.34	$\pm$	42.79	&	20.03	$\pm$	52.82	&	27.40	$\pm$	71.82	\\
Abal$_2$	&	5.39	$\pm$	12.21	&	9.24	$\pm$	33.99	&	24.37	$\pm$	93.08	&	33.17	$\pm$	111.46	&	37.14	$\pm$	118.02	&	41.74	$\pm$	125.53	\\ \hline

   \end{tabular}
  }
\label{tab:freq:spea2}
\end{table*}
 When we turn our attention to Table~\ref{tab:freq:spea2}, reporting the duplicates when using the SPEA2 algorithm and our proposed SDO, we can observe the same trend as before: the number of duplicates is drastically reduced in the semantic-based method compared to the canonical SPEA2 in all the datasets used in this study, except in the Yeast$_1$ dataset, where a marginally reduction is observed for the semantic-based method going from  8.51 (canonical SPEA2) to 6.52. It is important to note that our proposed SDO reduces the number of duplicates thanks to its mechanism to promote diversity through the use of another objective to be optimised, rather than having an explicit mechanism, for example, to penalise individuals already present in the population with the hope to eliminate this undesired effect during evolution.
 
\begin{figure*}[tbh!]
  \centering
  \begin{tabular}{cc}\\
     \multicolumn{2}{c}{Ion}\\
     \hspace{-0.82cm}  
     \includegraphics[width=0.55\textwidth]{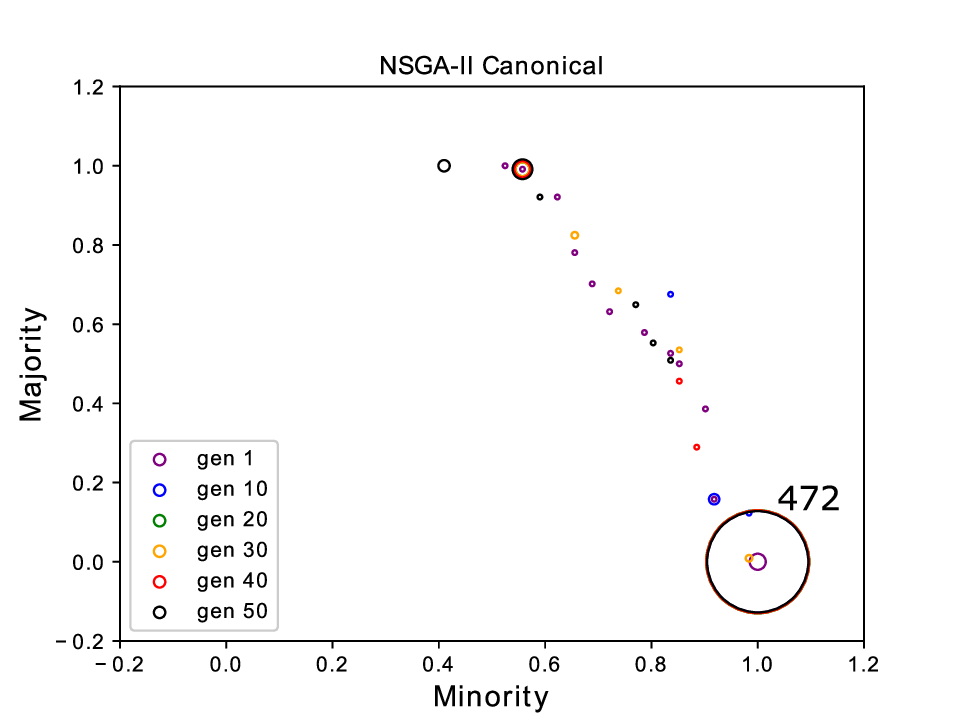}   
     & \hspace{-0.95cm}  
     \includegraphics[width=0.55\textwidth]{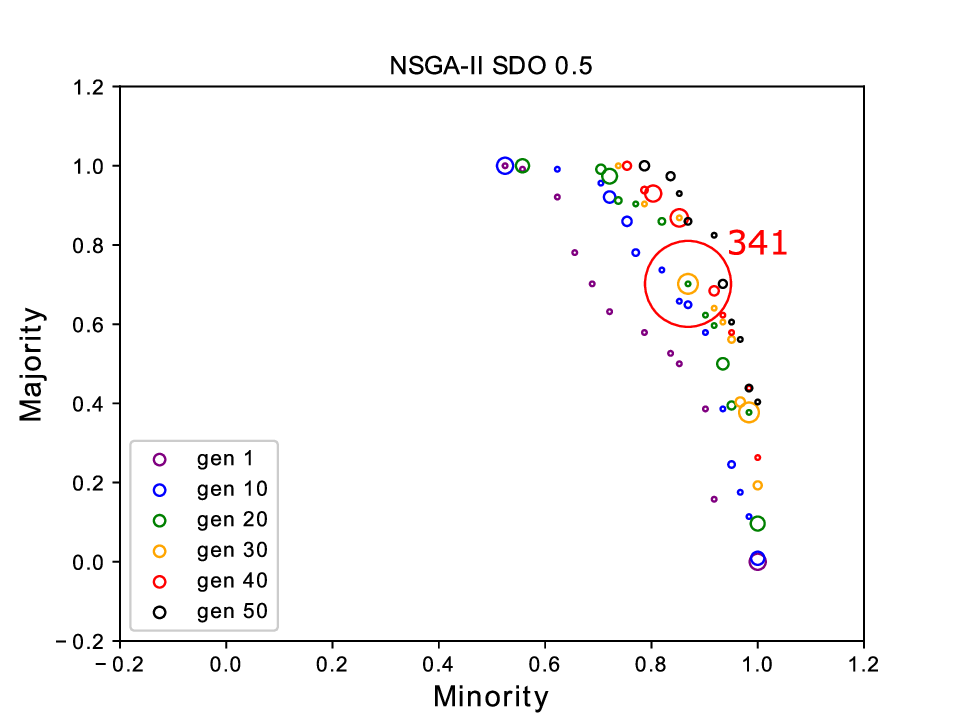}\\
     \multicolumn{2}{c}{Spect}\\
     \hspace{-0.82cm}  
     \includegraphics[width=0.55\textwidth]{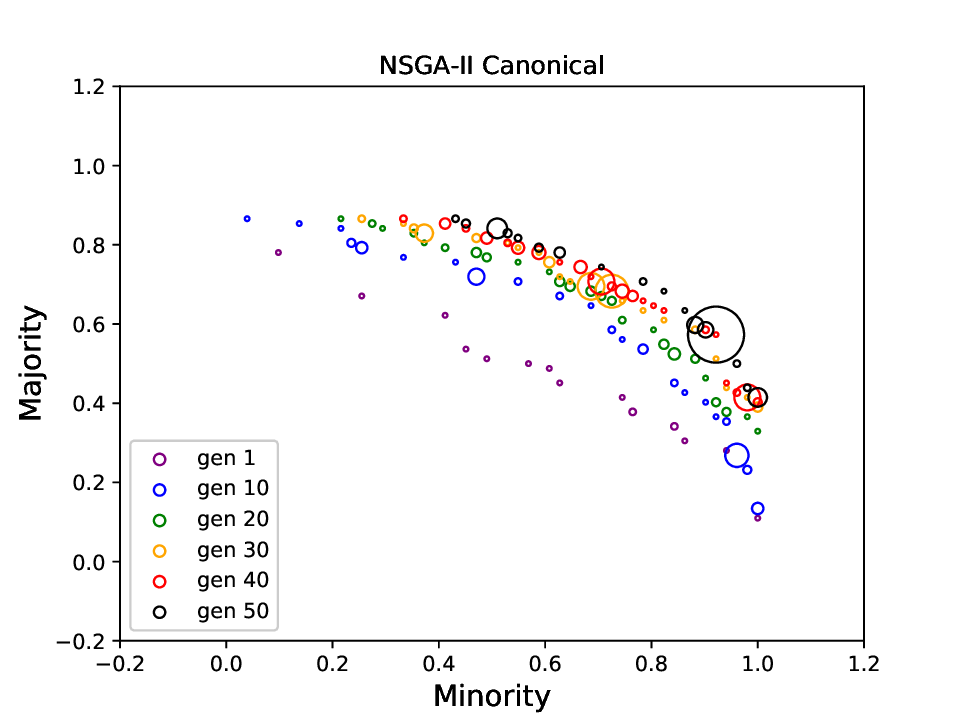}   
     & \hspace{-0.95cm}  
     \includegraphics[width=0.55\textwidth]{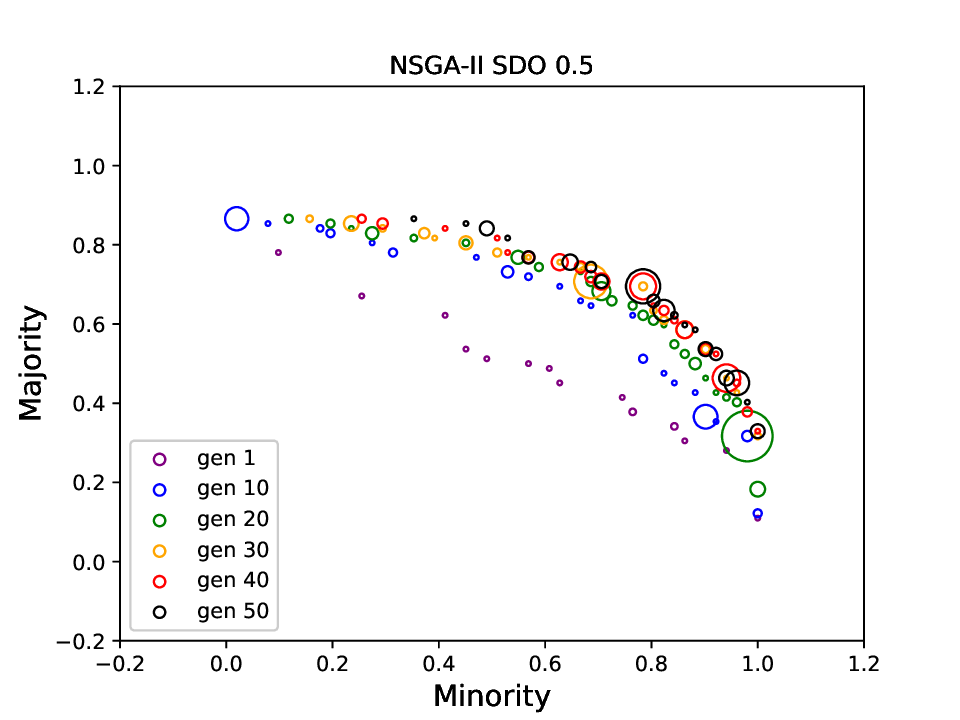} \\
     \multicolumn{2}{c}{Yeast$_1$}\\
    \hspace{-0.82cm}  
     \includegraphics[width=0.55\textwidth]{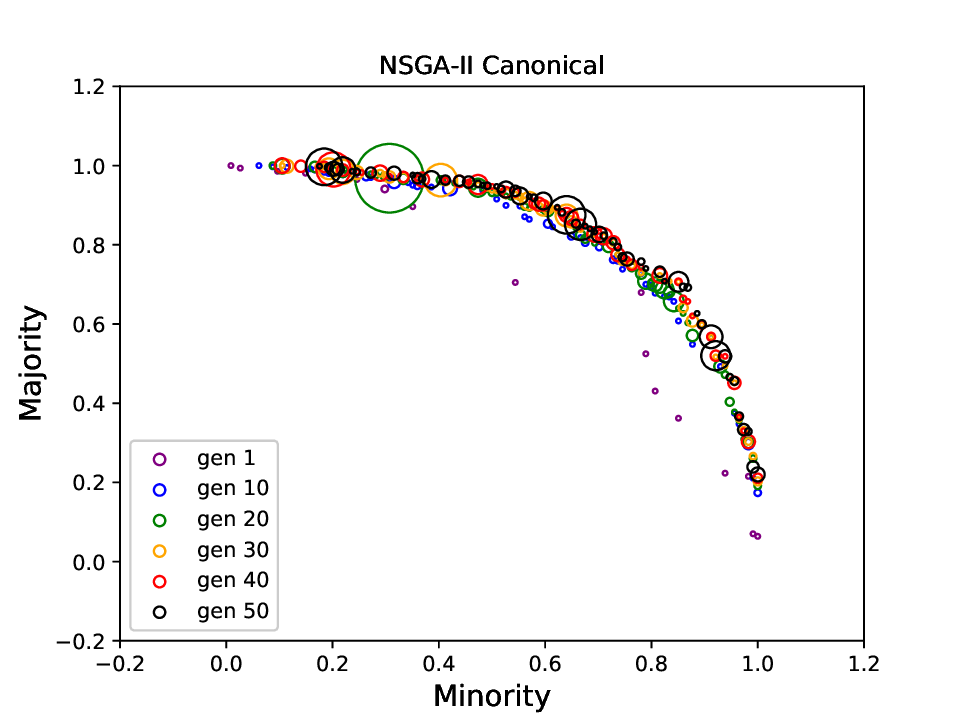}   
     & \hspace{-0.95cm}  
     \includegraphics[width=0.55\textwidth]{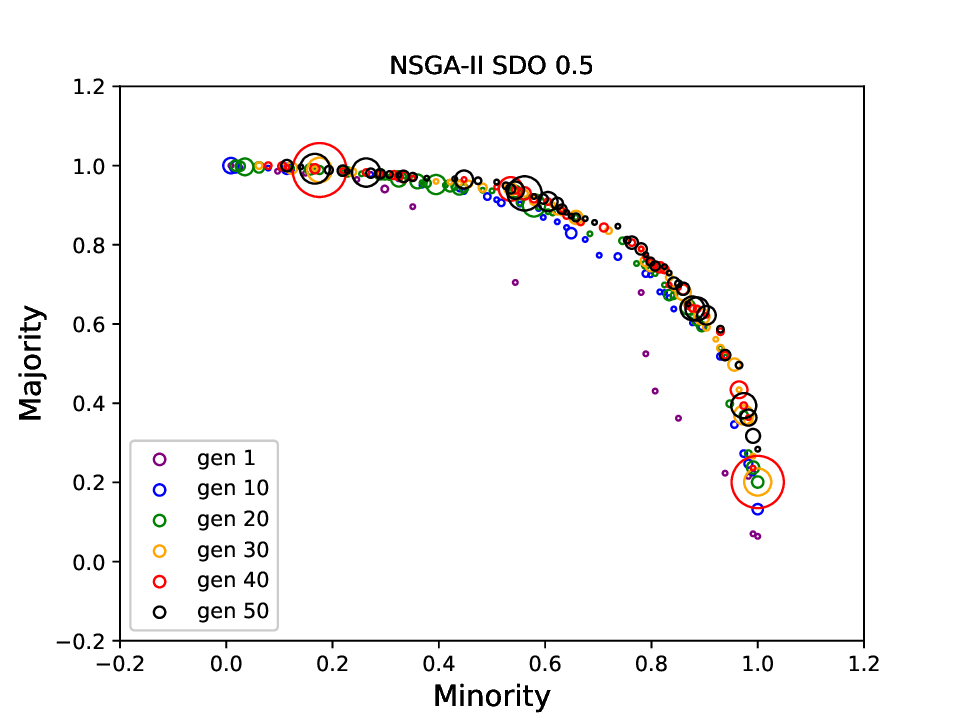} \\

\end{tabular}
\caption{Duplicate frequency of individuals at first Pareto front for Ion, Spect and Yeast$_1$ data set for generation 1, 10, 20, 30, 40 and 50 for NSGA-II and NSGA-II SDO for a single run.} 
\label{fig:freq:nsga-ii_ion_spect_yeast1}
\end{figure*}

\begin{figure*}[tbh!]
  \centering
  \begin{tabular}{cc}\\
     \multicolumn{2}{c}{Yeast$_2$}\\
     \hspace{-0.82cm}  
     \includegraphics[width=0.55\textwidth]{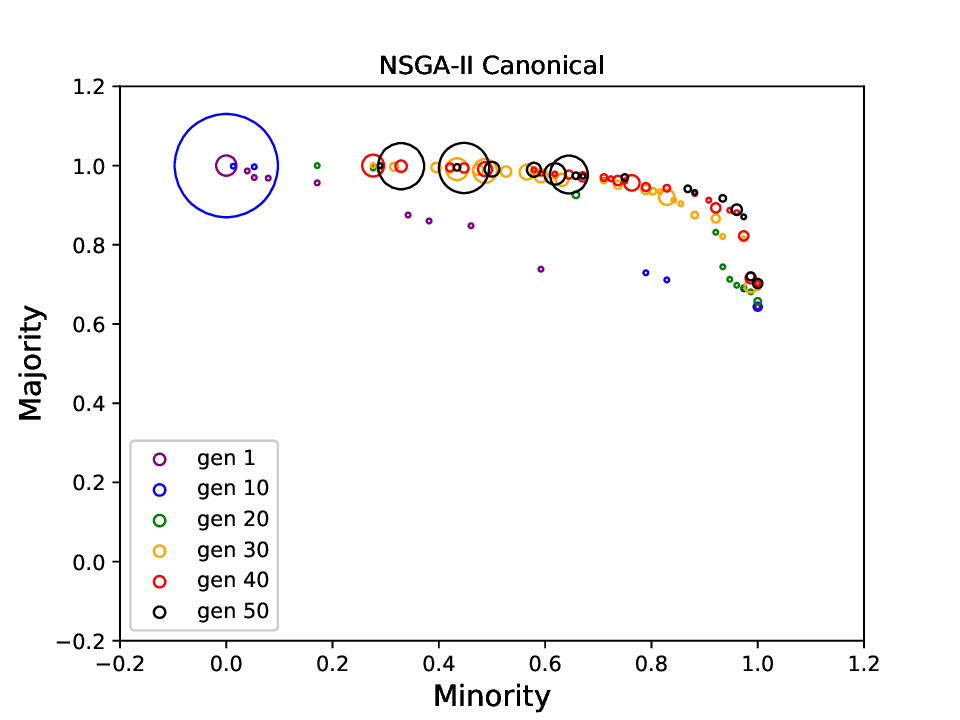}   
     & \hspace{-0.95cm}  
     \includegraphics[width=0.55\textwidth]{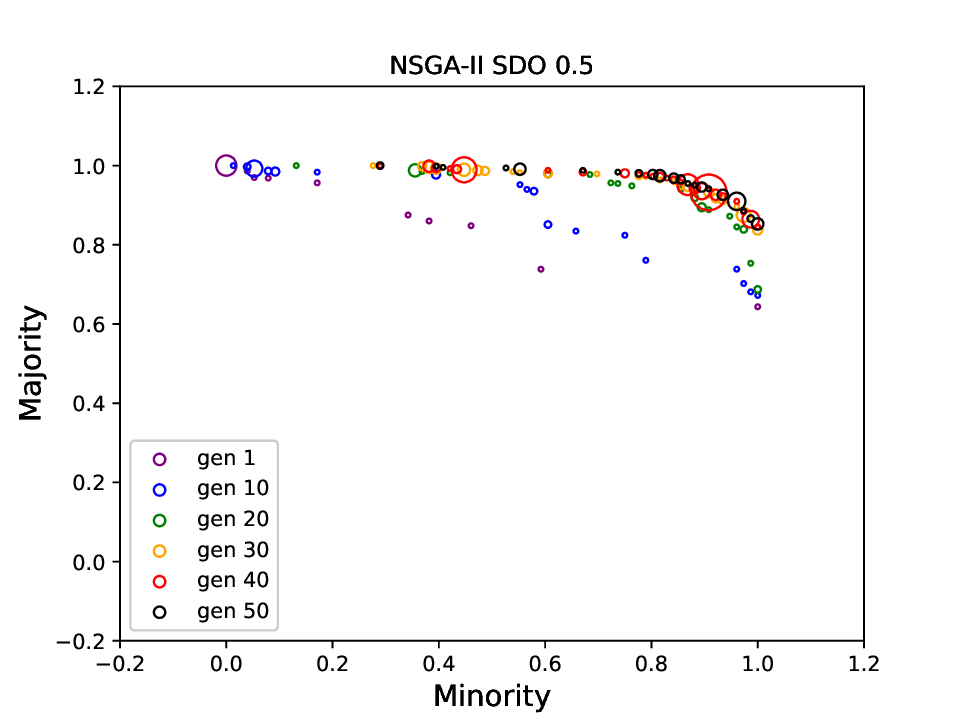}\\
     \multicolumn{2}{c}{Abal$_1$}\\
     \hspace{-0.82cm}  
     \includegraphics[width=0.55\textwidth]{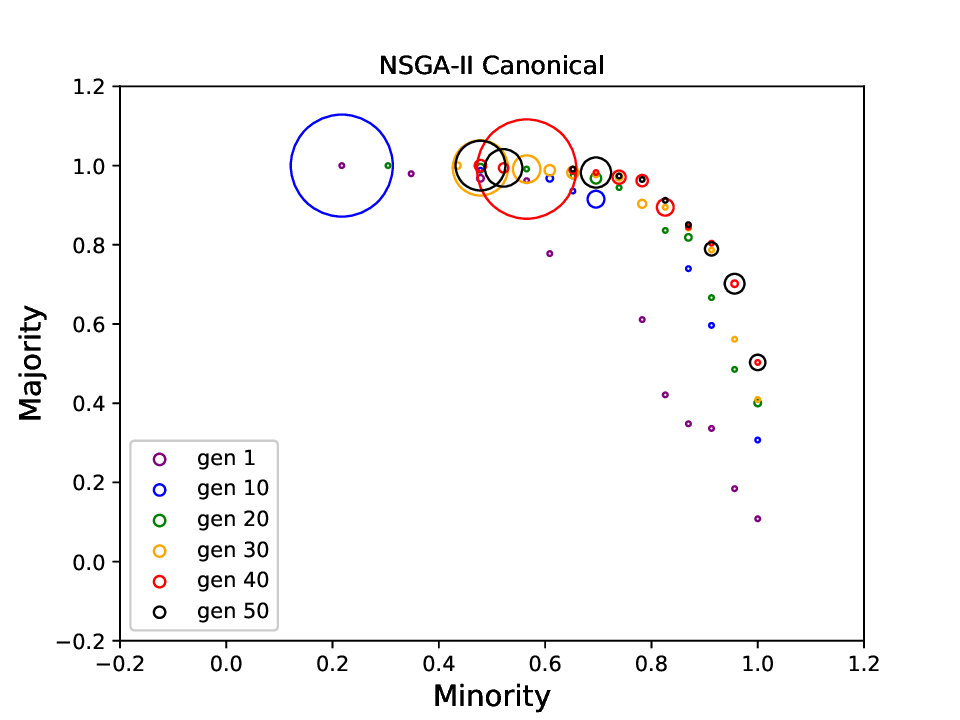}   
     & \hspace{-0.95cm}  
     \includegraphics[width=0.55\textwidth]{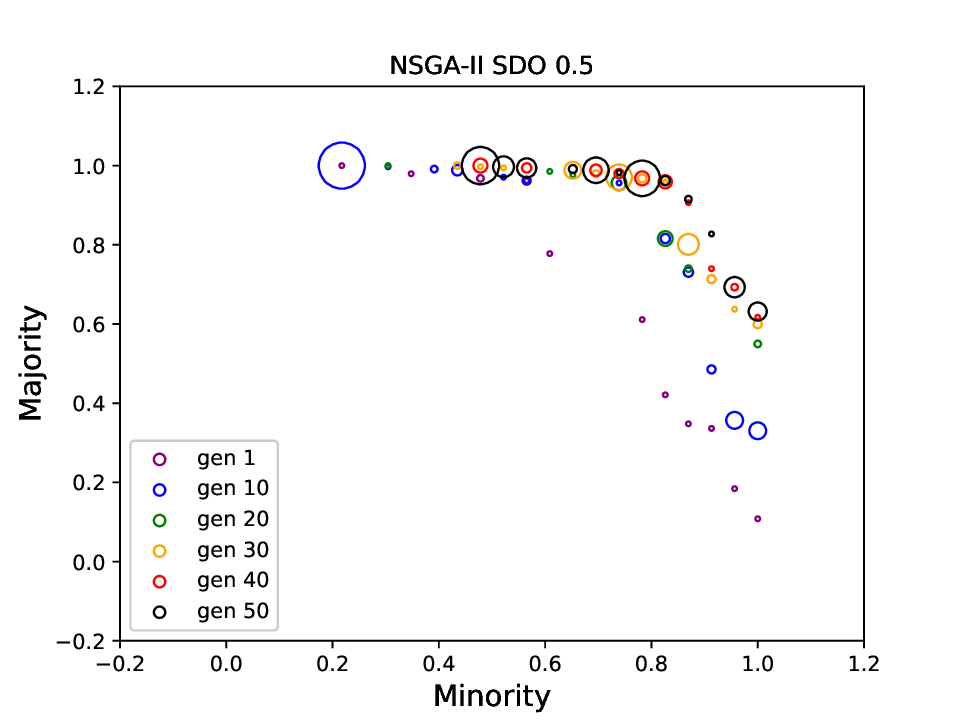} \\
    \multicolumn{2}{c}{Abal$_2$}\\
    \hspace{-0.82cm}  
     \includegraphics[width=0.55\textwidth]{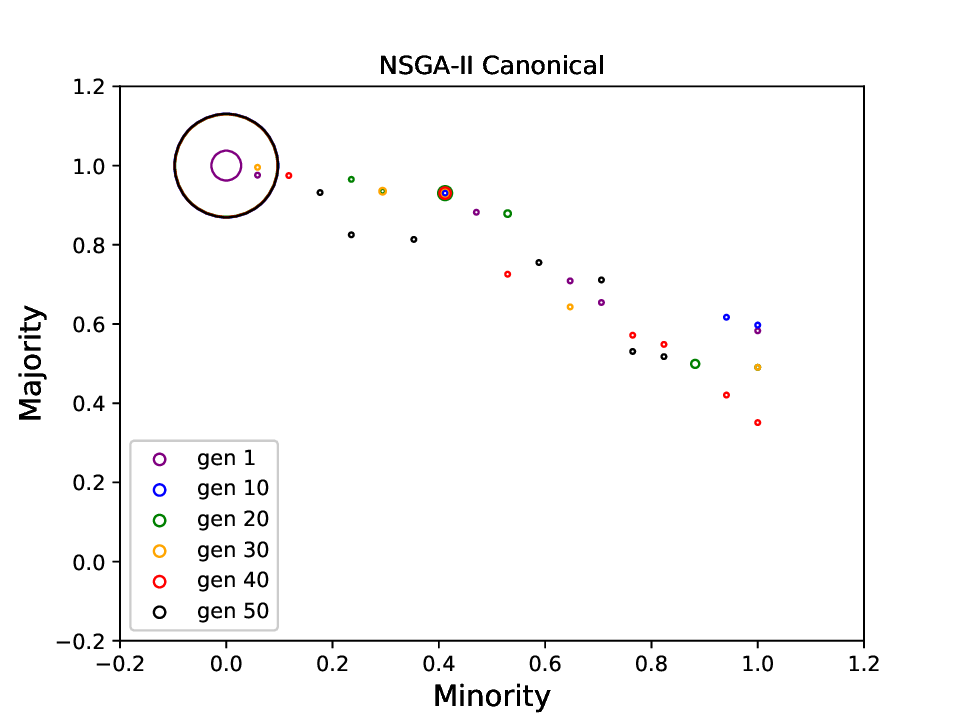}   
     & \hspace{-0.95cm}  
     \includegraphics[width=0.55\textwidth]{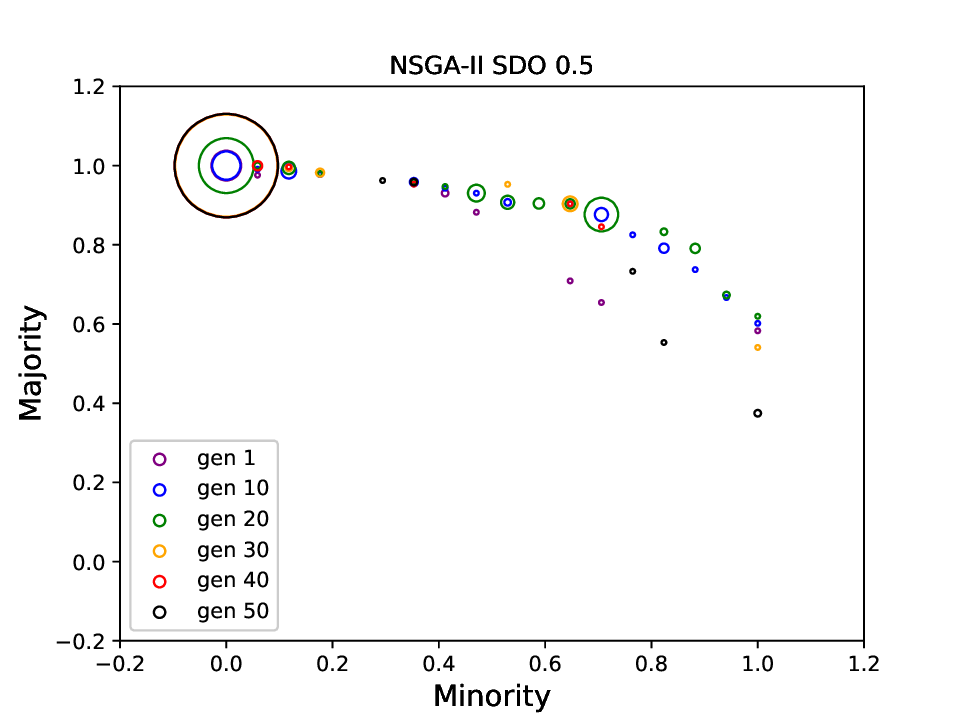} \\

\end{tabular}
  \caption{Duplicate frequency of individuals at first Pareto front for Yeast$_2$, Abal$_1$ and Abal$_2$ data set for generation 1, 10, 20, 30, 40 and 50 for NSGA-II and NSGA-II SDO for a single run.} 
\label{fig:freq:nsga-ii_yeast2_abal1_abal2}
\end{figure*}

\begin{figure*}[tbh!]
  \centering
  \begin{tabular}{cc}\\
     \multicolumn{2}{c}{Ion}\\
     \hspace{-0.82cm}  
     \includegraphics[width=0.55\textwidth]{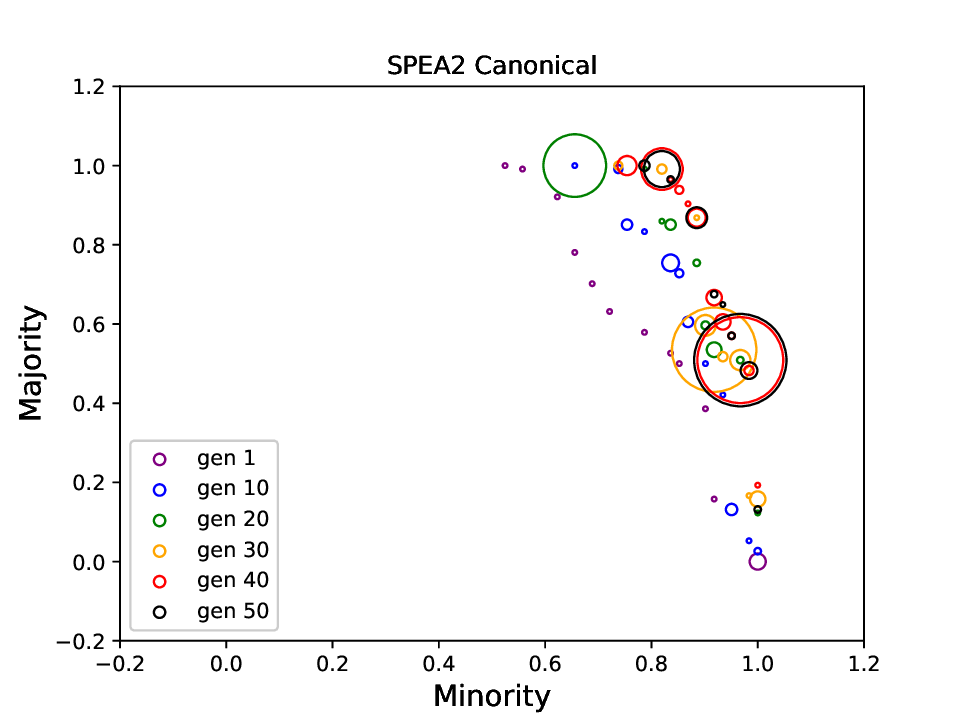}   
     & \hspace{-0.95cm}  
     \includegraphics[width=0.55\textwidth]{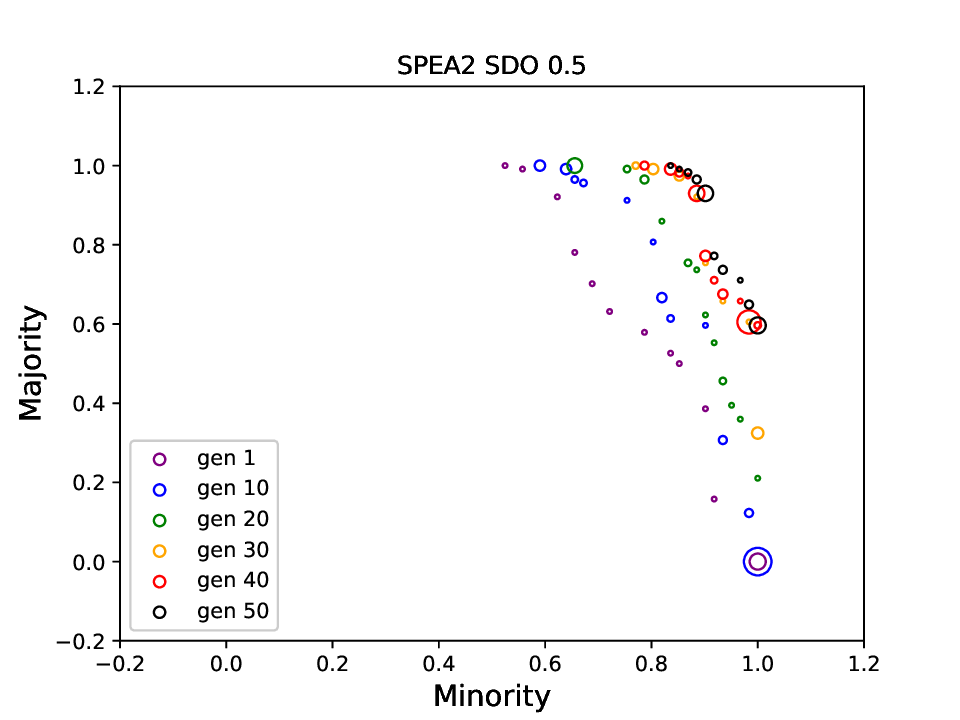}\\
     \multicolumn{2}{c}{Spect}\\
     \hspace{-0.82cm}  
     \includegraphics[width=0.55\textwidth]{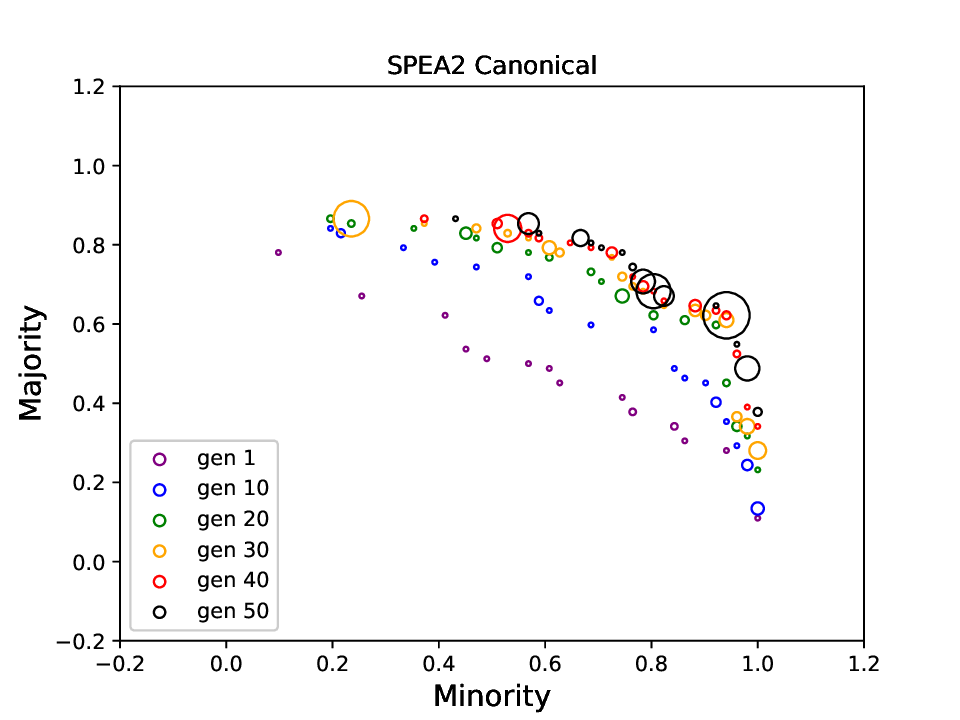}   
     & \hspace{-0.95cm}  
     \includegraphics[width=0.55\textwidth]{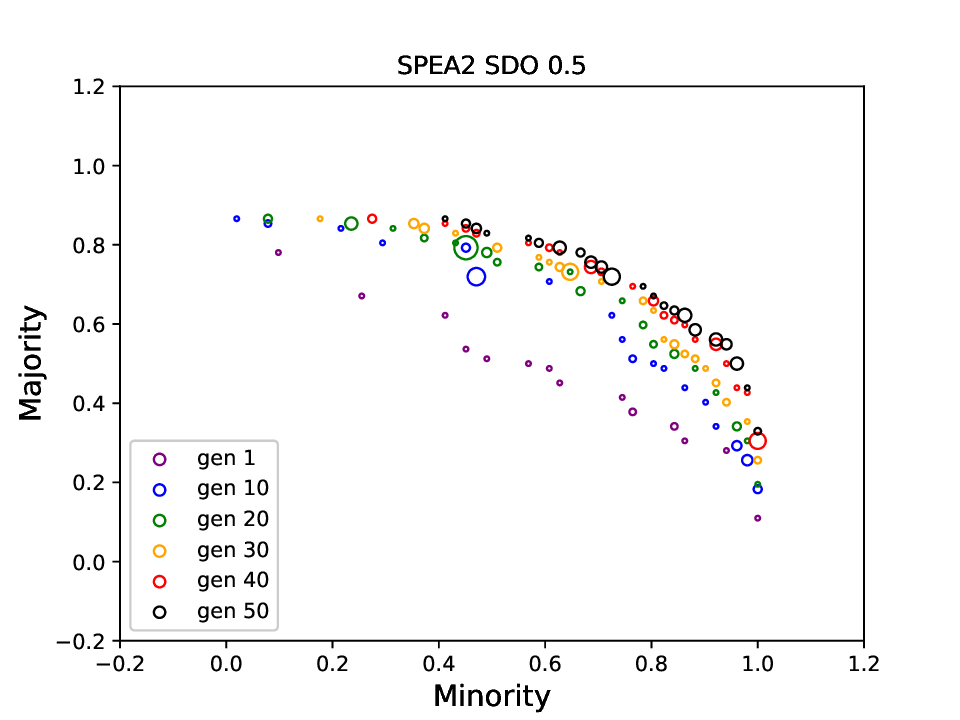} \\
     \multicolumn{2}{c}{Yeast$_1$}\\
    \hspace{-0.82cm}  
     \includegraphics[width=0.55\textwidth]{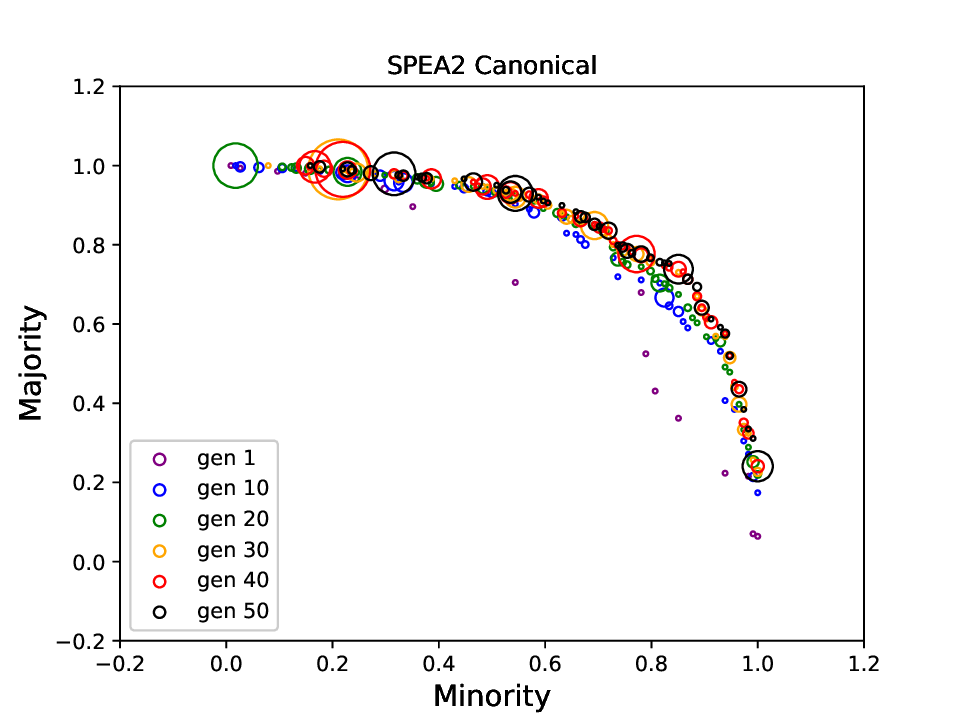}   
     & \hspace{-0.95cm}  
     \includegraphics[width=0.55\textwidth]{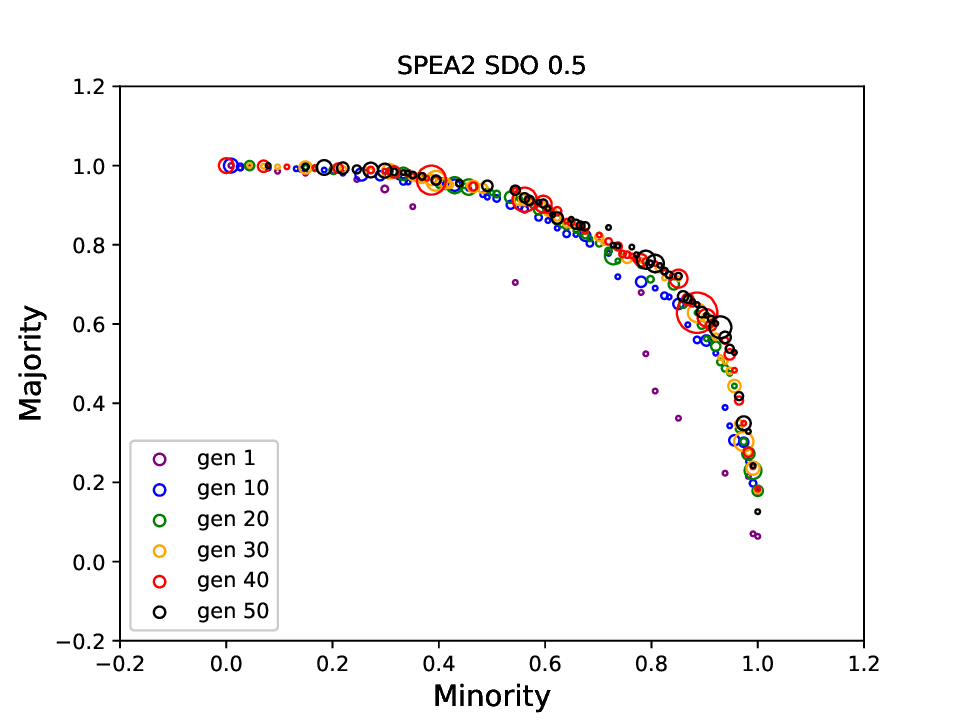} \\

\end{tabular}
\caption{Duplicate frequency of individuals at first Pareto front for Ion, Spect and Yeast$_1$ data set for generation 1, 10, 20, 30, 40 and 50 for SPEA2 and SPEA2 SDO for a single run.} 
\label{fig:freq:spea2_ion_spect_yeast1}
\end{figure*}

\begin{figure*}[tbh!]
  \centering
  \begin{tabular}{cc}\\
     \multicolumn{2}{c}{Yeast$_2$}\\
     \hspace{-0.82cm}  
     \includegraphics[width=0.55\textwidth]{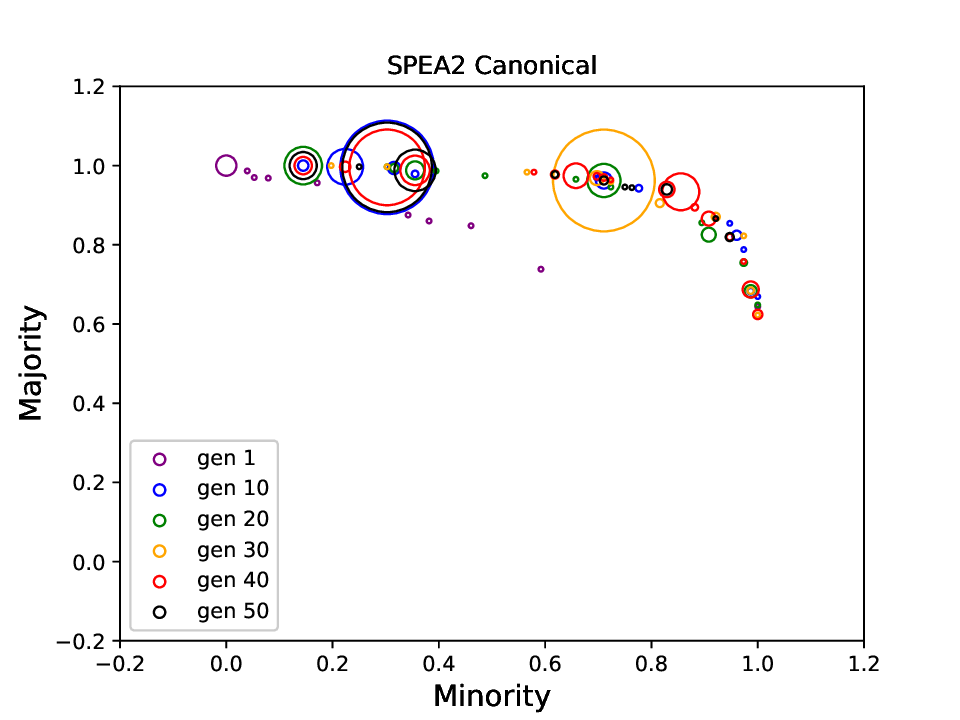}   
     & \hspace{-0.95cm}  
     \includegraphics[width=0.55\textwidth]{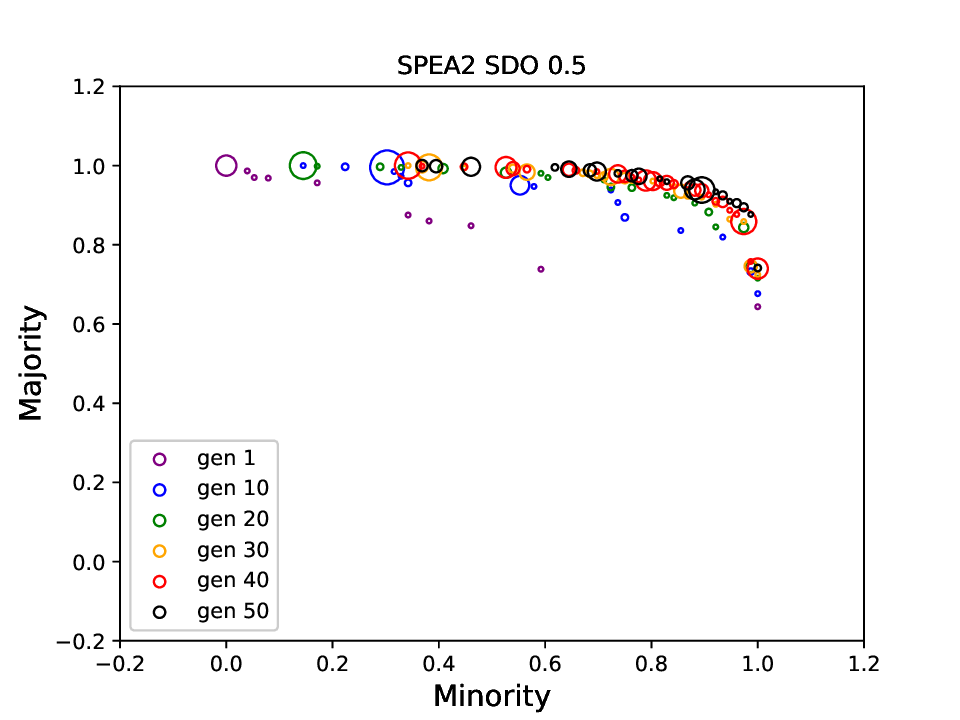}\\
     \multicolumn{2}{c}{Abal$_1$}\\
     \hspace{-0.82cm}  
     \includegraphics[width=0.55\textwidth]{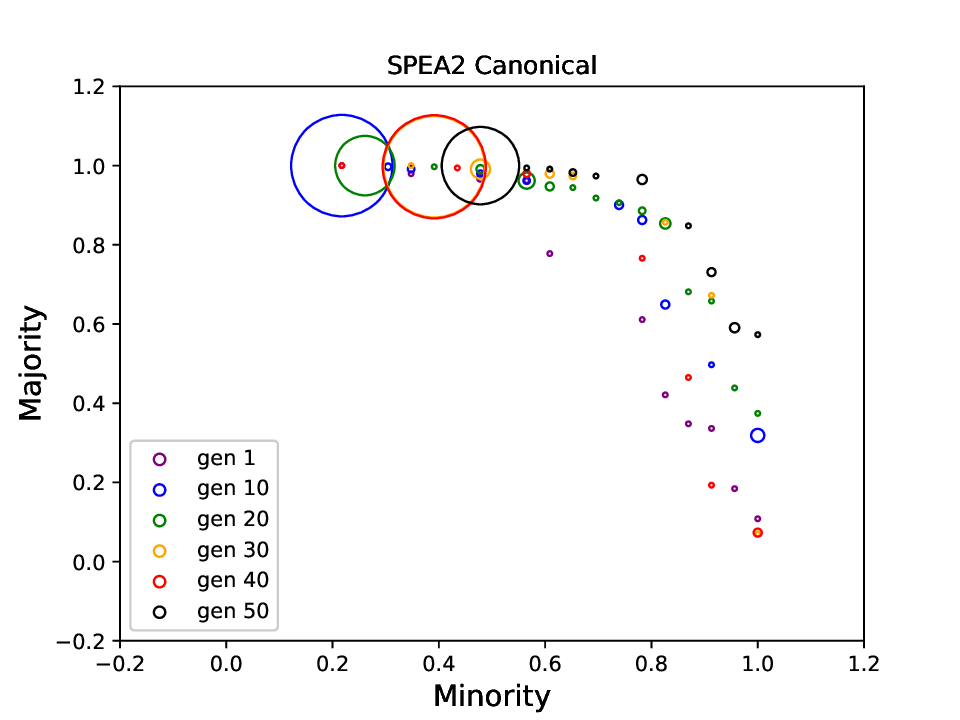}   
     & \hspace{-0.95cm}  
     \includegraphics[width=0.55\textwidth]{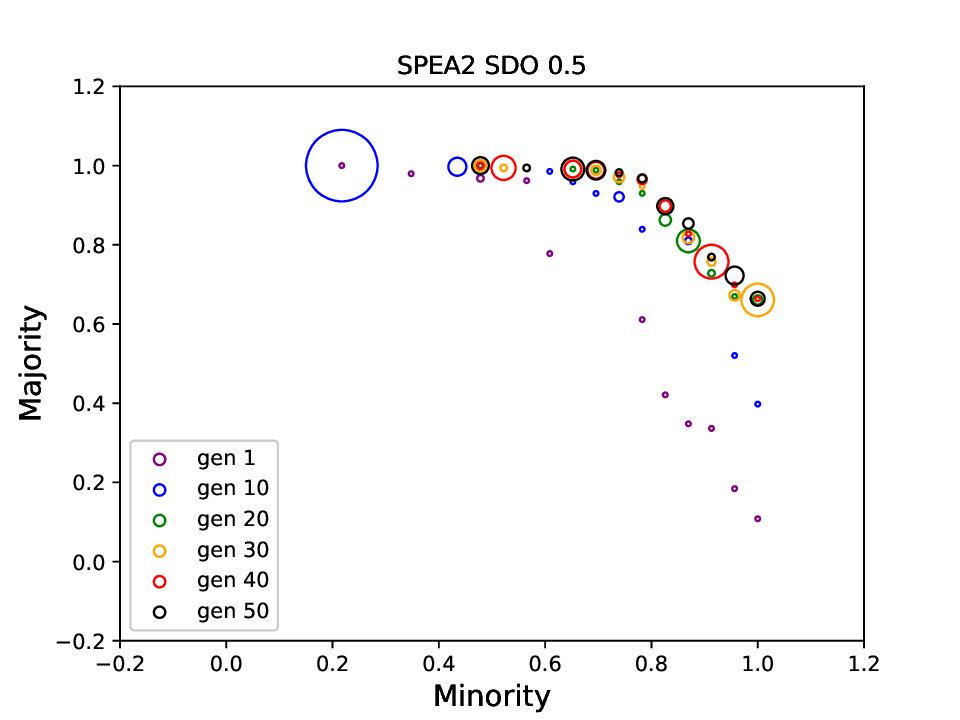} \\
    \multicolumn{2}{c}{Abal$_2$}\\
    \hspace{-0.82cm}  
     \includegraphics[width=0.55\textwidth]{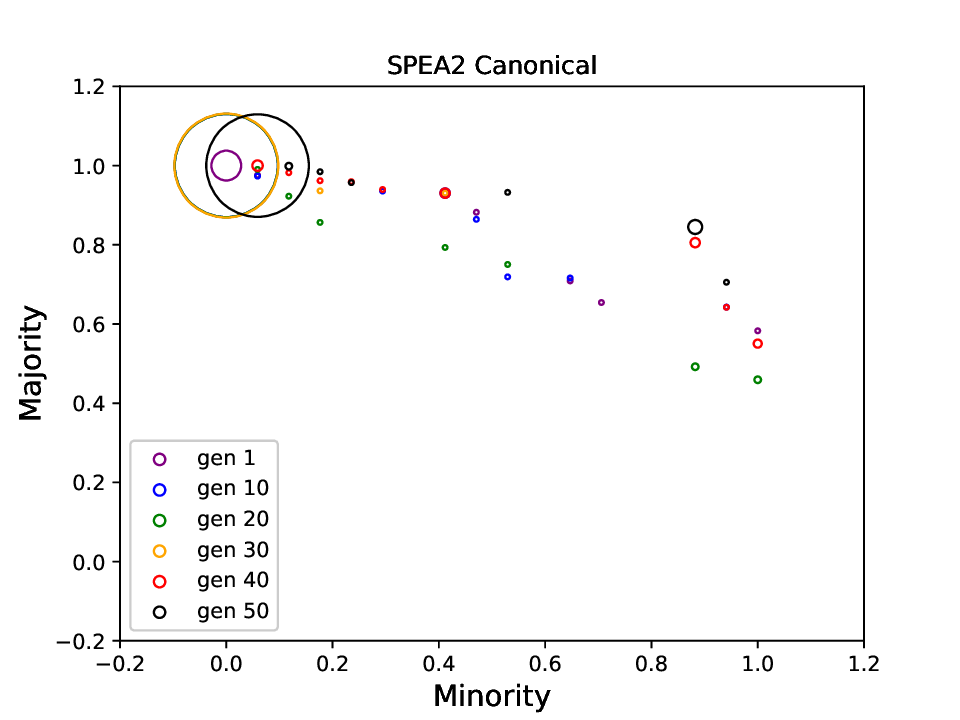}   
     & \hspace{-0.95cm}  
     \includegraphics[width=0.55\textwidth]{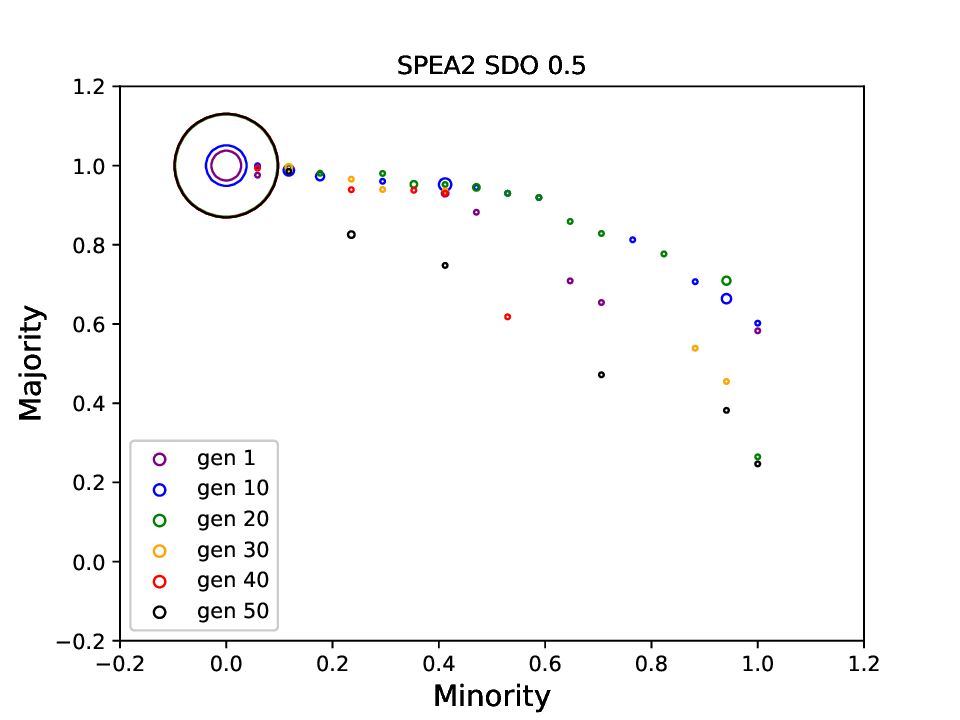} \\

\end{tabular}
  \caption{Duplicate frequency of individuals at first Pareto front for Yeast$_2$, Abal$_1$ and Abal$_2$ data set for generation 1, 10, 20, 30, 40 and 50 for SPEA2 and SPEA2 SDO for a single run.} 
\label{fig:freq:spea2_yeast2_abal1_abal2}
\end{figure*}

 There is one more element worth noting from Tables~\ref{tab:freq:nsgaii} and~\ref{tab:freq:spea2}, which is the standard deviation of duplicated solutions. The standard deviation is notably large when compared to its respective mean, however, again these standard deviations drop significantly for the semantic method compared to the canonical method. These comparatively large standard deviations suggest that while the number of distinct individuals that experience large levels of duplication in the population is somewhat low, there may be a small subset of individuals which experience a level of duplication on an order of magnitude greater than the rest of the population. There are further ramifications of having such large standard deviations in duplication size, which will be discussed in greater detail when further analysing the plots depicted in Figures~\ref{fig:freq:nsga-ii_ion_spect_yeast1},~\ref{fig:freq:nsga-ii_yeast2_abal1_abal2},~\ref{fig:freq:spea2_ion_spect_yeast1} and~\ref{fig:freq:spea2_yeast2_abal1_abal2}.

To explain how the semantic distance-based method is promoting diversity we will briefly return to some of the core functionality of the EMO framework. NSGA-II and SPEA2 are partial ordering methods, that is that when the new population is being formulated, the new population is filled with individuals from the entire approximated fronts first (that is the full Pareto front of a given dominance rank) and subsequently the remainder of the population is filled based on either the crowding distance operation, in the case of canonical methods, or by the semantic crowding-distance for the semantic distance-based methods. An important aspect of this mechanism is that the population size is fixed, therefore removing duplication from the lower dominance rank Pareto fronts allows more individuals from the higher ranked Pareto fronts to be retained in the population at each generation. This in turn leads to a greater spread in the available genetic material for the algorithm to work with, thus promoting diversity. Furthermore, since the pivot is attracting more unique solutions to the first front, this in combination with the decreased duplication of solutions lead to greater overall performance in the semantic distance-based approach.
%\ref{fig:freq:nsga-ii_ion_spect_yeast1}
%\ref{fig:freq:nsga-ii_yeast2_abal1_abal2}
%\ref{fig:freq:spea2_ion_spect_yeast1}
%\ref{fig:freq:spea2_yeast2_abal1_abal2}
 Figures~\ref{fig:freq:nsga-ii_ion_spect_yeast1} and~\ref{fig:freq:nsga-ii_yeast2_abal1_abal2} show solutions from the first approximated Pareto front for NSGA-II and NSGA-II SDO, for the Ion, Spect and Yeast$_1$ data sets and for the Yeast$_2$, Abal$_1$ and Abal$_2$ data sets, respectively, and likewise, Figures~\ref{fig:freq:spea2_ion_spect_yeast1} and~\ref{fig:freq:spea2_yeast2_abal1_abal2} show solutions from the first approximated Pareto front for generations 1, 10, 20, 30, 40 and 50, represented with different coloured hollow circles. All plots represent results from a single seeded run, chosen at random, in order to avoid a biased analysis of the results. The left-hand column and right-hand column of Figures~\ref{fig:freq:nsga-ii_ion_spect_yeast1} to~\ref{fig:freq:spea2_yeast2_abal1_abal2} show the results of the canonical EMO method and the SDP method, respectively.

 The marker size represents the number of candidate solutions found at a particular location in the objective space and thus is an indication of the frequency of duplication. Since the results are for a single run there is some variability when compared to our analysis of Tables~\ref{tab:uniq:nsgaii},~\ref{tab:uniq:spea2},~\ref{tab:freq:nsgaii} and~\ref{tab:freq:spea2}, as these results are based on the average across 50 independent runs, but in general they tend to conform. In particular Yeast$_2$ and Abal$_1$ demonstrate clearly the increase in duplication, with markers of much greater area being observed in the EMO canonical methods.
 % This should move to a new paragraph but \textcolor is preventing
 The Ion and Abal$_2$ data sets for NSGA-II and Abal$_2$ data set for NSGA-II SDO exhibit rather unusual behaviours in terms of the evolution of their respective approximated Pareto fronts. It would be expected that fronts of subsequent generations ought to move towards and ideal region of maximisation, i.e closer to point [1,1], as clearly observed with Spect data. In other words, the expected behaviour is that generation 50 (black hollow circle) is closer to [1,1] as opposed to generation 1 (purple hollow circle) being closer to [1,1], however the is not the case in with the aforementioned data sets. This can again be explained by the duplication of results. In Figure~\ref{fig:freq:nsga-ii_ion_spect_yeast1} for the Ion data set NSGA-II method (top left), the point at [1,0] has a large number of duplicates with 472 solutions at this location, as shown in the plot. However, the total population size is set at 500. Therefore as a result of the large duplication occurring at this point preferable solutions are being `pushed' out of the solution set as the first front is exceeding the population size. This causes the approximate front to gradually recede and leads to a decrease in performance. This behaviour can be implicitly understood from the standard deviation of the duplicates Tables~\ref{tab:freq:nsgaii} and~\ref{tab:freq:spea2}. In particular, both Ion, Abal$_1$ and Abal$_2$ have standard deviations greater than 100 for both NSGA-II and SPEA2. Since the semantic distance-based methods reduce the overall duplication and better promote individuals to the first approximated Pareto front this issue is not as readily observed for these methods but does still occur. For instance, if we look at the semantic approach for Ion NSGA-II SDO (top-right of Figure~\ref{fig:freq:nsga-ii_ion_spect_yeast1}) we see that point [0.93, 0.72] grows to a total number of duplicates of 341 (as shown in the plot) at that location for generation 40 but by generation 50 all of these points have been removed as more recently created candidate solutions now dominate this point. Abal$_2$ was the only one notable data set for the semantic method that failed to tackle this issue adequately (bottom row in Figures~\ref{fig:freq:nsga-ii_yeast2_abal1_abal2} and~\ref{fig:freq:spea2_yeast2_abal1_abal2}). Even though Abal$_2$ saw a drop in the duplication averages and standard deviation from the canonical to semantic methods, the poorer performance of Abal$_2$ can be attributed to the relatively large standard deviation in duplication average still present (Tables~\ref{tab:freq:nsgaii} and~\ref{tab:freq:spea2}).  
% Duplication tables start here

\subsection{Size of GP solutions}

\begin{figure}
  \centering
    \begin{tabular}{cc}
     \scriptsize{NSGA-II and variants} & \scriptsize{SPEA2 and variants}\\

     %\hspace{-0.82cm}  \includegraphics[width=0.370\textwidth]{figures/mo_spect_spea2}   & \hspace{-0.95cm}  \includegraphics[width=0.370\textwidth]{figures/mo_abalone_9_18_spea2} & \hspace{-0.95cm}  \includegraphics[width=0.370\textwidth]{figures/mo_abalone_9_other_spea2} 

\hspace{-0.82cm}  \includegraphics[width=0.50\columnwidth]{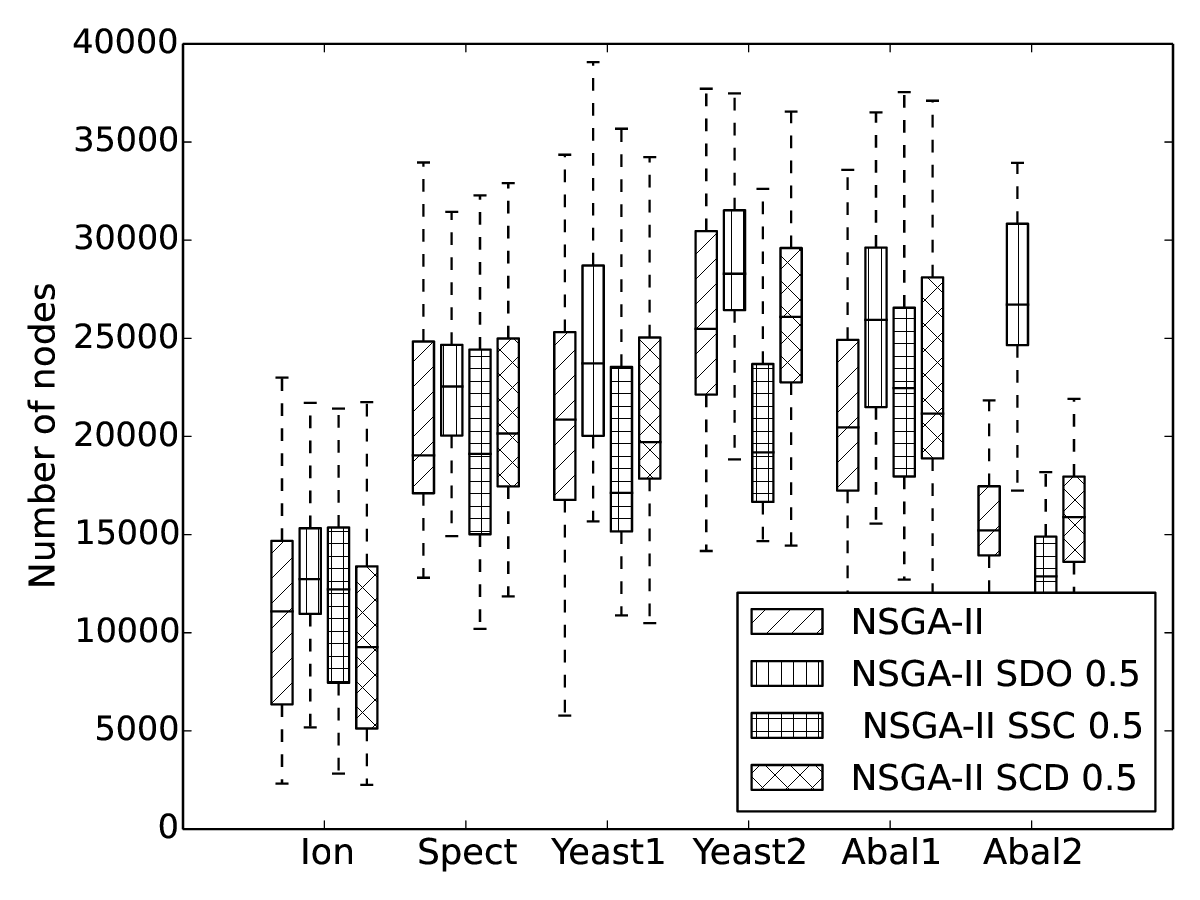}&
\hspace{-0.0cm}    \includegraphics[width=0.50\columnwidth]{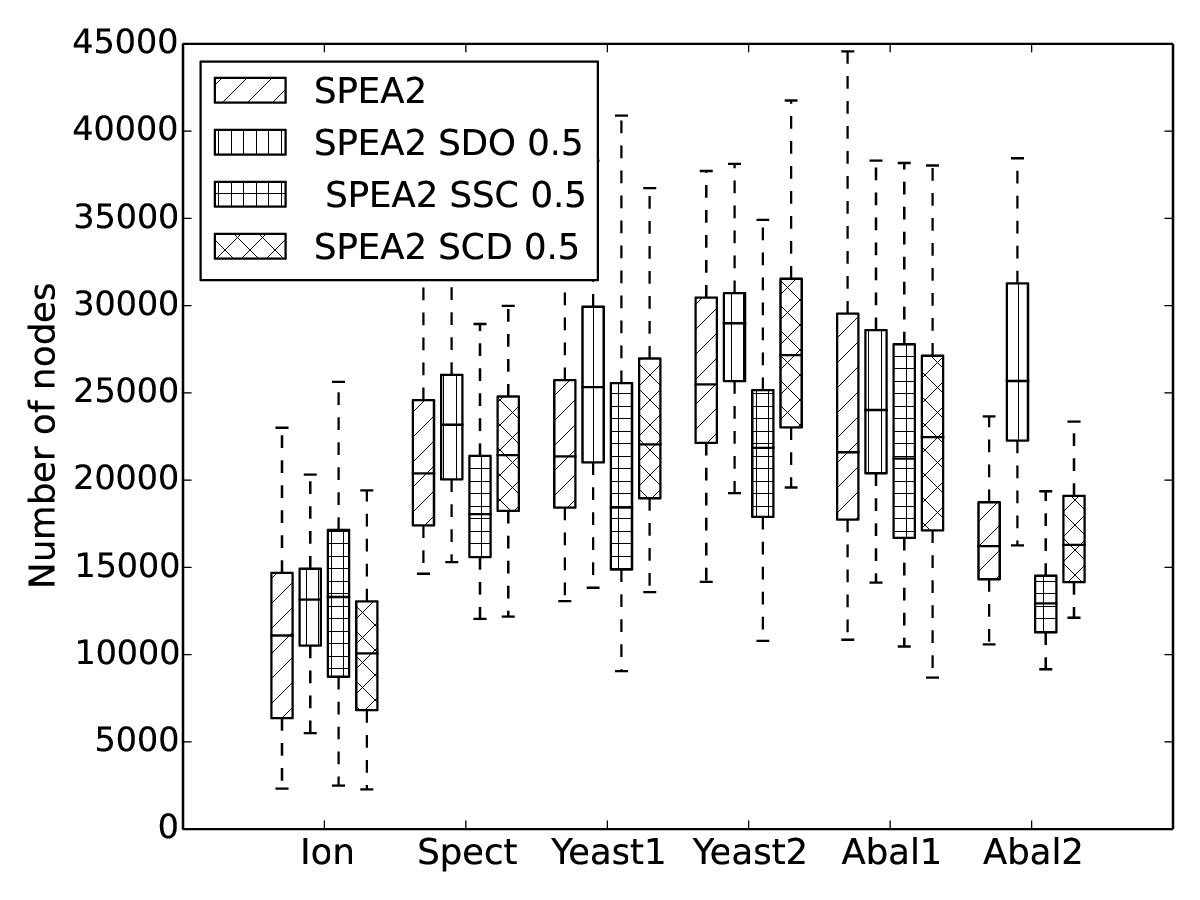}

\end{tabular}
\caption{Average size of GP individuals, computed by using the entire population per generation for each independent run, using NSGA-II (left) and SPEA2 (right) and their corresponding semantic-based variants (SDO, SSC and SCD, setting UBSS = 0.5).}
\label{fig:bloat}
\end{figure}

Bloat, increase in average tree size without a corresponding increase in performance, is a phenomenon commonly observed in GP variable length representations, such as the one used in this study. Figure~\ref{fig:bloat} shows the average number of nodes per run for each one of the 50 independent runs for each dataset and for each approach used in this work, including NSGA-II and SPEA2, shown in the left and in the right of Figure~\ref{fig:bloat}, respectively, including their semantic-based variants, setting UBSS = 0.5  for each of these three variants (SDO, SSC and SCD).

We have grouped, for each of the six datasets used in this work (see Table~\ref{tab:datasets}), the average number of nodes evaluated by either NSGA-II/SPEA2, SDO, SSC or SCD. From Figure~\ref{fig:bloat}, it is easy to observe that when we compare the semantic-based approaches (last three box-plots, from left to right, for each of the six datasets groups) either using NSGA-II (left) or SPEA (right), SDO (second box-plot, from left to right) tends to evaluate more nodes, hence  larger trees, compared to the other two semantic-based approaches (last two box-plots for each of these six groups) as well as their EMO counterparts (first box-plot for each of these groups). This is particularly visible in the Abal$_2$ dataset.

When we compare the number of nodes evaluated by SDO \textit{vs.} NSGA-II or SPEA2, left and right of Figure~\ref{fig:bloat}, respectively, we can see that SDO does not evaluate many more nodes compared to the two canonical EMO approaches when using the Ion, Spect and Yeast$_2$ datasets. From this analysis, it is interesting to note that there is a tendency to get a better Pareto front when more nodes are evaluated. For example, see in Figure~\ref{fig:pareto}, how the Pareto front by the SDO approach is better compared to the other three approaches in the Abal$_2$ dataset (right-hand side of the figure). SDO evaluates significantly more nodes compared to the other three approaches in this dataset (see right-hand side of Figure~\ref{fig:bloat}).

The same tendency is observed when using the Abal$_1$ dataset. SDO evaluates more nodes compared to NSGA-II (left of Figure~\ref{fig:bloat}) and achieves a better Pareto front (see middle-top of Figure~\ref{fig:pareto}). This analysis suggests that evaluating more nodes leads to better results in the objective space, which means this additional code growth should not be characterised as bloat.

\section{Conclusions}
\label{sec:conclusions} 
This work proposes a new approach, named Semantic-based Distance as an additional criteriOn (SDO), which consists of using semantic distance values as another criterion to optimise and preferences solutions that are semantically attracted to the sparsest region of the first approximated Pareto front. We also use this distance in lieu of the crowding distance at the heart of the aforementioned EMO algorithm. Results for the new approach were tested against the canonical frameworks of NSGA-II and SPEA2 and additionally two semantic-based methods were use as baselines, namely Semantic Similarity-based Crossover and Semantic-based Crowding Distance. It was found that SDO produced significantly better results when compared against each of these methods in terms of the hypervolume metric.

Our analysis shows that the SDO method produces more unique individuals compared to the other methods. A comparison of the first approximated Pareto fronts at specific generations showed that SDO not only attracts new individuals to the sparsest regions of the front but also reduces the amount of duplication, thus improving diversity.% Furthermore it was demonstrated that duplication not only reduces the amount of genetic variability in a population but can can also lead to a decrease in performance over the course of many generations when the solution set for the first approximated Pareto front grows larger than the population size, and the duplicated individual is disproportionately represented in the solution set. It was found that evaluating more nodes leads to better results in the objective space and as such the additional code growth observed in the SDO method is distinct from bloat as performance was not negatively affected.}  

%\textcolor{blue}{We have shown that our proposed approach achieves better results compared to other forms of semantics,} as well as the results obtained by canonical NSGA-II and SPEA2 algorithms. We have learned how it is feasible to promote semantic diversity in a MOGP by using a well-defined semantic-based distance.

\section*{Acknowledgments}

\noindent This publication has emanated from research conducted with the financial support of Science Foundation Ireland under Grant number 18/CRT/6049. The opinions, findings and conclusions or recommendations expressed in this material are those of the author(s) and do not necessarily reflect the views of the Science Foundation Ireland. The authors wish to acknowledge the \sloppy{DJEI/DES/SFI/HEA} Irish Centre for High-End Computing (ICHEC) for the provision of computational facilities and support. We would like to thank the reviewers for the comments made on the early version of this manuscript.

%\noindent This publication has emanated from research conducted with the financial support of Science Foundation Ireland under Grant number 18/CRT/6049.  The authors wish to acknowledge the DJEI/DES/SFI/HEA Irish Centre for High-End Computing (ICHEC) for the provision of computational facilities and support. We would like to thank the reviewers for the comments made on the early version of this manuscript.

%\section*{Acknowledgments}
%We would like to thanks the reviewers  for their useful comments that helped us to improve our work. We thank E. Mezura for his early involvement in this work. The authors wish to acknowledge the DJEI/DES/SFI/HEA Irish Centre for High-End Computing (ICHEC) for the provision of computational facilities and support.   

\bibliography{mogp_semantics_arXiv}

\end{document}